\documentclass[runningheads,11pt]{llncs}
\usepackage{blindtext}
\usepackage{geometry}
 \usepackage{setspace}
\usepackage{graphicx}
\usepackage[export]{adjustbox}

\usepackage{diagbox} %JON
\usepackage{cite} %JON

\usepackage{amssymb}

\usepackage{amsmath,mleftright}
\usepackage{xparse}

\NewDocumentCommand{\evalat}{sO{\big}mm}{%
  \IfBooleanTF{#1}
   {\mleft. #3 \mright|_{#4}}
   {#3#2|_{#4}}%
}

\usepackage[title]{appendix} %JON

\usepackage[table,xcdraw]{xcolor} %JON
\usepackage{subcaption} %JON
\usepackage{multirow} %JON
\usepackage{comment} %JON
\PassOptionsToPackage{hyphens}{url}\usepackage{hyperref} %JON

\usepackage{tablefootnote}
\usepackage[absolute,overlay]{textpos}
\usepackage{blindtext}
\usepackage{relsize}
\usepackage{hyperref}
\usepackage{tikz}

\tikzset{fontscale/.style={font=\relsize{#1}}}

\begin{document}

\begin{textblock*}{21cm}(1cm, 0.55cm)
  \begin{tikzpicture}
    \draw (0,0) rectangle (18,1); 
    \end{tikzpicture}
\end{textblock*} 

\begin{textblock*}{21cm}(-0.5cm, 0.55cm)
  \begin{tikzpicture}
    \node (center) {};
    \path (center)+(10.5,4-0.25) node [fontscale=-1] (name) {Final published version of the manuscript can be found at \href{https://doi.org/10.1016/j.sysarc.2023.102878}{10.1016/j.sysarc.2023.102878}.};
    \path (center)+(10.5,4+0.25) node [fontscale=-1] (name) {\copyright 2023. This manuscript version is made available under the CC-BY-NC-ND 4.0 license.};
    \end{tikzpicture}
\end{textblock*}

\title{On-chip Hyperspectral Image Segmentation with Fully Convolutional Networks for Scene Understanding in Autonomous Driving\thanks{This work was partially supported by the Basque Government under grants KK-2021/00111 and PRE\_2021\_1\_0113 and by the Spanish Ministry of Science and Innovation under grant PID2020-115375RB-I00. We thank the University of the Basque Country for its allocation of computational resources.}}% <-this % stops a space

\begin{spacing}{1}

\author{Jon Gutiérrez-Zaballa\inst{1}\orcidID{0000-0002-6633-4148} \and \\ Koldo Basterretxea\inst{2}\orcidID{0000-0002-5934-4735} \and \\ Javier Echanobe\inst{3}\orcidID{0000-0002-1064-2555} \and \\ M. Victoria Martínez\inst{3} \and \\ Unai Martinez-Corral\inst{2}\orcidID{0000-0003-1752-9181} \and \\ Óscar Mata-Carballeira\inst{2}\orcidID{0000-0002-1468-6280} \and \\ Inés del Campo \inst{3}\orcidID{0000-0002-6378-5357}}
\authorrunning{J. Gutiérrez-Zaballa et al.}
% First names are abbreviated in the running head.
% If there are more than two authors, 'et al.' is used.
%
\institute{Jon Gutiérrez-Zaballa is with the Department of Electronics Technology, \\ University of the Basque Country, 48013, Bilbao, Spain \email{j.gutierrez@ehu.eus} \and Researchers are with the Department of Electronics Technology, \\ University of the Basque Country, 48013, Bilbao, Spain \and Researchers are with the Department of Electricity and Electronics, \\
University of the Basque Country, 48940 Leioa, Spain}

\maketitle              % typeset the header of the contribution
\end{spacing}

\begin{abstract}
Most of current computer vision-based advanced driver assistance systems (ADAS) perform detection and tracking of objects quite successfully under regular conditions. However, under adverse weather and changing lighting conditions, and in complex situations with many overlapping objects, these systems are not completely reliable. The spectral reflectance of the different objects in a driving scene beyond the visible spectrum can offer additional information to increase the reliability of these systems, especially under challenging driving conditions. Furthermore, this information may be significant enough to develop vision systems that allow for a better understanding and interpretation of the whole driving scene. In this work we explore the use of snapshot, video-rate hyperspectral imaging (HSI) cameras in ADAS on the assumption that the near infrared (NIR) spectral reflectance of different materials can help to better segment the objects in real driving scenarios. To do this, we have used the HSI-Drive 1.1 dataset to perform various experiments on spectral classification algorithms. However, the information retrieval of hyperspectral recordings in natural outdoor scenarios is challenging, mainly because of deficient colour constancy and other inherent shortcomings of current snapshot HSI technology, which poses some limitations to the development of pure spectral classifiers. In consequence, in this work we analyze to what extent the spatial features codified by standard, tiny fully convolutional network (FCN) models can improve the performance of HSI segmentation systems for ADAS applications. In order to be realistic from an engineering viewpoint, this research is focused on the development of a feasible HSI segmentation system for ADAS, which implies considering implementation constraints and latency specifications throughout the algorithmic development process. For this reason, it is of particular importance to include the study of the raw image preprocessing stage into the data processing pipeline. Accordingly, this paper describes the development and deployment of a complete machine learning-based HSI segmentation system for ADAS, including the characterization of its performance on different embedded computing platforms, including a single board computer, an embedded GPU SoC and a programmable system on chip (PSoC) with embedded FPGA. We verify the superiority of the FPGA-PSoC over the GPU-SoC in terms of energy consumption and, particularly, processing latency, and demonstrate that it is feasible to achieve segmentation speeds within the range of ADAS industry specifications using standard development tools.

\end{abstract}

%\linenumbers

\section{Introduction}\label{sec:intro}
Today, thanks to the availability of small-size, portable, snapshot hyperspectral cameras, it is possible to set up HSI processing systems on moving platforms. The use of drones for precision agriculture and ecosystem monitoring is probably one of the most active and mature application domains of this technology \cite{govender2007review}. The research into how hyperspectral information can be used to develop more capable and robust ADAS is, on the contrary, in its infancy. HSI provides rich information about how materials reflect the light of different wavelengths (spectral reflection), and this can be used to identify and classify surfaces and objects in a scene \cite{colomb2019spectral, weikl2022potentials}. In fact, RGB images tend to suffer from metamerism (two objects with different reflection spectra but having the same colour under a certain light source), so HSI can help to resolve this phenomenon and become a powerful solution for object segmentation \cite{huang2021weakly}. Thus, with the application of appropriate information processing techniques, HSI can help to enhance the accuracy and robustness of current ADAS for object identification and tracking and, eventually, can be used for scene understanding, which is a step forward in the achievement of more capable and intelligent ADAS/ADS (Autonomous Driving Systems) \cite{pinchon2018all, winkens2019road}.

HSI segmentation of real driving scenes is, however, very challenging for a variety of reasons that are relative to both the lack of control over the environment (lighting, weather, distance of objects etc.) and the technology of portable snapshot HSI cameras, i.e. the use of spatial mosaic and tiled spectral filters, the limited spectral range of CMOS sensors and the lack of a photometer-based exposure-time setting. First, acquiring HSI outdoors implies working with varying illumination and environmental conditions and dealing with the presence of moving objects and changing backgrounds, which strongly conditions the setup of the acquisition system. In consequence, acquired images need to undergo a time-consuming preprocessing stage to transform a 2D image with radiance data into a 3D cube of reflectance values which, under these variable conditions and changing camera setups, can hardly guarantee the required robustness for the subsequent information processing. Secondly, the spectral signatures of the different objects of interest for ADAS/ADS in a driving scene may show a weak spectral separability when acquired under the conditions described herein. Thirdly, the extraction of additional spatial features that could help to improve the segmentation of items with similar spectral reflectance is problematic as a result of the enormous diversity of shapes, view angles, distances and scales in an image. Finally, ADS/ADAS development implies executing image processing and AI algorithms onboard and in real-time (the latter meaning with a sufficiently low latency), and this must be performed on resource constrained hardware platforms with tight power consumption restrictions. This entails not only evaluating the performance indexes (recall/precision) of the inferred classification or segmentation maps but carefully analyzing the computational complexity of the algorithms involved and how to efficiently implement them on the target devices.

In this work, which is an extended version of \cite{gutierrez22}, we describe an ML approach based on FCN for the segmentation of HSI video recorded in real driving conditions as part of a research project which aims at exploring the applicability of small-size snapshot hyperspectral sensor technology to the enhancement of vision-based ADAS. This is a multidisciplinary work that includes investigating in an integrated manner HSI processing techniques, machine learning and AI algorithms and models, and processor and advanced SoC design techniques to develop a functional real-time image segmentation system that can realistically be implemented in an embedded processing system. We present two application examples of the design of a HSI segmentation system that could be applied to ADS/ADAS. These examples are used throughout the paper to show the system development process, from the algorithmic design, (i.e. ML model training and optimization), to the final system prototyping and testing on different embedded processing systems: embedded multicore microprocessors, an embedded GPU SoC, and a PSoC with an embedded FPGA. The first, simpler example aims at the segmentation of images by the pixelwise classification of three categories: tarmac, road marks and "rest" or non-drivable areas, which could be used to enhance automatic lane keeping and trajectory planning systems for ADS. The second example incorporates two additional categories, vegetation and sky, to make a total of five different categories to be classified, which extends the system's ability for a broader scene understanding. The aim was to evaluate, first, to what extent deep convolutional segmentation models such as U-Net, which perform successfully in image segmentation tasks in other application fields, can help to overcome the limitations revealed in previous stages of our research of purely spectral classifiers. The second aim was to evaluate whether it is pertinent to try out the implementation of such models into embedded processing platforms, considering the real-time requirements of the application.

To do so, we have trained and evaluated the segmentation algorithms under study using a HSI database specifically created for this purpose \cite{basterretxeahsi}. Various FCN variants have been used in order to analyze their performance on the experimental data. However, in this approach we have avoided complex, compound FCN models to keep the model sizes under control. Moreover, the selected FCN model hyperparameters have finally been optimized in order to achieve a good trade-off between segmentation accuracy and model complexity to reduce computing time (FLOPS) and memory footprint (reduced number of parameters). As stated before, small-size hyperspectral snapshot cameras require quite a complex preprocessing pipeline to get the 3D cubes containing spectral reflectance data from the recorded radiance raw images. For an ADS/ADAS application, this preprocessing must be carried out on-the-fly and on the onboard processing device. In addition, we include an ablation study to observe the effect of those techniques separately on each of the stages of the preprocessing. Finally, we have conducted a comprehensive benchmarking to exhaustively characterize the prototype performances in terms of latency, power and energy consumption, both at the image preprocessing stage and at the inference of the deployed network. Besides this, the benchmarking also considers the comparison between the execution of the FCN models using only the processing systems (microprocessors) with the combined execution of the processing system and the dedicated coprocessors for AI inference (GPU and FPGA), which comprises a previous quantization of the models. In fact, the effect of the quantization process is also addressed to check how it affects the previously mentioned measurements as well as the memory footprint and the segmentation accuracy

The main contributions of this paper are summarized as follows:\begin{itemize}
    \item The spectral data analysis and the development of HSI segmentation algorithms for scene understanding in ADAS are performed on images obtained with a snapshot camera in real driving scenarios and under diverse lighting and weather conditions. This guarantees that the results obtained are realistic and can be extrapolated to an eventual real implementation based on this same technology. 
    \item Some inherent limitations of the spectral information obtained from multispectral filter-array based HSI snapshot cameras are identified and addressed in the development of HSI segmentation algorithms.
    \item A lightweight FCN is developed for the efficient segmentation of HSI images for scene understanding in ADAS. This improves the performance of pure spectral classifiers while remaining simple enough to be deployed successfully on an embedded computing platform.
    \item Real HSI segmentation prototypes are implemented and their performance characterized in terms of memory footprint, latency, power and energy consumption by the deployment of the whole processing pipeline on different commercial embedded heterogeneous computing platforms: an embedded multicore microprocessor, an embedded GPU SoC, and a PSoC with embedded FPGA. This involves analyzing not only the FCN inference stage for image segmentation but also the raw image preprocessing stage that precedes the formation of the hyperspectral cubes.
\end{itemize}

The rest of the paper is organized as follows: Section \ref{sec:relatedWork} contains the most relevant works regarding Deep Learning-based segmentation with either RGB or HSI in ADAS as well as articles where the superiority of HSI over RGB is shown. In Section \ref{sec:expSetup} we present the experimentation setup, including an analysis of the database which shows the implications of acquiring HSI images outdoors and in challenging environments. Some initial results with baseline spectral are included too, where a comparison between segmentation with HSI or pseudoRGB is included. Section \ref{sec:FCNs} includes the FCN model development for which a detailed hyperparameter tuning process has been carried out. Besides this, it also covers the way in which the training has been conducted and presents the segmentation results. Section \ref{sec:workflow} contains the deploying of both the FCN and the cube preprocessing algorithm on the three embedded computing platforms whose performance is deeply benchmarked in terms of latency, power and energy consumption, resource usage and accuracy. Section \ref{sec:conc} concludes the paper and discusses possible future work.

\section{Related work}\label{sec:relatedWork}
There is very limited prior work on ADAS segmentation in HSI, so we first summarize related work on segmentation with HSI outside ADAS and deep learning-based ADAS segmentation in RGB data.

\subsection{Deep learning-based segmentation with RGB data in ADAS}
Semantic segmentation of urban scenes has been attracting the attention of researchers since the publication of widely known databases such as KITTI Stereo and Flow Benchmark \cite{Alhaija2018IJCV} (200 annotated training images and 200 test images of urban environments), CamVid \cite{BrostowSFC:ECCV08, BrostowFC:PRL2008} (700 cityscape images), Cityscapes \cite{cordts2016cityscapes} (5000 finely-annotated and 20 000 weakly-annotated images), although there are also other databases that focus on other environments such as \cite{kim2020highway}, which provides 1200 annotated images from highway scenarios. In most cases, these databases are used to compare different models according to various metrics that are usually focused on accuracy results without paying attention to either network complexity or energy consumption, neglecting their possible deployment in an embedded application.

In fact, when analyzing the architectures that report remarkable results in the Cityscapes database, it can be seen that most of them contain tens of millions of parameters and require billions of floating point operations (GFLOPS) at inference. For instance, Deeplab \cite{chen2017deeplab}, a deep convolutional neural network which includes atrous convolution for dense feature extraction and field-of-view enlargement, contains tens of millions of parameters. SwiftNet \cite{orsic2019defense}, another deep convolutional encoder-decoder network which is based on an interleaved pyramid fusion model, contains, depending on the configuration, from 2.4M to 24.7M parameters and performs from 41 to 218 GFLOPs when run on an NVIDIA GTX 1080Ti GPU. STDC \cite{fan2021rethinking}, a convolutional model which features various short-term dense concatenate modules (STDC) to fuse outputs from convolutions with different receptive fields and ends with a fully connected layer, contains, depending on the architecture, from 8.44M to 813M parameters and performs from 0.01 to 1.45 GFLOPS when run on an NVIDIA GTX 1080Ti. Finally, TinyHMSeg \cite{li2020humans} is an encoder-decoder network which instantiates MobileNetV2 as the lightweight backbone and step-wisely reuses the intermediate feature maps, downsamples them and feeds them to medium- and low-resolution branches. Unlike those mentioned above, TinyHMSeg is the only one that aims to achieve high performance results while remaining lightweight; it contains 0.7M parameters and  performs 3 GFLOPs when run on an NVIDIA GTX 1080Ti for real-time applications.

Finally, it should be noted that there is a line of research discussed in \cite{courdier2020real} that promotes precisely latency-aware, real-time video segmentation for which it seeks to leverage the temporal correlation of consecutive frames in a video to improve the next-frame prediction and reduce computation burden and latency.

\subsection{RGB versus HSI for image segmentation}
The hyperspectral reflectance data of different material surfaces outside the visible spectrum can provide very valuable information to improve intelligent vision tasks and, particularly, for image segmentation. Outside ADAS, HSI is being applied in various fields such as remote sensing for geoscience applications, food assessment and biomedical image analysis. With regards to remote sensing, in \cite{fricker2019convolutional}, to identify tree species in mixed-conifer forests, high spatial resolution airborne hyperspectral imagery and a convolutional neural network are used. The most relevant aspect of this article is that the authors show that the HSI CNN model outperformed the RGB CNN model by 23\% on average, taking into account all metrics and species. This indicates that the additional spectral information provided by HSI is essential for increasing model performance. In the field of food assessment, in \cite{taghizadeh2011comparison}, they show
the greater potential of HSI compared with conventional RGB imaging for predicting L-value of mushrooms, a method for mushroom quality grading. Finally, concerning biomedical image analysis, especially interesting is the work presented in \cite{seidlitz2022robust} where they compare five deep learning models, varying both the spatial (from pixel-based segmentation to image-based segmentation) and spectral (RGB, domain-specific tissue parameter images (TPI) and HSI) granularities. They show that unprocessed HSI data (not TPI) offers a huge benefit compared to RGB or TPI data for organ segmentation, although this superiority is reduced when using more complex networks such as FCNs, which take advantage not only of the richness of the spectral signature of the materials but also of the spatial properties of the images.

\subsection{Deep learning-based segmentation with HSI data in ADAS}
The main hypothesis of this line of research is that spectral information beyond the visible range can improve the accuracy and robustness of current RGB-based ADAS for object identification and tracking and, eventually, can be used for scene understanding. Leaving aside the studies carried out with thermal cameras (far infrared), there is not much previous work published in the field \cite{borges2022survey}. Nevertheless, there are some researchers that have started to explore the applicability of HSI cameras to the field of ADS. In \cite{lu20}, a hyperspectral image dataset for road segmentation in urban and rural scenes is introduced and in \cite{gutierrez22} the authors explore the use of FCNs to segment HSI applied to ADAS. The underlying idea is that the incorporation of richer spectral information can provide a distinct spectral fingerprint for each entity in the image, helping to achieve more precise and robust detection systems. 

In the study conducted by \cite{colomb2019spectral}, the authors characterize the reflection of a wide variety of road objects, such as a work zone signaling cone, a road reflector, an asphalt pavement, a dark blue fleece sweater, a tyre and an exhaust pipe in the 350 to 2450nm range using a spectroradiometer. They show that there is a great difference among the spectral signatures of all the measured materials and remark that the zone of greatest reflectance is between 750 and 1550nm (NIR-SWIR bands). In the same way, in \cite{pinchon2018all} they evaluate the relevance of four different spectral bands: RGB, NIR, SWIR and LWIR when detecting and recognizing pedestrians, vehicles, traffic signs and lanes both in outdoor natural conditions and in an artificial fog tunnel. They conclude that, overall, the NIR band is the most useful spectral band followed by SWIR, RGB and LWIR.
Another work which follows the same line is \cite{weikl2022potentials}, where the authors generate realistic spectral radiance data of synthetic scenes in the VIS and NIR spectral range and assess the benefits of VIS-NIR imaging in their synthetic images using exemplary night time and daylight traffic scenes. Their results for daylight scenes confirm the benefits of using the additional information provided by the NIR spectral band for metamerism resolution, although they acknowledge that the nighttime scene results are not conclusive and more models of artificial light sources should be added to the database. In \cite{liyanage20}, based on the uniqueness of the spectral signature of different materials, the authors propose a pixel-wise labeling process aided by hyperspectral imaging which can, to a great extent, reduce the manual labelling process. Another paper that deals with the metamerism phenomenon is \cite{huang2021weakly}, where they introduce their hyperspectral dataset for semantic segmentation of cityscapes and remark that metamerism is particularly challenging in these scenes because they contain too many classes, complex lighting and spatial structures. They perform tSNE visualization of RGB and HSI data, and show that HSI images have a stronger separability. One of the earliest studies in this regard, although not specifically aimed at ADAS, draws interesting conclusions regarding the ability of images taken with hyperspectral cameras to detect people in urban environments \cite{Herweg:13}. Quite interestingly, one of the outcomes of the work is the finding that the use of the VNIR spectrum (up to 1000 nm) offers similar results to using the full spectrum of the used sensor (up to 2500nm). This is very relevant to the present proposal, since current CMOS sensor-based low-cost HSI snapshot cameras do not offer spectrometric information beyond 1000 nm.

Another initial hypothesis of this research consists of the assumption that deep-learning algorithms developed for image segmentation in the field of ADAS will take advantage of hyperspectral data, either to get lighter network models for the same task  (since the input information is richer, it is expected that the more informative HSI images will require a less complex processing pipeline and so, they will be more suitable for embedded applications) or by improving the performance obtained with RGB images. \cite{cavigelli2016computationally} is a paper that focuses exactly on analyzing these two aspects for a video-surveillance system. They show that while using a pixel-wise classification neural network, the error rate decreases by 30\% when using RGB-NIR data together compared to using RGB data alone. In addition, they take into account the hard power constraints of embedded platforms and train the RGB-NIR data on two convolutional networks with smaller computational effort, getting the same error rate as the RGB-only pixel-wise neural network, while decreasing the computational load by 300\%. A pioneering research group in the investigation of the use of HSI in ADS is the Active Vision Group (AGAS) of the University of Koblenz-Landau. They have published various papers on this topic, reporting interesting results on image segmentation and terrain classification applied to images combining VIS and NIR spectrum information from HSI snapshot cameras. In \cite{winkens2019shadow} the authors satisfactorily combine a 16-band image in the visible spectrum with a 25-band image in the NIR spectrum obtaining promising results and improve the performance of a method which only relies on a RGB image and a 1-band NIR image. In \cite{winkens2017features}, the work is focused on extracting robust spatial and spectral features for snapshot hyperspectral data which are invariant to illumination changes and then train a per-pixel classification network. In \cite{winkens2018hyperspectral}, the authors propose a snapshot hyperspectral scene analysis pipeline which combines a per-pixel classification with context-aware, fully connected, conditional random fields leading to a more accurate pixel-level classification performance. Of special interest is \cite{winkens2019road}, where the authors use a deep autoencoder with additional regularization terms to learn the latent space of the input hyperspectral snapshot data and obtain a reduced feature set which, when trained with deep learning methods, outperforms the RGB data networks.

In summary, prior work on semantic segmentation with real HSI data from snapshot cameras in ADAS is scarce and, in general, little attention has been paid to the realistic deployment of these system and to their characterization in terms of processing performance. This includes analyzing not only the segmentation process itself, but also the data acquisition and the raw image preprocessing stages. There are some works that use synthetic databases that are not capable of accurately reproducing the complex illumination of real traffic scenarios \cite{weikl2022potentials}. Many papers are aimed at detecting or identifying road objects but not at performing semantic segmentation of images for scene understanding \cite{pinchon2018all,colomb2019spectral}. The few works that combine real data with image segmentation tasks do not address the problem of analyzing how these systems can be realistically deployed on a resource and power constrained embedded computing platform while keeping processing latency values in the range of what is acceptable for ADAS \cite{winkens2017features, winkens2018hyperspectral, winkens2019road, winkens2019shadow, huang2021weakly}.

\section{Experimental Setup}\label{sec:expSetup}

\subsection{The Dataset}\label{subsec:dataset}
As is reported in \cite{basterretxeahsi}, there are very few HSI datasets aimed at developing ADAS and ADS applications. In \cite{basterretxeahsi}, the authors present a dataset named HSI-Drive, which was specifically conceived to provide the research community with data obtained at real driving scenarios with a small size, portable, snapshot hyperspectral camera capable of recording 25-band HSI at video rates. The current version of the dataset, i.e. HSI Drive v1.1, contains 276 manually labelled images recorded while driving a car in urban, road and highway scenarios under diverse weather (sunny, cloudy, rainy and foggy) and lighting (dawn, midday, sunset) conditions. The recordings were performed during Spring (121 images) and Summer (155 images). The recording setup included a single Photonfocus camera featuring an Imec 25-band VIS-NIR (535nm-975nm) mosaic spectral filter over a CMOSIS CMV200 image wafer sensor. The resolution of the raw images was 1088 x 2048 pixels with 5$\mu m$ x 5$\mu m$ size. However, as the spectral bands were extracted from a mosaic formed by 5x5 pixel window Fabri-Perot filters, the output resolution of the HSI cubes was actually reduced to 216 x 409 x 25 \cite{mv1-d2048x1088-hs02-96-g2}. Indeed, the use of this technology implies performing an image preprocessing stage prior to the processing pipeline to get the spectral cubes from the recorded radiance raw images, an issue that is thoroughly addressed in Subsection \ref{subsec:imagePre} of this paper.

\begin{figure}[h!]
\begin{subfigure}{0.48\linewidth}
\centering
\includegraphics[width = 8.7cm]{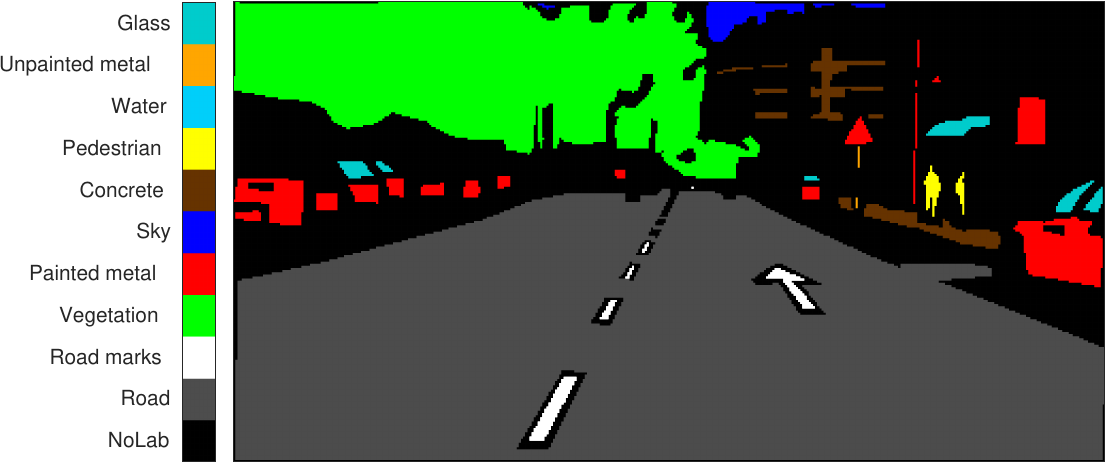}
\caption{Ground truth image.}
\label{fig:detailedLabellingVisible}
\end{subfigure}
\begin{subfigure}{0.48\linewidth}
\centering
\includegraphics[width = 6.9cm]{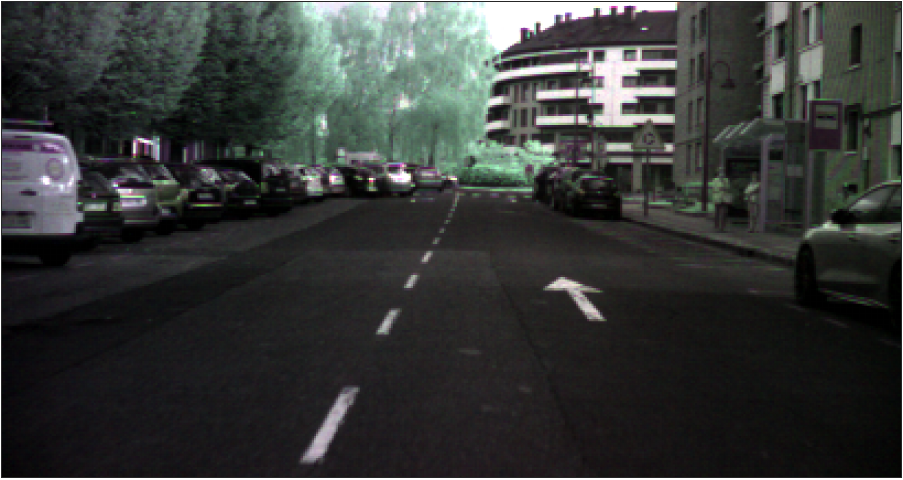}
\caption{False-color visible image.}
\label{fig:detailedLabellingGT}
\end{subfigure}
\caption{Ground truth image (a) and its corresponding false-color visible image (b) of an urban scenario as an example of the weak labelling methodology followed to develop the HSI-Drive dataset.}
\label{fig:detailedLabelling}
\end{figure}

The original labelling of the scene images in the dataset comprises 10 different classes related mostly to the expected differences in the surface reflectance signatures of various materials: metal, vegetation, concrete, tarmac etc. Available ground truth masks however are not fully dense, since the labelling method was aimed at favoring the accuracy in the spectral information provided for each class by reducing spectral mixing. This means, for instance, that contour pixels of many items and the surfaces on the background that cannot be precisely distinguished have not been labeled (Figure \ref{fig:detailedLabelling}). While this method of labelling is not an issue for non-convolutional ML models, it may have some adverse effects on the performance of CNN-based segmentation models. In fact, the training of convolutional neural networks with "weak supervision" on sparsely labelled datasets is a line of research itself (see for instance \cite{wang2020weakly} and references therein).

\newpage

Regarding the spectral characteristics of the data, in \cite{gutierrez22} the authors provide spectral separability figures of all 10 classes calculated on all available data in the HSI-Drive v1.1 dataset. In particular, they use the Jeffreys-Matsushita (JM) index, which ranges from 0 (null separability) to 2.0 (perfect separability) as the preferred criterion, which is relevant in this work since this index provides an estimation of the probability of a correct prediction to happen \cite{forestier2013comparison}. As can be observed in Table \ref{tab:separability}, which we reproduce here for clarity, the separability of some of the class pairs such as Road/Road Marks (1.92), Road/Sky (1.98) or Road Marks/Unpainted Metal (1.92) is promisingly high, while class-pairs such as Road/Concrete (1.44), Road Marks/Painted Metal (1.63) and Painted Metal/Unpainted Metal (1.35) show low separability indexes. In general, JM index values below 1.9 imply that designing classification algorithms based on pure spectral reflectance data may be challenging or infeasible, and that more meaningful and relevant features should be extracted from the raw data to achieve good performance in an image segmentation task.

\begin{table}[h!]
\centering
\resizebox{12.25cm}{!}{%
\begin{tabular}{c|c|c|c|c|c|c|c|c|c|}
\cline{2-10}
 & \textbf{Road} & \textbf{Road M.} & \textbf{Veg.} & \textbf{P. Met.} & \textbf{Sky} & \textbf{Conc.} & \textbf{Ped.} & \textbf{Unp. Met.} & \textbf{Glass} \\ \hline
\multicolumn{1}{|c|}{\textbf{Road}} &  & 1.92 & 1.83 & 1.65 & 1.98 & 1.44 & 1.84 & 1.42 & 1.49 \\ \cline{1-1}
\multicolumn{1}{|c|}{\textbf{Road Marks}} & 1.92 &  & 1.79 & 1.63 & 1.87 & 1.68 & 1.92 & 1.92 & 1.93 \\ \cline{1-1}
\multicolumn{1}{|c|}{\textbf{Vegetation}} & 1.83 & 1.79 &  & 1.44 & 1.96 & 1.64 & 1.81 & 1.73 & 1.81 \\ \cline{1-1}
\multicolumn{1}{|c|}{\textbf{Painted Metal}} & 1.65 & 1.63 & 1.44 &  & 1.91 & 1.48 & 1.70 & 1.35 & 1.48 \\ \cline{1-1}
\multicolumn{1}{|c|}{\textbf{Sky}} & 1.98 & 1.87 & 1.96 & 1.91 &  & 1.97 & 1.98 & 1.98 & 1.80 \\ \cline{1-1}
\multicolumn{1}{|c|}{\textbf{Concrete}} & 1.44 & 1.68 & 1.64 & 1.48 & 1.97 &  & 1.74 & 1.62 & 1.68 \\ \cline{1-1}
\multicolumn{1}{|c|}{\textbf{Pedestrian}} & 1.84 & 1.92 & 1.81 & 1.70 & 1.98 & 1.74 &  & 1.79 & 1.66 \\ \cline{1-1}
\multicolumn{1}{|c|}{\textbf{Unpainted Metal}} & 1.42 & 1.92 & 1.73 & 1.35 & 1.98 & 1.62 & 1.79 &  & 1.38 \\ \cline{1-1}
\multicolumn{1}{|c|}{\textbf{Glass}} & 1.49 & 1.93 & 1.81 & 1.48 & 1.80 & 1.68 & 1.66 & 1.38 &  \\ \hline
\multicolumn{1}{|c|}{\textbf{Mean}} & \textbf{1.73} & \textbf{1.80} & \textbf{1.78} & \textbf{1.63} & \textbf{1.96} & \textbf{1.69} & \textbf{1.82} & \textbf{1.68} & \textbf{1.71} \\ \hline
\end{tabular}}
\caption{Jeffreys-Matsushita (JM) interclass distances which range from 0 (null separability) to 2.0 (perfect separability).}
\label{tab:separability}
\end{table}

Beyond possible similarities of the ``real" physical spectral reflectance footprints of different material surfaces in the spectral range of the sensor, low separability indexes between categories in the dataset can be a consequence of three main factors: redundant information in the spectral dimension, high variability within the same category due to changing lighting and different camera set-ups, and the inherent technological limitations of mosaic-filters to accurately separate the spectral radiance that enters the sensor. Examples of this are the presence of crosstalks, the variability produced by different light beam angles of incidence and the spectral mixing produced in the demosaicing process. 

\begin{figure}[h!]
\begin{subfigure}{0.48\linewidth}
\centering
\includegraphics[width = 8cm]{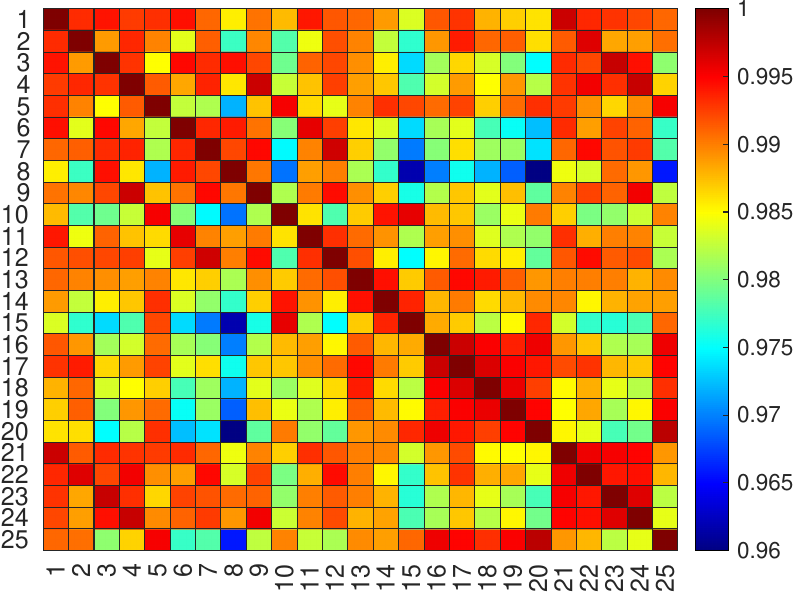}
\caption{}
\label{fig:bandCorrelation}
\end{subfigure}
\begin{subfigure}{0.48\linewidth}
\centering
\includegraphics[width = 7.6cm]{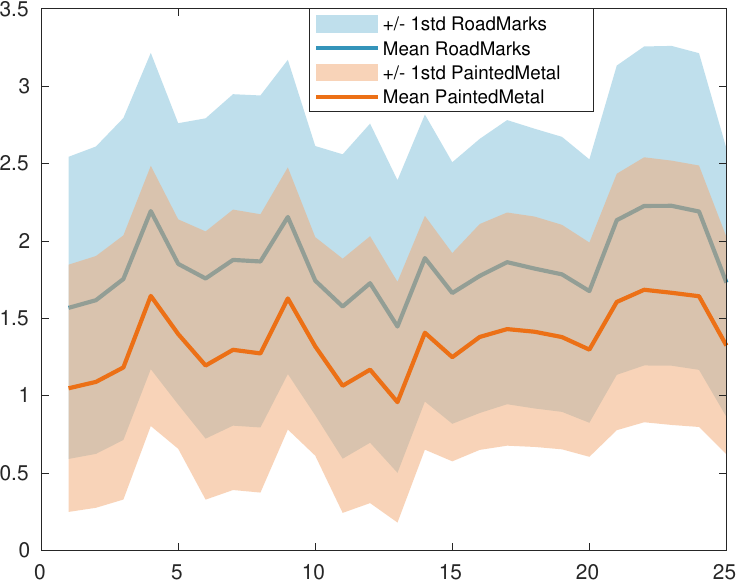}
\caption{Spectral signature of certain classes.}
\label{fig:minMeanMaxSpectralSignature}
\end{subfigure}
\caption{Pearson Correlation Coefficients among the 25 spectral bands (a) and spectral signature of some of the representative classes (b) of the HSI-Drive 1.1 dataset.}
\end{figure}

The aforementioned limitations associated with snapshot technology lead to a correlation matrix (Figure \ref{fig:bandCorrelation}) which shows a strong Pearson Correlation Coefficient among the bands.

\newpage

\begin{figure}[h!]
\begin{subfigure}{0.48\linewidth}
\centering
\includegraphics[width=8cm]{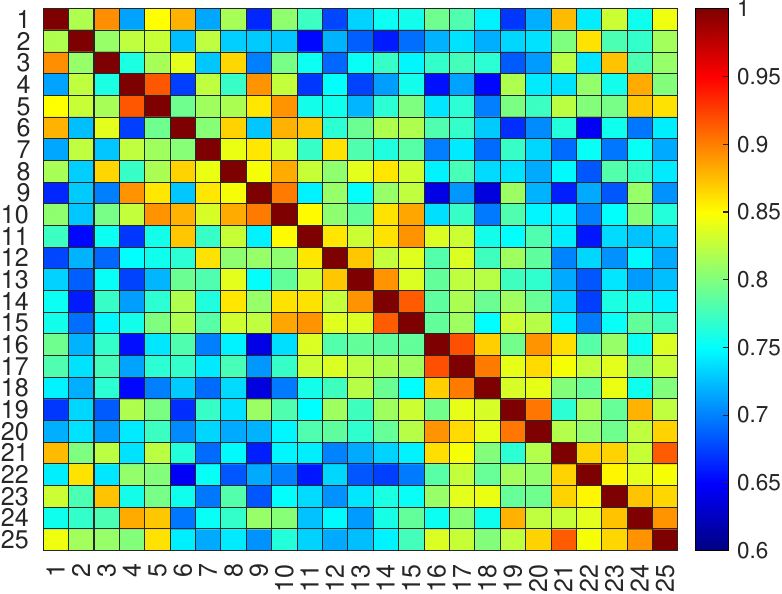}
\caption{}
\label{fig:intraClassCorrelationRoadMarks}
\end{subfigure}
\begin{subfigure}{0.48\linewidth}
\centering
\includegraphics[width=8cm]{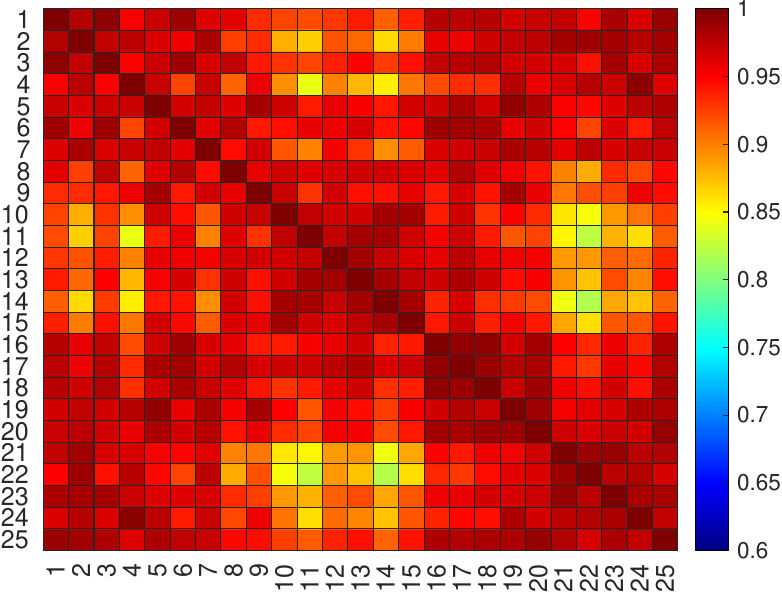}
\caption{}
\label{fig:intraClassCorrelationPaintedMetal}
\end{subfigure}
\caption{Intraclass Pearson Correlation Coefficients for the 25 bands of two representative classes of the dataset: Road Marks (a) and Painted Metal (b).}
\label{fig:intraClassCorrelations}
\end{figure}

In order to determine the existing intraclass variability, we have used our own spectral library. This library consists of 10 images (one for each class) which have been taken on the same day, at the same time and with the same camera configuration. The objective is to ascertain what the properties of the materials are in the most controlled conditions as possible (while still acquiring them outdoors with the snapshot hyperspectral camera). Figure \ref{fig:intraClassCorrelations} shows the maximum intrinsic intraclass Pearson Correlation Coefficients of two representative classes of the dataset. It can be seen that although Painted Metal is quite homogeneous, the dispersion of Road Marks is heavily affected by the nature of the materials.

Finally, Figure \ref{fig:minMeanMaxSpectralSignature} displays the area covered by the mean $\pm$ 1std of the spectral signatures of the same two classes, using more than 250,000 pixels for each class. The Figure confirms that interclass variance is high and results in a wide overlapping among spectral signatures which, on the other hand, is something common  in other HSI datasets (see \cite{bai2023achieving} for instance).

\begin{figure}[h!]
\begin{subfigure}{0.48\linewidth}
\centering
\includegraphics[width = 7.5cm]{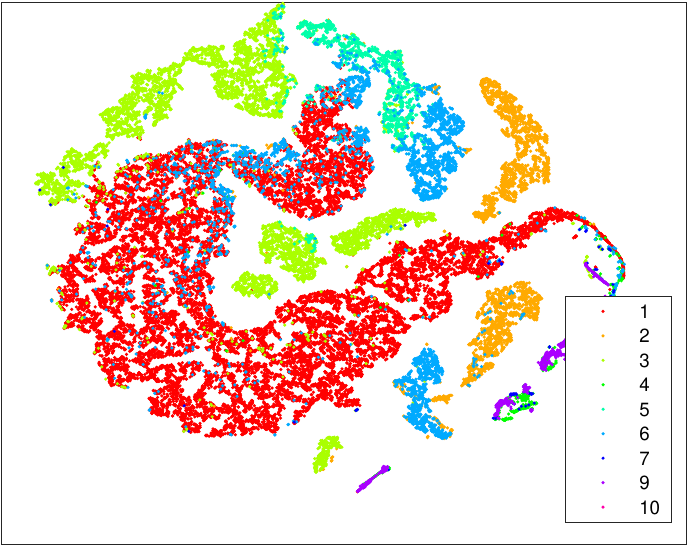}
\caption{3-channel t-SNE.}
\label{fig:tsne3clases}
\end{subfigure}
\begin{subfigure}{0.48\linewidth}
\centering
\includegraphics[width = 7.5cm]{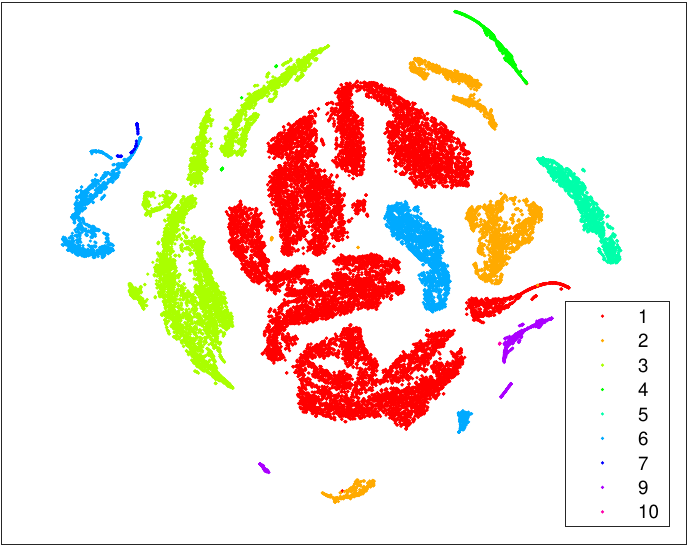}
\caption{25-channel t-SNE.}
\label{fig:tsne25clases}
\end{subfigure}
\caption{Comparison of t-SNE output when using either the 3 most representative channels (a) or the full 25 channels (b).}
\label{fig:tsne}
\end{figure}

Despite the issues described above and the challenges posed by the use of data acquired in real outdoor driving scenarios, as mentioned in Section \ref{sec:relatedWork}, there are numerous works that confirm the benefits of using HSI in combination (or not) with RGB over using only RGB images for image segmentation. Since the camera used to develop the dataset does not cover the entire visible spectral range, it is not possible to generate equivalent RGB images to perform a proper comparison of classification performance to standard RGB. It is possible however to select the three most informative bands out of the original 25 in the dataset and generate pseudoRGB images to perform meaningful comparative analyses. In particular, we used the orthogonal space projection method proposed in \cite{du2008similarity} to identify the most informative spectral bands from every image in the dataset and then selected the most frequent (mode) values, which happened to be the triplet (8, 21, 24) in the mosaic filter. Applying a t-Distributed Stochastic Neighbor Embedding (t-SNE) algorithm \cite{van2008visualizing}, it is possible to map the data on a 2D to graphically visualize the separability of the classes. Figure \ref{fig:tsne3clases} shows the result of applying the t-SNE to the pseudoRGB images, while in Figure \ref{fig:tsne25clases}, the results when using all 25 bands are shown. It can be seen that when using all bands, the data representation is not only free of overlapping among classes, but is also more compact, as clusters of the same class are closer together. The contribution of the hyperspectral information compared to pseudoRGB data is further analyzed in Subsection \ref{subsec:baseline} by evaluating the performance of baseline spectral classifiers.

\subsection{Data Partitioning and Application Examples}\label{subsec:FCNs}
In order to investigate the potential performance of segmentation systems based on the training of ML models with this dataset, throughout this work we use two application examples that can be employed in the development of ADAS/ADS. The first example focuses on segmenting images into three classes: road (tarmac), road marks and non-drivable areas, the latter including the remaining classes in the labelled masks. This low-complexity segmentation example would be aimed at developing a system for the discrimination of drivable and non-drivable zones, together with a lane-keeping aid for ADAS or even a trajectory planning system for ADS. In the second example, two additional classes have been added to the model training (vegetation and sky), which show quite satisfactory spectral separability indexes (see Table \ref{tab:separability}) and add relevant information for a scene understanding application. The exploration of more complex segmentation models including all 10 classes in the dataset has also been investigated and some segmented images are shown in Section \ref{sec:FCNs}. However, the results obtained are irregular and inconclusive, and will require further research.  

In order to perform an ML training on this dataset, the 276 available images have been divided into training, validation and testing subsets. The selection of images has not been aleatory but conducted following a data-diversity criterion. The generated subsets contain 162 images for training, 57 for validation and 57 for testing, preserving class proportionality in all three subsets. As for the metrics used in the evaluation of the performance tests, these have been Recall (R), Precision (P), and Intersection over union (IoU), which is an equivalent for the IoU for object detection. As Equations \ref{equ:recall}, \ref{equ:precision} and \ref{equ:iou} show, recall accounts for the false negatives (FN) and true positives (TP), precision takes into account the false positives (FP) and the true positives too and IoU combines both aspects:

\begin{equation}
  R_i = \frac{TP_i}{TP_i + FN_{i}}
  \qquad
  WR = \sum\limits_{i=1}^n w_i R_i 
  \label{equ:recall}
\end{equation}

\begin{equation}
  P_i = \frac{TP_i}{TP_i + FP_{i}}
  \qquad
  WP = \sum\limits_{i=1}^n w_i P_i
  \label{equ:precision}
\end{equation}

\begin{equation}
  IoU_i = \frac{TP_i}{TP_i + FN_{i} + FP_{i}}
  \qquad
  WIoU = \sum\limits_{i=1}^n w_i IoU_i
\label{equ:iou}
\end{equation}

\noindent where $i$ is the class index such that, for example, $FN_{i}$ accounts for the pixels that have been predicted as not belonging to class i, but are actually part of class i. 
As a consequence of the dataset being heavily imbalanced (Table \ref{tab:datasetPartition}), a correct interpretation of the indexes requires not only a representation of the global metrics, but also the mean values and, more specifically, the weighted scores of performance indexes. Accordingly, weighting factors related to the inverse of the frequency of the classes in the dataset have been added to the final calculation of the indexes.

\begin{table}[h!]
\centering
\resizebox{16cm}{!}{%
\begin{tabular}{ccccccccccc}
\cline{4-11}
 &
   &
  \multicolumn{1}{c|}{} &
  \multicolumn{3}{c|}{\textbf{3-class experiment}} &
  \multicolumn{5}{c|}{\textbf{5-class experiment}} \\ \cline{3-11} 
 &
  \multicolumn{1}{c|}{} &
  \multicolumn{1}{c|}{\textbf{Total}} &
  \multicolumn{1}{c|}{\textbf{Road}} &
  \multicolumn{1}{c|}{\textbf{Marks}} &
  \multicolumn{1}{c|}{\textbf{No Dri.}} &
  \multicolumn{1}{c|}{\textbf{Road}} &
  \multicolumn{1}{c|}{\textbf{Marks}} &
  \multicolumn{1}{c|}{\textbf{Veg.}} &
  \multicolumn{1}{c|}{\textbf{Sky}} &
  \multicolumn{1}{c|}{\textbf{Other}} \\ \hline
\multicolumn{1}{|c|}{\multirow{2}{*}{\textbf{Train}}} &
  \multicolumn{1}{c|}{\textbf{Num. pixels}} &
  \multicolumn{1}{c|}{13 343 314} &
  \multicolumn{1}{c|}{7 947 254} &
  \multicolumn{1}{c|}{450 703} &
  \multicolumn{1}{c|}{4 945 627} &
  \multicolumn{1}{c|}{7 947 254} &
  \multicolumn{1}{c|}{450 703} &
  \multicolumn{1}{c|}{3 127 191} &
  \multicolumn{1}{c|}{568 411} &
  \multicolumn{1}{c|}{1 250 025} \\ \cline{2-11} 
\multicolumn{1}{|c|}{} &
  \multicolumn{1}{c|}{\textbf{\%}} &
  \multicolumn{1}{c|}{100} &
  \multicolumn{1}{c|}{59.56} &
  \multicolumn{1}{c|}{3.38} &
  \multicolumn{1}{c|}{37.06} &
  \multicolumn{1}{c|}{59.56} &
  \multicolumn{1}{c|}{3.38} &
  \multicolumn{1}{c|}{23.44} &
  \multicolumn{1}{c|}{4.26} &
  \multicolumn{1}{c|}{9.37} \\ \hline
 &
   &
   &
   &
   &
   &
   &
   &
   &
   &
   \\ \cline{3-11} 
 &
  \multicolumn{1}{c|}{} &
  \multicolumn{1}{c|}{\textbf{Total}} &
  \multicolumn{1}{c|}{\textbf{Road}} &
  \multicolumn{1}{c|}{\textbf{Marks}} &
  \multicolumn{1}{c|}{\textbf{No Dri.}} &
  \multicolumn{1}{c|}{\textbf{Road}} &
  \multicolumn{1}{c|}{\textbf{Marks}} &
  \multicolumn{1}{c|}{\textbf{Veg.}} &
  \multicolumn{1}{c|}{\textbf{Sky}} &
  \multicolumn{1}{c|}{\textbf{Other}} \\ \hline
\multicolumn{1}{|c|}{\multirow{2}{*}{\textbf{Test}}} &
  \multicolumn{1}{c|}{\textbf{Num. pixels}} &
  \multicolumn{1}{c|}{3 514 081} &
  \multicolumn{1}{c|}{2 067 379} &
  \multicolumn{1}{c|}{99 426} &
  \multicolumn{1}{c|}{1 347 276} &
  \multicolumn{1}{c|}{2 067 379} &
  \multicolumn{1}{c|}{99 426} &
  \multicolumn{1}{c|}{820 804} &
  \multicolumn{1}{c|}{163 127} &
  \multicolumn{1}{c|}{363 345} \\ \cline{2-11} 
\multicolumn{1}{|c|}{} &
  \multicolumn{1}{c|}{\textbf{\%}} &
  \multicolumn{1}{c|}{100} &
  \multicolumn{1}{c|}{58.83} &
  \multicolumn{1}{c|}{2.83} &
  \multicolumn{1}{c|}{38.34} &
  \multicolumn{1}{c|}{58.83} &
  \multicolumn{1}{c|}{2.83} &
  \multicolumn{1}{c|}{23.36} &
  \multicolumn{1}{c|}{4.64} &
  \multicolumn{1}{c|}{10.34} \\ \hline
\end{tabular}}
\caption{Frequency of each of the classes in the train/val and test subsets for the 3-class and 5-class experiments.}
\label{tab:datasetPartition}
\end{table}

\subsection{Reference Baseline Spectral Classifiers}\label{subsec:baseline}
In a first approach to the problem, baseline neural classifiers were trained relying only on spectral data. The results obtained, in addition to exposing the foreseeable limitations of segmentation based on pure, non-processed spectral features, will be useful for assessing fairly the performance results of the convolutional models presented in the following sections.

We trained medium-depth, fully connected ANNs with three hidden layers to perform image segmentation on the per-pixel classification basis. First, spectral reflectance values were normalized through division by the sum of the reflectance at all 25 wavelengths to alleviate the adverse effects of differing illumination at the image level. The models featured Hyperbolic Tangent activation functions at every layer and included an input z-score normalization layer and a Softmax output layer.  Due to large dataset size and memory limits in the NVIDIA GFORCE RTX-3090 used for training (24 GB), the optimization was performed by an Adam (Adaptive Moment estimation) algorithm with 2\textsuperscript{20} mini-batch size, data shuffling at every epoch and a limit of 300 epochs. Selected network parameter settings corresponded to the best cross-entropy loss value in the validation set. Due to severe class imbalance in the training dataset, various techniques such as under-sampling to the smallest class dataset size, data augmentation by SMOTE algorithm, weighting of the loss function and changing to a Focal Loss function were tried out. None of these techniques produced entirely satisfactory results, we thus applied a combination of techniques by generating some synthetic data to reduce class imbalance. Then, a weighted, focal loss function was set, based on data frequency. Nevertheless, the weights were finally manually adjusted until satisfactory results were achieved.

After various attempts, ic-25-100-100-c ANN models, where ic stands for the number of input channels and c for the number of classes to be categorised, showed good performance, while adding more nodes/parameters did not produce any accuracy improvements. Table \ref{tab:metricsFloatingANN} sums up the results obtained for the two example applications under study. While performance figures are generally acceptable, when per image indexes are analyzed, there is a lack of robustness in the predictions. In particular, the precision of minority classes tends to be low, with a high rate of false positives in some images, which is reflected in the lower weighted scores. Figures \ref{fig:comparison3class} and \ref{fig:comparison5class} show produced segmentation for three example images in the testing set, one for each type of driving scenario: urban, road and highway. As can be seen, produced segmented images are overall correct, producing interpretable scenarios, but the lack of precision, which produces incorrectly classified pixel infiltration, mostly from the majority class into minority class segments, is noticeable. 

\begin{table}[h!]
\centering
\resizebox{9cm}{!}{%
\begin{tabular}{ccccccc}
\cline{2-7}
\multicolumn{1}{c|}{} & \multicolumn{3}{c|}{\textbf{25 channels}} & \multicolumn{3}{c|}{\textbf{pseudoRGB}}\\
\cline{2-7}
\multicolumn{1}{c|}{} & \multicolumn{1}{c|}{\textbf{Recall}} & \multicolumn{1}{c|}{\textbf{Precision}} & \multicolumn{1}{c|}{\textbf{IoU}}& \multicolumn{1}{c|}{\textbf{Recall}} & \multicolumn{1}{c|}{\textbf{Precision}} & \multicolumn{1}{c|}{\textbf{IoU}} \\
\hline
\multicolumn{1}{|c|}{\textbf{Road}} & \multicolumn{1}{c|}{89.39} & \multicolumn{1}{c|}{92.91} & \multicolumn{1}{c|}{83.67} & \multicolumn{1}{c|}{81.70} & \multicolumn{1}{c|}{78.37} & \multicolumn{1}{c|}{66.66}\\
\hline
\multicolumn{1}{|c|}{\textbf{Road Marks}} & \multicolumn{1}{c|}{48.88} & \multicolumn{1}{c|}{29.66} & \multicolumn{1}{c|}{22.64}& \multicolumn{1}{c|}{04.47} & \multicolumn{1}{c|}{19.93} & \multicolumn{1}{c|}{03.79} \\
\hline
\multicolumn{1}{|c|}{\textbf{Non-Drivable}} & \multicolumn{1}{c|}{90.34} & \multicolumn{1}{c|}{89.42} & \multicolumn{1}{c|}{81.62}& \multicolumn{1}{c|}{67.94} & \multicolumn{1}{c|}{68.48} & \multicolumn{1}{c|}{51.75} \\
\hline
\multicolumn{1}{|c|}{\textbf{Overall}} & \multicolumn{1}{c|}{\textbf{88.61}} & \multicolumn{1}{c|}{\textbf{89.78}} & \multicolumn{1}{c|}{\textbf{81.16}}  & \multicolumn{1}{c|}{\textbf{74.24}} & \multicolumn{1}{c|}{\textbf{72.92}} & \multicolumn{1}{c|}{\textbf{59.17}}\\
\hline
\multicolumn{1}{|c|}{\textbf{Mean}} & \multicolumn{1}{c|}{\textbf{76.21}} & \multicolumn{1}{c|}{\textbf{70.66}} & \multicolumn{1}{c|}{\textbf{62.64}}  & \multicolumn{1}{c|}{\textbf{51.37}} & \multicolumn{1}{c|}{\textbf{55.59}} & \multicolumn{1}{c|}{\textbf{40.73}}\\
\hline
\multicolumn{1}{|c|}{\textbf{Weighted}} & \multicolumn{1}{c|}{\textbf{53.35}} & \multicolumn{1}{c|}{\textbf{36.30}} & \multicolumn{1}{c|}{\textbf{29.13}}  & \multicolumn{1}{c|}{\textbf{11.95}} & \multicolumn{1}{c|}{\textbf{25.63}} & \multicolumn{1}{c|}{\textbf{09.64}}\\
\hline
 &  &  &  \\ \hline
\multicolumn{1}{|c|}{\textbf{Road}} & \multicolumn{1}{c|}{85.71} & \multicolumn{1}{c|}{94.18} & \multicolumn{1}{c|}{81.39}& \multicolumn{1}{c|}{59.91} & \multicolumn{1}{c|}{87.22} & \multicolumn{1}{c|}{55.07} \\
\hline
\multicolumn{1}{|c|}{\textbf{Road Marks}} & \multicolumn{1}{c|}{54.34} & \multicolumn{1}{c|}{26.85} & \multicolumn{1}{c|}{21.91} & \multicolumn{1}{c|}{38.80} & \multicolumn{1}{c|}{08.32} & \multicolumn{1}{c|}{07.36}\\
\hline
\multicolumn{1}{|c|}{\textbf{Vegetation}} & \multicolumn{1}{c|}{95.05} & \multicolumn{1}{c|}{90.29} & \multicolumn{1}{c|}{86.24} & \multicolumn{1}{c|}{75.95} & \multicolumn{1}{c|}{77.45} & \multicolumn{1}{c|}{62.20}\\
\hline
\multicolumn{1}{|c|}{\textbf{Sky}} & \multicolumn{1}{c|}{96.05} & \multicolumn{1}{c|}{82.76} & \multicolumn{1}{c|}{80.36} & \multicolumn{1}{c|}{69.21} & \multicolumn{1}{c|}{30.27} & \multicolumn{1}{c|}{26.68}\\
\hline
\multicolumn{1}{|c|}{\textbf{Other}} & \multicolumn{1}{c|}{56.72} & \multicolumn{1}{c|}{54.52} & \multicolumn{1}{c|}{38.50}& \multicolumn{1}{c|}{24.85} & \multicolumn{1}{c|}{19.97} & \multicolumn{1}{c|}{12.45} \\
\hline
\multicolumn{1}{|c|}{\textbf{Overall}} & \multicolumn{1}{c|}{\textbf{84.49}} & \multicolumn{1}{c|}{\textbf{86.74}} & \multicolumn{1}{c|}{\textbf{76.35}}  & \multicolumn{1}{c|}{\textbf{59.87}} & \multicolumn{1}{c|}{\textbf{73.11}} & \multicolumn{1}{c|}{\textbf{49.66}}\\
\hline
\multicolumn{1}{|c|}{\textbf{Mean}} & \multicolumn{1}{c|}{95.85} & \multicolumn{1}{c|}{69.719} & \multicolumn{1}{c|}{68.84}& \multicolumn{1}{c|}{53.75} & \multicolumn{1}{c|}{44.65} & \multicolumn{1}{c|}{32.75} \\
\hline
\multicolumn{1}{|c|}{\textbf{Weighted}} & \multicolumn{1}{c|}{\textbf{70.16}} & \multicolumn{1}{c|}{\textbf{52.45}} & \multicolumn{1}{c|}{\textbf{46.55}}& \multicolumn{1}{c|}{\textbf{48.64}} & \multicolumn{1}{c|}{\textbf{22.32}} & \multicolumn{1}{c|}{\textbf{18.12}} \\ \hline
\end{tabular}}
\caption{Evaluation of the ANN on the 3-classes (above-top) and 5-classes (above) test datasets in terms of recall, precision and IoU (Intersection over Union).}
\label{tab:metricsFloatingANN}
\end{table}

\begin{figure}[h!]
\centering
\includegraphics[height=7cm]{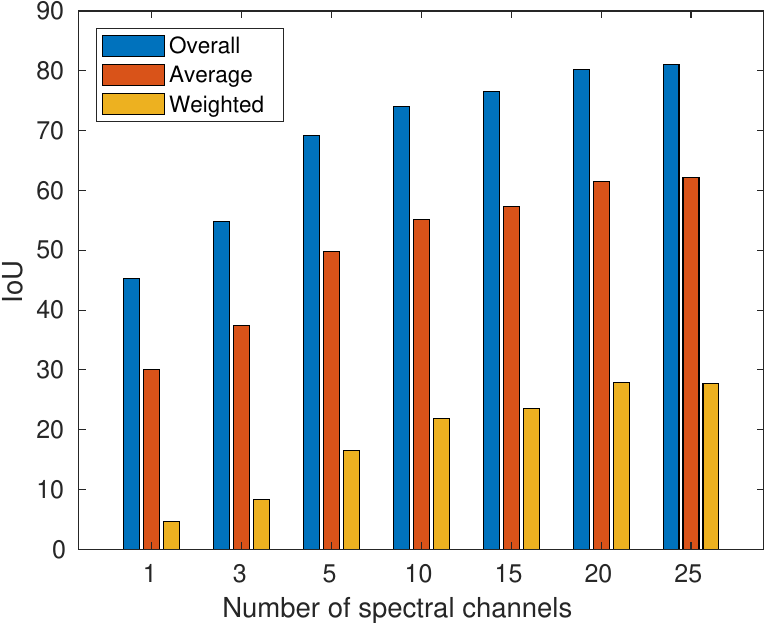}
\caption{Overall, average and weighted IoU (\%) as a function of the number of spectral channels.}
\label{fig:barPlot}
\end{figure}

The same experiment was performed using pseudoRGB images generated with the three most VNIR informative bands for each application example. After various training setting adjustments, obtained best results are summed up in Table \ref{tab:metricsFloatingANN}. As can be observed, the classifiers are unable to generate meaningful results, thus confirming, as expected, the impossibility of designing an image segmentation system based only on spectral reflectance data with just three channels, and demonstrating the superiority of HSI in this task. Additionally, the ANN model has been iteratively trained for an increasing number of spectral band data in the input vectors, ranging from just 1 band to all 25 bands. Figure \ref{fig:barPlot} shows obtained performance scores on the testing set for the first segmentation example (3 classes). It can be noted that, despite the strong spectral cross-correlation of bands, the accuracy of the classifier improves as more spectral information is provided to the input which, in principle, discourages performing spectral channel selection to reduce complexity.

\section{FCNs for HSI Image Segmentation}\label{sec:FCNs}
It is well known that incorporating spatial information to the spectral data in HSI can be an effective means to overcome the limits of the spectral imaging technology for image segmentation. The use of convolutional filters with tunable parameters in CNNs to extract spatial information has shown excellent performance in many applications. In particular, FCNs are neural computing architectures specifically aimed at image segmentation. However, when convolutional models are designed and trained with disregard for the computational constraints imposed by the target processing devices and the requirements of the application (power, latency), the results obtained may prove to be impractical. In this section we describe the development of an FCN that enhances the accuracy and robustness of the spectral classifiers while keeping model architecture in the range of tens of thousands of parameters for efficient embedded processing.

\subsection{Model development}
U-Net is an FCN-based architecture intended for accurate image segmentation. This model is of the type encoder-decoder, which means that after an encoding stage, there is a decoding or ``deconvolution" up-sampling stage that re-projects the extracted characteristics, and recovers the original image resolution to directly produce segmentation maps. Intermediate skip connections enable the fusion of low- and high-level features. U-Net was originally intended for biological image segmentation \cite{unet} but has been successfully used in other segmentation tasks, such as precision agriculture \cite{wang2020weakly}, food quality assessment \cite{son2021u} and aerial city recognition \cite{cui2019multiscale}. The idea of using an FCN to process HSI is to combine the intrinsic spectral characteristics of the different classes with the spatial relationships extracted by the convolution operations.

\begin{figure}[h!]
\centering
\includegraphics[height=9.5cm]{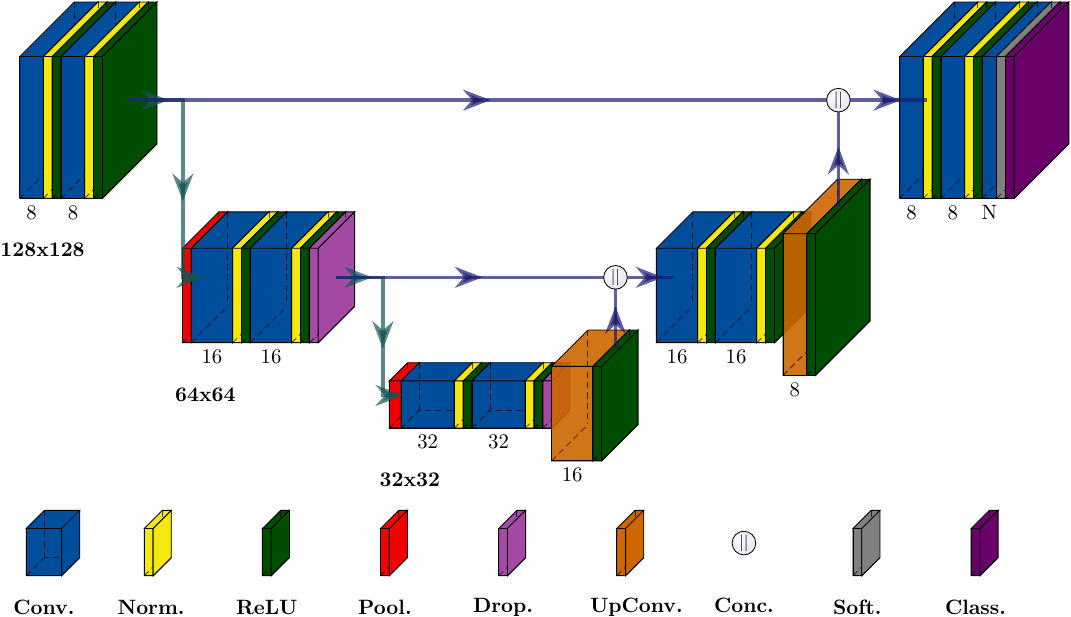}
\caption{Architecture of the modified U-Net.} \label{fig:miUnet}
\end{figure}

We have tailored the original architecture of U-Net to the unique characteristics of the dataset (see Figure \ref{fig:miUnet}, made with PlotNeuralNet software \cite{haris_iqbal_2018_2526396}) by including batch-normalization layers after convolutional filters, adding a dropout layer before the decoder, adapting the padding of the convolutional filters and modifying the network depth and the number of filters in the initial convolutional block. Besides this, we have also customized some training-related parameters which will be discussed when the training procedure is described.

\begin{figure}
\centering
\includegraphics[width=8cm]{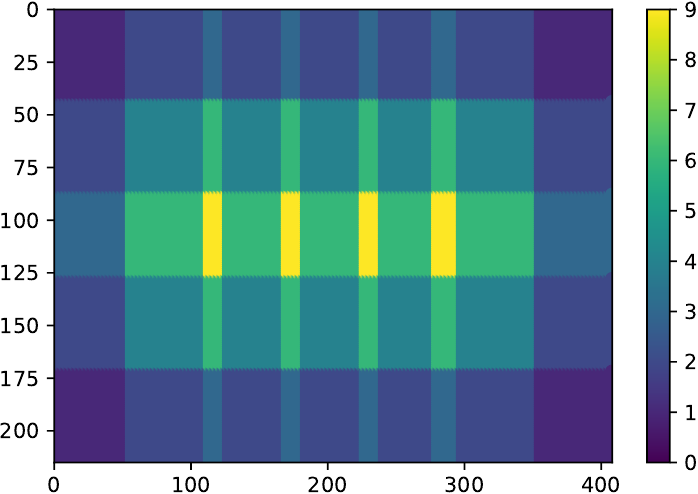}
\caption{Strides: 44 vertical and 57 horizontal.}
\label{fig:vert44hor57}
\caption{Example of the stride selection effect in the overlapping matrix index using 18 (3 x 6) patches.}
\label{fig:mosaicoSolapamiento}
\end{figure}

In addition to what has been explained in the previous paragraph, to get the best trade-off between segmentation performance and computational complexity, a hyperparameter optimization process has been carried out by programming a grid search of the optimum combination of a set of model hyperparameters. The optimization performance has been evaluated by calculating the segmentation accuracy on a subset of 45 images selected from all possible environments and weather conditions. After an initial evaluation of the model's performance, the selected set of analyzed hyperparameters included: the size of the input image patches [64x64, 128x128], the encoder depth [2, 3, 4] and the number of filters in the first convolutional block [8, 16, 32]. Additionally, the optimization procedure also included the influence of two parameters related to the training process: the initial learning rate of the optimization algorithm (Adam) and the stride of the patch creation process. Regarding the initial learning rate, we opted for varying it in decreasing powers of 10 from $10^{-1}$ to $10^{-4}$. As for the stride, we have opted for training the FCN with patches and not with the full resolution image so as to, on the one hand, have a lighter neural network and, on the other hand, benefit from the fact that as some of the patches overlap in certain pixels, there is more information to correctly predict the class to which those pixels belong. How the patches overlap depends both on the starting position of the initial patch (upper left side of the image in our case) and on the value given to the horizontal and vertical strides. We have decided that the areas where we want the segmentation to be as accurate as possible (and consequently locate the most overlapping patches) should be the central area of the image (a centrosymmetric overlapping scheme as Figure \ref{fig:mosaicoSolapamiento} shows) and accordingly set the vertical and horizontal stride values.

 The overlapping matrix is defined in Equation \ref{equ:OI}, which is simply the sum of the patch masks (binary matrices of the same size as the input image containing ones where the patch is defined and zeros everywhere else).

\begin{equation}
    OI = \sum_{n=1}^{N} {pm}_i \in \mathbb{R}^{3}
    \label{equ:OI}
\end{equation}

where $N$ is the number of patches and ${pm}_i$ is the patch mask associated to patch number $i$. 
The results from the hyperparameter tuning process are gathered in Tables \ref{tab:resultadosJavi1_64} and \ref{tab:resultadosJavi1_128}. It can be seen that using a larger 128 patch-size tends to produce a lower validation error than when using the smaller 64 patch-size. With regard to the number of initial filters and the encoder depth there is not a clear pattern and the results depend on the particular combination of different parameter values.

In order to establish a numerical criterion for the selection of the best model, we have evaluated two metrics that give an insight into the memory footprint and the computing complexity of the model architecture: the total number of parameters (NP) and the number of floating point operations (FLOPS). Both metrics depend on some other variables that were frozen as we did not take them into account in the hyperparameter tuning study. More explicitly, the number of parameters, as shown in Equation \ref{equ:n_params}, is a function of the encoder depth ($ed$), the initial number of filters ($if$), the convolution/up-convolution kernel sizes ($cks$, $ucks$), the number of classes to be predicted ($c$) and the number of input image channels ($ic$). The number of FLOPS (Equation \ref{equ:n_FLOPS}) depends on the preceding parameters as well as on the input image spatial dimensions ($id$) and other architectural choices such as the strides of the convolution/deconvolution and pooling layers.

\newpage

\begin{table}[h!]
\begin{center}
%\rotatebox{90}{
\scalebox{0.87}{
\begin{tabular}{ccccccccccccccccccc}
\cline{1-6} \cline{8-12} \cline{14-18}
\multicolumn{1}{|c|}{\textbf{Initial num. of filt.}} &
  \multicolumn{5}{c|}{{\color[HTML]{6200C9} \textbf{8}}} &
  \multicolumn{1}{c|}{} &
  \multicolumn{5}{c|}{{\color[HTML]{6200C9} \textbf{16}}} &
  \multicolumn{1}{c|}{} &
  \multicolumn{5}{c|}{{\color[HTML]{6200C9} \textbf{32}}} \\ \hline 
\multicolumn{1}{|c|}{\diagbox[]{\textbf{Img. size}}{\textbf{Enc. depth}}} &
  \multicolumn{1}{c|}{\color[HTML]{009901} \textbf{2}} &
  \multicolumn{1}{c|}{} &
  \multicolumn{1}{c|}{\color[HTML]{442605} \textbf{3}} &
  \multicolumn{1}{c|}{} &
  \multicolumn{1}{c|}{\color[HTML]{0FC6D8} \textbf{4}} &
  \multicolumn{1}{c|}{} &
  \multicolumn{1}{c|}{{\color[HTML]{009901} \textbf{2}}} &
  \multicolumn{1}{c|}{} &
  \multicolumn{1}{c|}{\color[HTML]{442605} \textbf{3}} &
  \multicolumn{1}{c|}{} &
  \multicolumn{1}{c|}{\color[HTML]{0FC6D8} \textbf{4}} &
  \multicolumn{1}{c|}{} &
  \multicolumn{1}{c|}{{\color[HTML]{009901} \textbf{2}}} &
  \multicolumn{1}{c|}{} &
  \multicolumn{1}{c|}{\color[HTML]{442605} \textbf{3}} &
  \multicolumn{1}{c|}{} &
  \multicolumn{1}{c|}{\color[HTML]{0FC6D8} \textbf{4}} \\ \hline 
\multicolumn{1}{|c|}{{\color[HTML]{FE0000} \textbf{64}}} &
 \multicolumn{17}{c|}{{\color[HTML]{F8A102} \textbf{0,100}}} \\ \hline
\multicolumn{1}{|c|}{\textbf{\textit{maxValidationAccuracy}}} &
  \multicolumn{1}{c|}{90.55} &
  \multicolumn{1}{c|}{} &
  \multicolumn{1}{c|}{90.48} &
  \multicolumn{1}{c|}{} &
  \multicolumn{1}{c|}{92.74} &
  \multicolumn{1}{c|}{} &
  \multicolumn{1}{c|}{91.11} &
  \multicolumn{1}{c|}{} &
  \multicolumn{1}{c|}{90.97} &
  \multicolumn{1}{c|}{} &
  \multicolumn{1}{c|}{93.29} &
  \multicolumn{1}{c|}{} &
  \multicolumn{1}{c|}{92.56} &
  \multicolumn{1}{c|}{} &
  \multicolumn{1}{c|}{94.49} &
  \multicolumn{1}{c|}{} &
  \multicolumn{1}{c|}{94.52} \\ \hline 
\multicolumn{1}{|c|}{\textbf{\textit{minValidationLoss}}} &
  \multicolumn{1}{c|}{0.28} &
  \multicolumn{1}{c|}{} &
  \multicolumn{1}{c|}{0.35} &
  \multicolumn{1}{c|}{} &
  \multicolumn{1}{c|}{0.23} &
  \multicolumn{1}{c|}{} &
  \multicolumn{1}{c|}{0.24} &
  \multicolumn{1}{c|}{} &
  \multicolumn{1}{c|}{0.28} &
  \multicolumn{1}{c|}{} &
  \multicolumn{1}{c|}{0.20} &
  \multicolumn{1}{c|}{} &
  \multicolumn{1}{c|}{0.25} &
  \multicolumn{1}{c|}{} &
  \multicolumn{1}{c|}{0.17} &
  \multicolumn{1}{c|}{} &
  \multicolumn{1}{c|}{0.20} \\ \hline 
\multicolumn{1}{|c|}{\textbf{\textit{maxTrainingAccuracy}}} &
  \multicolumn{1}{c|}{94.34} &
  \multicolumn{1}{c|}{} &
  \multicolumn{1}{c|}{94.66} &
  \multicolumn{1}{c|}{} &
  \multicolumn{1}{c|}{96.08} &
  \multicolumn{1}{c|}{} &
  \multicolumn{1}{c|}{95.58} &
  \multicolumn{1}{c|}{} &
  \multicolumn{1}{c|}{94.42} &
  \multicolumn{1}{c|}{} &
  \multicolumn{1}{c|}{96.52} &
  \multicolumn{1}{c|}{} &
  \multicolumn{1}{c|}{94.97} &
  \multicolumn{1}{c|}{} &
  \multicolumn{1}{c|}{96.17} &
  \multicolumn{1}{c|}{} &
  \multicolumn{1}{c|}{97.35} \\ \hline   
\multicolumn{1}{|c|}{\textbf{\textit{minTrainingLoss}}} &
  \multicolumn{1}{c|}{0.21} &
  \multicolumn{1}{c|}{} &
  \multicolumn{1}{c|}{0.17} &
  \multicolumn{1}{c|}{} &
  \multicolumn{1}{c|}{0.13} &
  \multicolumn{1}{c|}{} &
  \multicolumn{1}{c|}{0.16} &
  \multicolumn{1}{c|}{} &
  \multicolumn{1}{c|}{0.14} &
  \multicolumn{1}{c|}{} &
  \multicolumn{1}{c|}{0.11} &
  \multicolumn{1}{c|}{} &
  \multicolumn{1}{c|}{0.15} &
  \multicolumn{1}{c|}{} &
  \multicolumn{1}{c|}{0.12} &
  \multicolumn{1}{c|}{} &
  \multicolumn{1}{c|}{0.07} \\ \hline
  \multicolumn{1}{c|}{} &
  \multicolumn{17}{c|}{{\color[HTML]{F8A102} \textbf{0,01}}} \\ \hline
\multicolumn{1}{|c|}{\textbf{\textit{maxValidationAccuracy}}} &
  \multicolumn{1}{c|}{\textbf{94.65}} &
  \multicolumn{1}{c|}{} &
  \multicolumn{1}{c|}{\textbf{96.67}} &
  \multicolumn{1}{c|}{} &
  \multicolumn{1}{c|}{\textbf{96.48}} &
  \multicolumn{1}{c|}{} &
  \multicolumn{1}{c|}{94.79} &
  \multicolumn{1}{c|}{} &
  \multicolumn{1}{c|}{96.41} &
  \multicolumn{1}{c|}{} &
  \multicolumn{1}{c|}{\textbf{96.34}} &
  \multicolumn{1}{c|}{} &
  \multicolumn{1}{c|}{94.94} &
  \multicolumn{1}{c|}{} &
  \multicolumn{1}{c|}{96.32} &
  \multicolumn{1}{c|}{} &
  \multicolumn{1}{c|}{94.80} \\ \hline  
\multicolumn{1}{|c|}{\textbf{\textit{minValidationLoss}}} &
  \multicolumn{1}{c|}{\textbf{0.16}} &
  \multicolumn{1}{c|}{} &
  \multicolumn{1}{c|}{\textbf{0.13}} &
  \multicolumn{1}{c|}{} &
  \multicolumn{1}{c|}{\textbf{0.16}} &
  \multicolumn{1}{c|}{} &
  \multicolumn{1}{c|}{0.18} &
  \multicolumn{1}{c|}{} &
  \multicolumn{1}{c|}{0.14} &
  \multicolumn{1}{c|}{} &
  \multicolumn{1}{c|}{\textbf{0.12}} &
  \multicolumn{1}{c|}{} &
  \multicolumn{1}{c|}{0.17} &
  \multicolumn{1}{c|}{} &
  \multicolumn{1}{c|}{0.14} &
  \multicolumn{1}{c|}{} &
  \multicolumn{1}{c|}{0.20} \\ \hline 
\multicolumn{1}{|c|}{\textbf{\textit{maxTrainingAccuracy}}} &
  \multicolumn{1}{c|}{\textbf{97.50}} &
  \multicolumn{1}{c|}{} &
  \multicolumn{1}{c|}{\textbf{98.64}} &
  \multicolumn{1}{c|}{} &
  \multicolumn{1}{c|}{\textbf{98.66}} &
  \multicolumn{1}{c|}{} &
  \multicolumn{1}{c|}{96.70} &
  \multicolumn{1}{c|}{} &
  \multicolumn{1}{c|}{99.16} &
  \multicolumn{1}{c|}{} &
  \multicolumn{1}{c|}{\textbf{98.88}} &
  \multicolumn{1}{c|}{} &
  \multicolumn{1}{c|}{97.67} &
  \multicolumn{1}{c|}{} &
  \multicolumn{1}{c|}{98.95} &
  \multicolumn{1}{c|}{} &
  \multicolumn{1}{c|}{97.97}\\ \hline 
\multicolumn{1}{|c|}{\textbf{\textit{minTrainingLoss}}} &
  \multicolumn{1}{c|}{\textbf{0.08}} &
  \multicolumn{1}{c|}{} &
  \multicolumn{1}{c|}{\textbf{0.04}} &
  \multicolumn{1}{c|}{} &
  \multicolumn{1}{c|}{\textbf{0.04}} &
  \multicolumn{1}{c|}{} &
  \multicolumn{1}{c|}{0.10} &
  \multicolumn{1}{c|}{} &
  \multicolumn{1}{c|}{0.03} &
  \multicolumn{1}{c|}{} &
  \multicolumn{1}{c|}{\textbf{0.04}} &
  \multicolumn{1}{c|}{} &
  \multicolumn{1}{c|}{0.06} &
  \multicolumn{1}{c|}{} &
  \multicolumn{1}{c|}{0.03} &
  \multicolumn{1}{c|}{} &
  \multicolumn{1}{c|}{0.06} \\ \hline
\multicolumn{1}{c|}{} &
  \multicolumn{17}{c|}{{\color[HTML]{F8A102} \textbf{0,001}}} \\ \hline
\multicolumn{1}{|c|}{\textbf{\textit{maxValidationAccuracy}}} &
  \multicolumn{1}{c|}{93.69} &
  \multicolumn{1}{c|}{} &
  \multicolumn{1}{c|}{95.09} &
  \multicolumn{1}{c|}{} &
  \multicolumn{1}{c|}{95.40} &
  \multicolumn{1}{c|}{} &
  \multicolumn{1}{c|}{\textbf{95.69}} &
  \multicolumn{1}{c|}{} &
  \multicolumn{1}{c|}{\textbf{97.12}} &
  \multicolumn{1}{c|}{} &
  \multicolumn{1}{c|}{96.02} &
  \multicolumn{1}{c|}{} &
  \multicolumn{1}{c|}{\textbf{96.31}} &
  \multicolumn{1}{c|}{} &
  \multicolumn{1}{c|}{\textbf{96.99}} &
  \multicolumn{1}{c|}{} &
  \multicolumn{1}{c|}{\textbf{95.89}} \\ \hline 
\multicolumn{1}{|c|}{\textbf{\textit{minValidationLoss}}} &
  \multicolumn{1}{c|}{0.21} &
  \multicolumn{1}{c|}{} &
  \multicolumn{1}{c|}{0.17} &
  \multicolumn{1}{c|}{} &
  \multicolumn{1}{c|}{0.20} &
  \multicolumn{1}{c|}{} &
  \multicolumn{1}{c|}{\textbf{0.19}} &
  \multicolumn{1}{c|}{} &
  \multicolumn{1}{c|}{\textbf{0.13}} &
  \multicolumn{1}{c|}{} &
  \multicolumn{1}{c|}{0.16} &
  \multicolumn{1}{c|}{} &
  \multicolumn{1}{c|}{\textbf{0.16}} &
  \multicolumn{1}{c|}{} &
  \multicolumn{1}{c|}{\textbf{0.14}} &
  \multicolumn{1}{c|}{} &
  \multicolumn{1}{c|}{\textbf{0.17}} \\ \hline  
\multicolumn{1}{|c|}{\textbf{\textit{maxTrainingAccuracy}}} &
  \multicolumn{1}{c|}{95.10} &
  \multicolumn{1}{c|}{} &
  \multicolumn{1}{c|}{98.01} &
  \multicolumn{1}{c|}{} &
  \multicolumn{1}{c|}{98.85} &
  \multicolumn{1}{c|}{} &
  \multicolumn{1}{c|}{\textbf{98.22}} &
  \multicolumn{1}{c|}{} &
  \multicolumn{1}{c|}{\textbf{99.08}} &
  \multicolumn{1}{c|}{} &
  \multicolumn{1}{c|}{99.40} &
  \multicolumn{1}{c|}{} &
  \multicolumn{1}{c|}{\textbf{98.23}} &
  \multicolumn{1}{c|}{} &
  \multicolumn{1}{c|}{\textbf{98.73}} &
  \multicolumn{1}{c|}{} &
  \multicolumn{1}{c|}{\textbf{99.23}} \\ \hline 
\multicolumn{1}{|c|}{\textbf{\textit{minTrainingLoss}}} &
  \multicolumn{1}{c|}{0.15} &
  \multicolumn{1}{c|}{} &
  \multicolumn{1}{c|}{0.06} &
  \multicolumn{1}{c|}{} &
  \multicolumn{1}{c|}{0.06} &
  \multicolumn{1}{c|}{} &
  \multicolumn{1}{c|}{\textbf{0.06}} &
  \multicolumn{1}{c|}{} &
  \multicolumn{1}{c|}{\textbf{0.03}} &
  \multicolumn{1}{c|}{} &
  \multicolumn{1}{c|}{0.02} &
  \multicolumn{1}{c|}{} &
  \multicolumn{1}{c|}{\textbf{0.05}} &
  \multicolumn{1}{c|}{} &
  \multicolumn{1}{c|}{\textbf{0.04}} &
  \multicolumn{1}{c|}{} &
  \multicolumn{1}{c|}{\textbf{0.03}} \\ \hline
  \multicolumn{1}{c|}{} &
  \multicolumn{17}{c|}{{\color[HTML]{F8A102} \textbf{0,0001}}} \\ \hline
\multicolumn{1}{|c|}{\textbf{\textit{maxValidationAccuracy}}} &
  \multicolumn{1}{c|}{86.19} &
  \multicolumn{1}{c|}{} &
  \multicolumn{1}{c|}{79.98} &
  \multicolumn{1}{c|}{} &
  \multicolumn{1}{c|}{90.27} &
  \multicolumn{1}{c|}{} &
  \multicolumn{1}{c|}{86.88} &
  \multicolumn{1}{c|}{} &
  \multicolumn{1}{c|}{92.40} &
  \multicolumn{1}{c|}{} &
  \multicolumn{1}{c|}{92.04} &
  \multicolumn{1}{c|}{} &
  \multicolumn{1}{c|}{93.69} &
  \multicolumn{1}{c|}{} &
  \multicolumn{1}{c|}{95.29} &
  \multicolumn{1}{c|}{} &
  \multicolumn{1}{c|}{94.85} \\ \hline 
\multicolumn{1}{|c|}{\textbf{\textit{minValidationLoss}}} &
  \multicolumn{1}{c|}{0.43} &
  \multicolumn{1}{c|}{} &
  \multicolumn{1}{c|}{0.67} &
  \multicolumn{1}{c|}{} &
  \multicolumn{1}{c|}{0.37} &
  \multicolumn{1}{c|}{} &
  \multicolumn{1}{c|}{0.37} &
  \multicolumn{1}{c|}{} &
  \multicolumn{1}{c|}{0.27} &
  \multicolumn{1}{c|}{} &
  \multicolumn{1}{c|}{0.27} &
  \multicolumn{1}{c|}{} &
  \multicolumn{1}{c|}{0.26} &
  \multicolumn{1}{c|}{} &
  \multicolumn{1}{c|}{0.19} &
  \multicolumn{1}{c|}{} &
  \multicolumn{1}{c|}{0.21} \\ \hline 
\multicolumn{1}{|c|}{\textbf{\textit{maxTrainingAccuracy}}} &
  \multicolumn{1}{c|}{88.37} &
  \multicolumn{1}{c|}{} &
  \multicolumn{1}{c|}{84.54} &
  \multicolumn{1}{c|}{} &
  \multicolumn{1}{c|}{93.85} &
  \multicolumn{1}{c|}{} &
  \multicolumn{1}{c|}{89.00} &
  \multicolumn{1}{c|}{} &
  \multicolumn{1}{c|}{94.93} &
  \multicolumn{1}{c|}{} &
  \multicolumn{1}{c|}{97.20} &
  \multicolumn{1}{c|}{} &
  \multicolumn{1}{c|}{94.51} &
  \multicolumn{1}{c|}{} &
  \multicolumn{1}{c|}{97.41} &
  \multicolumn{1}{c|}{} &
  \multicolumn{1}{c|}{98.96} \\ \hline   
\multicolumn{1}{|c|}{\textbf{\textit{minTrainingLoss}}} &
  \multicolumn{1}{c|}{0.39} &
  \multicolumn{1}{c|}{} &
  \multicolumn{1}{c|}{0.48} &
  \multicolumn{1}{c|}{} &
  \multicolumn{1}{c|}{0.25} &
  \multicolumn{1}{c|}{} &
  \multicolumn{1}{c|}{0.28} &
  \multicolumn{1}{c|}{} &
  \multicolumn{1}{c|}{0.17} &
  \multicolumn{1}{c|}{} &
  \multicolumn{1}{c|}{0.11} &
  \multicolumn{1}{c|}{} &
  \multicolumn{1}{c|}{0.14} &
  \multicolumn{1}{c|}{} &
  \multicolumn{1}{c|}{0.08} &
  \multicolumn{1}{c|}{} &
  \multicolumn{1}{c|}{0.04} \\ \hline
\end{tabular}}
\caption{Hyperparameter analysis for the following configurations: patch size, 64; initial number of filters, 8, 16, 32; encoder depth, 2, 3, 4 and initial learning rate, 0.1, 0.01, 0.001 and 0.0001.}
\label{tab:resultadosJavi1_64}
\end{center}
\end{table}

\begin{table}[h!]
\begin{center}
%\rotatebox{90}{
\scalebox{0.85}{
\begin{tabular}{ccccccccccccccccccc}
\cline{1-6} \cline{8-12} \cline{14-18}
\multicolumn{1}{|c|}{\textbf{Initial num. of filt.}} &
  \multicolumn{5}{c|}{{\color[HTML]{6200C9} \textbf{8}}} &
  \multicolumn{1}{c|}{} &
  \multicolumn{5}{c|}{{\color[HTML]{6200C9} \textbf{16}}} &
  \multicolumn{1}{c|}{} &
  \multicolumn{5}{c|}{{\color[HTML]{6200C9} \textbf{32}}} \\ \hline
\multicolumn{1}{|c|}{\diagbox[]{\textbf{Img. size}}{\textbf{Enc. depth}}} &
  \multicolumn{1}{c|}{\color[HTML]{009901} \textbf{2}} &
  \multicolumn{1}{c|}{} &
  \multicolumn{1}{c|}{\color[HTML]{442605} \textbf{3}} &
  \multicolumn{1}{c|}{} &
  \multicolumn{1}{c|}{\color[HTML]{0FC6D8} \textbf{4}} &
  \multicolumn{1}{c|}{} &
  \multicolumn{1}{c|}{{\color[HTML]{009901} \textbf{2}}} &
  \multicolumn{1}{c|}{} &
  \multicolumn{1}{c|}{\color[HTML]{442605} \textbf{3}} &
  \multicolumn{1}{c|}{} &
  \multicolumn{1}{c|}{\color[HTML]{0FC6D8} \textbf{4}} &
  \multicolumn{1}{c|}{} &
  \multicolumn{1}{c|}{{\color[HTML]{009901} \textbf{2}}} &
  \multicolumn{1}{c|}{} &
  \multicolumn{1}{c|}{\color[HTML]{442605} \textbf{3}} &
  \multicolumn{1}{c|}{} &
  \multicolumn{1}{c|}{\color[HTML]{0FC6D8} \textbf{4}} \\ \hline
\multicolumn{1}{|c|}{{\color[HTML]{FE0000} \textbf{128}}} &
  \multicolumn{17}{c|}{{\color[HTML]{F8A102} \textbf{0,1}}} \\ \hline
\multicolumn{1}{|c|}{\textbf{\textit{maxValidationAccuracy}}} &
  \multicolumn{1}{c|}{94.31} &
  \multicolumn{1}{c|}{} &
  \multicolumn{1}{c|}{91.65} &
  \multicolumn{1}{c|}{} &
  \multicolumn{1}{c|}{96.41} &
  \multicolumn{1}{c|}{} &
  \multicolumn{1}{c|}{94.10} &
  \multicolumn{1}{c|}{} &
  \multicolumn{1}{c|}{96.37} &
  \multicolumn{1}{c|}{} &
  \multicolumn{1}{c|}{93.60} &
  \multicolumn{1}{c|}{} &
  \multicolumn{1}{c|}{95.48} &
  \multicolumn{1}{c|}{} &
  \multicolumn{1}{c|}{96.23} &
  \multicolumn{1}{c|}{} &
  \multicolumn{1}{c|}{96.32} \\ \hline
\multicolumn{1}{|c|}{\textbf{\textit{minValidationLoss}}} &
  \multicolumn{1}{c|}{0.17} &
  \multicolumn{1}{c|}{} &
  \multicolumn{1}{c|}{0.22} &
  \multicolumn{1}{c|}{} &
  \multicolumn{1}{c|}{0.15} &
  \multicolumn{1}{c|}{} &
  \multicolumn{1}{c|}{0.18} &
  \multicolumn{1}{c|}{} &
  \multicolumn{1}{c|}{0.12} &
  \multicolumn{1}{c|}{} &
  \multicolumn{1}{c|}{0.20} &
  \multicolumn{1}{c|}{} &
  \multicolumn{1}{c|}{0.15} &
  \multicolumn{1}{c|}{} &
  \multicolumn{1}{c|}{0.14} &
  \multicolumn{1}{c|}{} &
  \multicolumn{1}{c|}{0.14} \\ \hline 
\multicolumn{1}{|c|}{\textbf{\textit{maxTrainingAccuracy}}} &
  \multicolumn{1}{c|}{96.91} &
  \multicolumn{1}{c|}{} &
  \multicolumn{1}{c|}{95.11} &
  \multicolumn{1}{c|}{} &
  \multicolumn{1}{c|}{98.56} &
  \multicolumn{1}{c|}{} &
  \multicolumn{1}{c|}{96.91} &
  \multicolumn{1}{c|}{} &
  \multicolumn{1}{c|}{98.89} &
  \multicolumn{1}{c|}{} &
  \multicolumn{1}{c|}{95.98} &
  \multicolumn{1}{c|}{} &
  \multicolumn{1}{c|}{97.74} &
  \multicolumn{1}{c|}{} &
  \multicolumn{1}{c|}{98.46} &
  \multicolumn{1}{c|}{} &
  \multicolumn{1}{c|}{98.33} \\ \hline
\multicolumn{1}{|c|}{\textbf{\textit{minTrainingLoss}}} &
  \multicolumn{1}{c|}{0.09} &
  \multicolumn{1}{c|}{} &
  \multicolumn{1}{c|}{0.14} &
  \multicolumn{1}{c|}{} &
  \multicolumn{1}{c|}{0.04} &
  \multicolumn{1}{c|}{} &
  \multicolumn{1}{c|}{0.08} &
  \multicolumn{1}{c|}{} &
  \multicolumn{1}{c|}{0.03} &
  \multicolumn{1}{c|}{} &
  \multicolumn{1}{c|}{0.12} &
  \multicolumn{1}{c|}{} &
  \multicolumn{1}{c|}{0.07} &
  \multicolumn{1}{c|}{} &
  \multicolumn{1}{c|}{0.04} &
  \multicolumn{1}{c|}{} &
  \multicolumn{1}{c|}{0.05} \\ \hline
  \multicolumn{1}{c|}{} &
  \multicolumn{17}{c|}{{\color[HTML]{F8A102} \textbf{0,01}}} \\ \hline
\multicolumn{1}{|c|}{\textbf{\textit{maxValidationAccuracy}}} &
  \multicolumn{1}{c|}{\textbf{95.75}} &
  \multicolumn{1}{c|}{} &
  \multicolumn{1}{c|}{\textbf{97.25}} &
  \multicolumn{1}{c|}{} &
  \multicolumn{1}{c|}{\textbf{97.78}} &
  \multicolumn{1}{c|}{} &
  \multicolumn{1}{c|}{96.05} &
  \multicolumn{1}{c|}{} &
  \multicolumn{1}{c|}{\textbf{96.64}} &
  \multicolumn{1}{c|}{} &
  \multicolumn{1}{c|}{\textbf{97.43}} &
  \multicolumn{1}{c|}{} &
  \multicolumn{1}{c|}{95.06} &
  \multicolumn{1}{c|}{} &
  \multicolumn{1}{c|}{97.23} &
  \multicolumn{1}{c|}{} &
  \multicolumn{1}{c|}{97.26} \\ \hline 
\multicolumn{1}{|c|}{\textbf{\textit{minValidationLoss}}} &
  \multicolumn{1}{c|}{\textbf{0.17}} &
  \multicolumn{1}{c|}{} &
  \multicolumn{1}{c|}{\textbf{0.13}} &
  \multicolumn{1}{c|}{} &
  \multicolumn{1}{c|}{\textbf{0.11}} &
  \multicolumn{1}{c|}{} &
  \multicolumn{1}{c|}{0.16} &
  \multicolumn{1}{c|}{} &
  \multicolumn{1}{c|}{\textbf{0.16}} &
  \multicolumn{1}{c|}{} &
  \multicolumn{1}{c|}{\textbf{0.11}} &
  \multicolumn{1}{c|}{} &
  \multicolumn{1}{c|}{0.19} &
  \multicolumn{1}{c|}{} &
  \multicolumn{1}{c|}{0.13} &
  \multicolumn{1}{c|}{} &
  \multicolumn{1}{c|}{0.12} \\ \hline  
\multicolumn{1}{|c|}{\textbf{\textit{maxTrainingAccuracy}}} &
  \multicolumn{1}{c|}{\textbf{98.52}} &
  \multicolumn{1}{c|}{} &
  \multicolumn{1}{c|}{\textbf{99.49}} &
  \multicolumn{1}{c|}{} &
  \multicolumn{1}{c|}{\textbf{99.65}} &
  \multicolumn{1}{c|}{} &
  \multicolumn{1}{c|}{98.90} &
  \multicolumn{1}{c|}{} &
  \multicolumn{1}{c|}{\textbf{99.49}} &
  \multicolumn{1}{c|}{} &
  \multicolumn{1}{c|}{\textbf{99.62}} &
  \multicolumn{1}{c|}{} &
  \multicolumn{1}{c|}{98.65} &
  \multicolumn{1}{c|}{} &
  \multicolumn{1}{c|}{99.67} &
  \multicolumn{1}{c|}{} &
  \multicolumn{1}{c|}{99.68} \\ \hline 
\multicolumn{1}{|c|}{\textbf{\textit{minTrainingLoss}}} &
  \multicolumn{1}{c|}{\textbf{0.04}} &
  \multicolumn{1}{c|}{} &
  \multicolumn{1}{c|}{\textbf{0.01}} &
  \multicolumn{1}{c|}{} &
  \multicolumn{1}{c|}{\textbf{0.01}} &
  \multicolumn{1}{c|}{} &
  \multicolumn{1}{c|}{0.03} &
  \multicolumn{1}{c|}{} &
  \multicolumn{1}{c|}{\textbf{0.01}} &
  \multicolumn{1}{c|}{} &
  \multicolumn{1}{c|}{\textbf{0.01}} &
  \multicolumn{1}{c|}{} &
  \multicolumn{1}{c|}{0.04} &
  \multicolumn{1}{c|}{} &
  \multicolumn{1}{c|}{0.01} &
  \multicolumn{1}{c|}{} &
  \multicolumn{1}{c|}{0.01} \\ \hline
  \multicolumn{1}{c|}{} &
  \multicolumn{17}{c|}{{\color[HTML]{F8A102} \textbf{0,001}}} \\ \hline
\multicolumn{1}{|c|}{\textbf{\textit{maxValidationAccuracy}}} &
  \multicolumn{1}{c|}{95.56} &
  \multicolumn{1}{c|}{} &
  \multicolumn{1}{c|}{97.01} &
  \multicolumn{1}{c|}{} &
  \multicolumn{1}{c|}{96.97} &
  \multicolumn{1}{c|}{} &
  \multicolumn{1}{c|}{\textbf{96.38}} &
  \multicolumn{1}{c|}{} &
  \multicolumn{1}{c|}{96.57} &
  \multicolumn{1}{c|}{} &
  \multicolumn{1}{c|}{97.02} &
  \multicolumn{1}{c|}{} &
  \multicolumn{1}{c|}{\textbf{96.73}} &
  \multicolumn{1}{c|}{} &
  \multicolumn{1}{c|}{\textbf{97.24}} &
  \multicolumn{1}{c|}{} &
  \multicolumn{1}{c|}{\textbf{97.31}} \\ \hline 
\multicolumn{1}{|c|}{\textbf{\textit{minValidationLoss}}} &
  \multicolumn{1}{c|}{0.17} &
  \multicolumn{1}{c|}{} &
  \multicolumn{1}{c|}{0.16} &
  \multicolumn{1}{c|}{} &
  \multicolumn{1}{c|}{0.17} &
  \multicolumn{1}{c|}{} &
  \multicolumn{1}{c|}{\textbf{0.16}} &
  \multicolumn{1}{c|}{} &
  \multicolumn{1}{c|}{0.17} &
  \multicolumn{1}{c|}{} &
  \multicolumn{1}{c|}{0.13} &
  \multicolumn{1}{c|}{} &
  \multicolumn{1}{c|}{\textbf{0.16}} &
  \multicolumn{1}{c|}{} &
  \multicolumn{1}{c|}{\textbf{0.15}} &
  \multicolumn{1}{c|}{} &
  \multicolumn{1}{c|}{\textbf{0.11}} \\ \hline 
\multicolumn{1}{|c|}{\textbf{\textit{maxTrainingAccuracy}}} &
  \multicolumn{1}{c|}{97.02} &
  \multicolumn{1}{c|}{} &
  \multicolumn{1}{c|}{99.34} &
  \multicolumn{1}{c|}{} &
  \multicolumn{1}{c|}{99.72} &
  \multicolumn{1}{c|}{} &
  \multicolumn{1}{c|}{\textbf{98.63}} &
  \multicolumn{1}{c|}{} &
  \multicolumn{1}{c|}{99.62} &
  \multicolumn{1}{c|}{} &
  \multicolumn{1}{c|}{99.77} &
  \multicolumn{1}{c|}{} &
  \multicolumn{1}{c|}{\textbf{99.26}} &
  \multicolumn{1}{c|}{} &
  \multicolumn{1}{c|}{\textbf{99.83}} &
  \multicolumn{1}{c|}{} &
  \multicolumn{1}{c|}{\textbf{99.85}} \\ \hline   
\multicolumn{1}{|c|}{\textbf{\textit{minTrainingLoss}}} &
  \multicolumn{1}{c|}{0.08} &
  \multicolumn{1}{c|}{} &
  \multicolumn{1}{c|}{0.03} &
  \multicolumn{1}{c|}{} &
  \multicolumn{1}{c|}{0.02} &
  \multicolumn{1}{c|}{} &
  \multicolumn{1}{c|}{\textbf{0.04}} &
  \multicolumn{1}{c|}{} &
  \multicolumn{1}{c|}{0.01} &
  \multicolumn{1}{c|}{} &
  \multicolumn{1}{c|}{0.01} &
  \multicolumn{1}{c|}{} &
  \multicolumn{1}{c|}{\textbf{0.02}} &
  \multicolumn{1}{c|}{} &
  \multicolumn{1}{c|}{\textbf{0.01}} &
  \multicolumn{1}{c|}{} &
  \multicolumn{1}{c|}{\textbf{0.01}} \\ \hline
  \multicolumn{1}{c|}{} &
  \multicolumn{17}{c|}{{\color[HTML]{F8A102} \textbf{0,0001}}} \\ \hline
\multicolumn{1}{|c|}{\textbf{\textit{maxValidationAccuracy}}} &
  \multicolumn{1}{c|}{73.08} &
  \multicolumn{1}{c|}{} &
  \multicolumn{1}{c|}{91.65} &
  \multicolumn{1}{c|}{} &
  \multicolumn{1}{c|}{93.33} &
  \multicolumn{1}{c|}{} &
  \multicolumn{1}{c|}{93.41} &
  \multicolumn{1}{c|}{} &
  \multicolumn{1}{c|}{94.96} &
  \multicolumn{1}{c|}{} &
  \multicolumn{1}{c|}{94.23} &
  \multicolumn{1}{c|}{} &
  \multicolumn{1}{c|}{93.88} &
  \multicolumn{1}{c|}{} &
  \multicolumn{1}{c|}{95.56} &
  \multicolumn{1}{c|}{} &
  \multicolumn{1}{c|}{96.19} \\ \hline  
\multicolumn{1}{|c|}{\textbf{\textit{minValidationLoss}}} &
  \multicolumn{1}{c|}{0.64} &
  \multicolumn{1}{c|}{} &
  \multicolumn{1}{c|}{0.31} &
  \multicolumn{1}{c|}{} &
  \multicolumn{1}{c|}{0.25} &
  \multicolumn{1}{c|}{} &
  \multicolumn{1}{c|}{0.24} &
  \multicolumn{1}{c|}{} &
  \multicolumn{1}{c|}{0.24} &
  \multicolumn{1}{c|}{} &
  \multicolumn{1}{c|}{0.21} &
  \multicolumn{1}{c|}{} &
  \multicolumn{1}{c|}{0.29} &
  \multicolumn{1}{c|}{} &
  \multicolumn{1}{c|}{0.20} &
  \multicolumn{1}{c|}{} &
  \multicolumn{1}{c|}{0.19} \\ \hline 
\multicolumn{1}{|c|}{\textbf{\textit{maxTrainingAccuracy}}} &
  \multicolumn{1}{c|}{81.81} &
  \multicolumn{1}{c|}{} &
  \multicolumn{1}{c|}{94.12} &
  \multicolumn{1}{c|}{} &
  \multicolumn{1}{c|}{97.20} &
  \multicolumn{1}{c|}{} &
  \multicolumn{1}{c|}{95.28} &
  \multicolumn{1}{c|}{} &
  \multicolumn{1}{c|}{97.95} &
  \multicolumn{1}{c|}{} &
  \multicolumn{1}{c|}{99.07} &
  \multicolumn{1}{c|}{} &
  \multicolumn{1}{c|}{97.19} &
  \multicolumn{1}{c|}{} &
  \multicolumn{1}{c|}{99.10} &
  \multicolumn{1}{c|}{} &
  \multicolumn{1}{c|}{99.77} \\ \hline 
\multicolumn{1}{|c|}{\textbf{\textit{minTrainingLoss}}} &
  \multicolumn{1}{c|}{0.50} &
  \multicolumn{1}{c|}{} &
  \multicolumn{1}{c|}{0.31} &
  \multicolumn{1}{c|}{} &
  \multicolumn{1}{c|}{0.10} &
  \multicolumn{1}{c|}{} &
  \multicolumn{1}{c|}{0.16} &
  \multicolumn{1}{c|}{} &
  \multicolumn{1}{c|}{0.08} &
  \multicolumn{1}{c|}{} &
  \multicolumn{1}{c|}{0.03} &
  \multicolumn{1}{c|}{} &
  \multicolumn{1}{c|}{0.09} &
  \multicolumn{1}{c|}{} &
  \multicolumn{1}{c|}{0.02} &
  \multicolumn{1}{c|}{} &
  \multicolumn{1}{c|}{0.01} \\ \hline
\end{tabular}}
\caption{Hyperparameter analysis for the following configurations: patch size, 128; initial number of filters, 8, 16, 32; encoder depth, 2,3,4 and initial learning rate, 0.1, 0.01, 0.001 and 0.0001.}
\label{tab:resultadosJavi1_128}
\end{center}
\end{table}

\newpage

\begin{equation}
NP = \evalat{f(ed, if, cks, ucks, c, ic)}{cks = 3, ucks = 2, c = 3/5, ic = 25}
\label{equ:n_params}
\end{equation}

\begin{equation}
FLOPS = \evalat{f(ed, if, cks, ucks, c, ic, id)}{cks = 3, ucks = 2, c = 3/5, ic = 25, id = 128}
\label{equ:n_FLOPS}
\end{equation}

\begin{figure}[h!]
\centering
  \begin{minipage}{0.45\textwidth}
    \centering
    \includegraphics[width=1.0\textwidth]{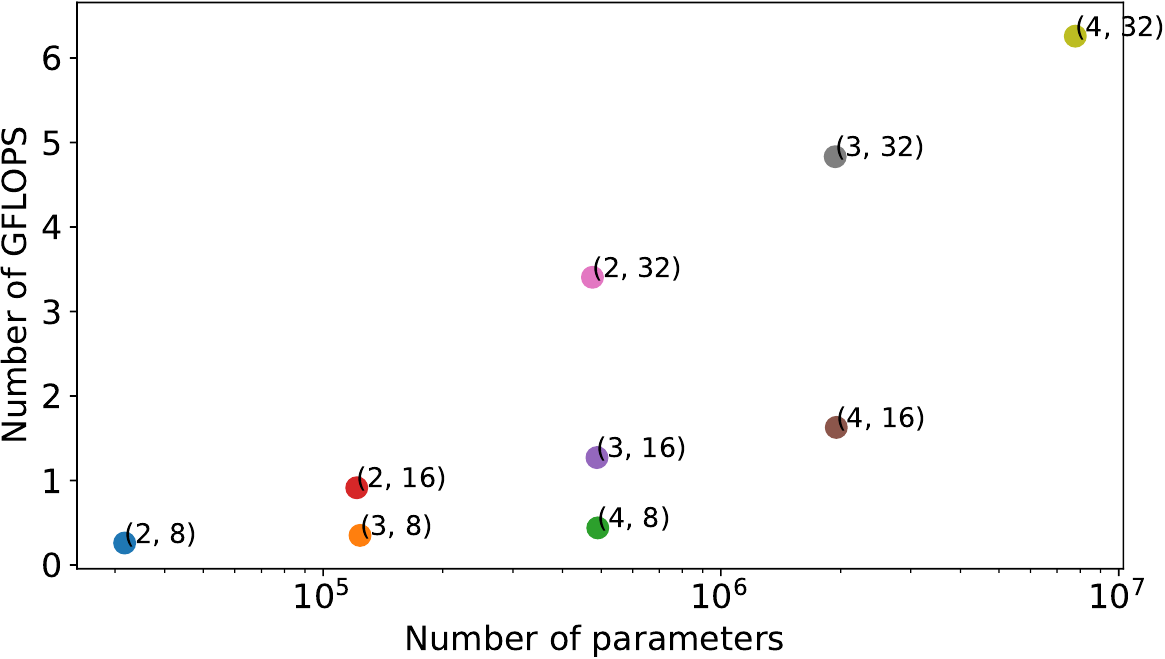}
    \caption{Variation in the number of FLOPS and parameters as a function of ($ed$, $if$).}
    \label{fig:paramsVSflops}
    \end{minipage}\hfill
    \begin{minipage}{0.45\textwidth}
    \centering
    \includegraphics[width=1.0\textwidth]{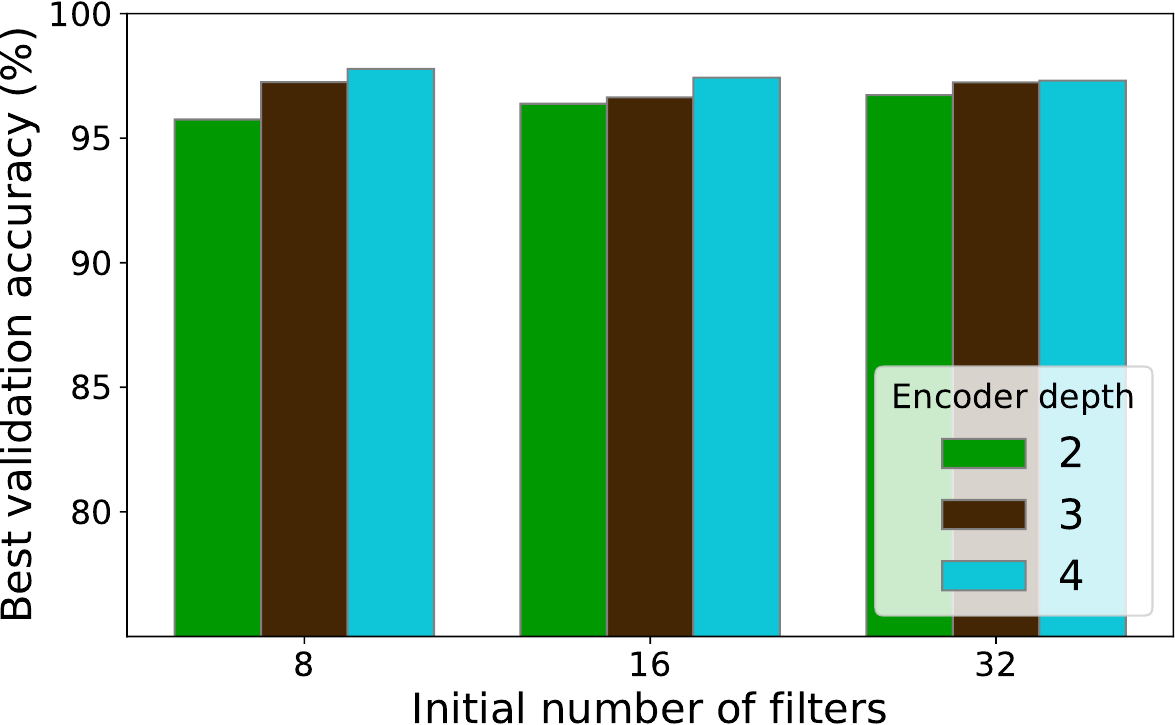}
    \caption{Best validation accuracy in terms of encoder depth and initial number of filters.}
\label{fig:valAcc}
    \end{minipage}
\end{figure}

Figure \ref{fig:paramsVSflops} shows the evaluation of the NP and FLOPS in the range of the hyperparameters under analysis. It can be seen, on the one hand, that the NP is almost quadrupled as we increase either the encoder depth or the initial number of filters. On the other hand, the number of FLOPS significantly increases with the initial number of filters, and not quite with the growth of the encoder depth. 

Figure \ref{fig:valAcc} depicts the highest achieved validation accuracy for each pair ($ed$, $if$) after training. Increasing $ed$ slightly improves the validation accuracy regardless of the initial number of filters.

All in all, combining the information extracted from both graphics, we selected the combination ($id$, $ed$, $if$) = (128, 2, 8) as the architecture with the best trade-off between model complexity and classification performance. Not only does it perform very similarly to the other architectures in terms of segmentation, but choosing a higher number of initial filters would considerably increase the memory footprint and the number of FLOPS (without improving precision/recall), and increasing the model depth would significantly increase the ratio of memory accesses with respect to the number of FLOPS (overshadowing the minor increase in precision/recall).

\begin{table}[h!]
\centering
\resizebox{7cm}{!}{%
\begin{tabular}{c|c|c|}
\cline{2-3}
                                                                   & \textbf{U-Net} & \textbf{ANN} \\ \hline
\multicolumn{1}{|c|}{\textbf{Num. of params.}}                     &      31.707 K          &  13.653 K    \\ \hline
\multicolumn{1}{|c|}{\textbf{FLOPS}}                               &      259.9 M         &  27.093 K    \\ \hline
\multicolumn{1}{|c|}{\textbf{FLOPS per image inference}}           &      4.678 G          &  1.206 G     \\ \hline
\end{tabular}}
\caption{Comparison of U-Net and ANN with regard to model size and computational complexity}
\label{tab:FLOPSparamsUNetANN}
\end{table}

We can now compare the U-Net and the baseline ANN in terms of model size and computational complexity (MACS, Multiply and Accumulate operations or FLOPS) as it is reflected on Table \ref{tab:FLOPSparamsUNetANN}: while the ANN has only 13,653 parameters and performs 1,206,000,000 FLOPS (27,093 FLOPS per pixel) during inference, U-Net model has 31,707 parameters (320 of them are non-trainable) and needs 4,678,200,00 FLOPS (259,900,000 FLOPS per patch) to produce an output. The difference between the FLOPS ratio (3.879x) and the parameters ratio (2.3x) affects the time needed to make the forward pass, which will be assessed in the next section.

\subsection{Segmentation Results}\label{sec:segRes}
Having first selected the best model architecture, we have trained U-Net on the complete training set. The training has been performed on a NVIDIA GeForce RTX 3090-24GB GPU with an Adam optimizer (0.9 gradient decay factor and 0.999 squared gradient decay factor), a mini batch size of 128, an initial learning rate of 0.005 with drop period and weighted cross-entropy loss function, among other parameters, for, at most, 60 epochs. Table \ref{tab:metricsFloatingPointUnet} collects the classification performance on the patches in the testing set. In addition, as previously stated, the images are reconstructed to the original resolution from the 18 overlapping patches into which each of them has been divided, so Table \ref{tab:metricsFloatingPointUnet} also displays the metrics for the complete reconstructed images. The comparison shows how the use of overlapping patches improves the segmentation performance, especially the precision, compared to the non overlapping reconstruction. This is because the FCNs tend to fail to correctly predict the classes on the patch contours since they lack part of the contextual information for the contour pixels.

\begin{table}[h!]
\centering
\resizebox{10cm}{!}{%
\begin{tabular}{ccccccc}
\cline{2-7}
\multicolumn{1}{c|}{} & \multicolumn{3}{c|}{\textbf{\begin{tabular}[c]{@{}c@{}}Patches\\ (128x128x25)\end{tabular}}} & \multicolumn{3}{c|}{\textbf{\begin{tabular}[c]{@{}c@{}}Rebuilt images from\\ overlapping patches\end{tabular}}} \\ \cline{2-7} 
\multicolumn{1}{c|}{} & \multicolumn{1}{c|}{\textbf{Recall}} & \multicolumn{1}{c|}{\textbf{Precision}} & \multicolumn{1}{c|}{\textbf{IoU}} & \multicolumn{1}{c|}{\textbf{Recall}} & \multicolumn{1}{c|}{\textbf{Precision}} & \multicolumn{1}{c|}{\textbf{IoU}} \\ \hline
\multicolumn{1}{|c|}{\textbf{Road}} & \multicolumn{1}{c|}{97.90} & \multicolumn{1}{c|}{95.66} & \multicolumn{1}{c|}{93.74} & \multicolumn{1}{c|}{98.54} & \multicolumn{1}{c|}{94.56} & \multicolumn{1}{c|}{93.25} \\ \hline
\multicolumn{1}{|c|}{\textbf{Road Marks}} & \multicolumn{1}{c|}{90.25} & \multicolumn{1}{c|}{73.11} & \multicolumn{1}{c|}{67.75} & \multicolumn{1}{c|}{87.89} & \multicolumn{1}{c|}{77.22} & \multicolumn{1}{c|}{\textbf{69.80}} \\ \hline
\multicolumn{1}{|c|}{\textbf{Non-Drivable}} & \multicolumn{1}{c|}{91.07} & \multicolumn{1}{c|}{97.16} & \multicolumn{1}{c|}{88.71} & \multicolumn{1}{c|}{91.20} & \multicolumn{1}{c|}{98.57} & \multicolumn{1}{c|}{\textbf{90.01}} \\ \hline
\multicolumn{1}{|c|}{\textbf{Overall}} & \multicolumn{1}{c|}{95.37} & \multicolumn{1}{c|}{95.55} & \multicolumn{1}{c|}{91.31} & \multicolumn{1}{c|}{95.42} & \multicolumn{1}{c|}{95.44} & \multicolumn{1}{c|}{91.50} \\ \hline
\multicolumn{1}{|c|}{\textbf{Mean}} & \multicolumn{1}{c|}{93.07} & \multicolumn{1}{c|}{88.64} & \multicolumn{1}{c|}{83.40} & \multicolumn{1}{c|}{92.54} & \multicolumn{1}{c|}{90.12} & \multicolumn{1}{c|}{84.35} \\ \hline
\multicolumn{1}{|c|}{\textbf{Weighted}} & \multicolumn{1}{c|}{90.60} & \multicolumn{1}{c|}{75.71} & \multicolumn{1}{c|}{70.27} & \multicolumn{1}{c|}{88.54} & \multicolumn{1}{c|}{79.43} & \multicolumn{1}{c|}{\textbf{72.60}} \\ \hline
 &  &  &  &  &  & \\ \hline
\multicolumn{1}{|c|}{\textbf{Road}} & \multicolumn{1}{c|}{92.61} & \multicolumn{1}{c|}{99.05} & \multicolumn{1}{c|}{91.36} & \multicolumn{1}{c|}{93.28} & \multicolumn{1}{c|}{99.00} & \multicolumn{1}{c|}{\textbf{92.41}} \\ \hline
\multicolumn{1}{|c|}{\textbf{Road Marks}} & \multicolumn{1}{c|}{80.93} & \multicolumn{1}{c|}{75.39} & \multicolumn{1}{c|}{64.02} & \multicolumn{1}{c|}{78.32} & \multicolumn{1}{c|}{79.11} & \multicolumn{1}{c|}{64.90} \\ \hline
\multicolumn{1}{|c|}{\textbf{Vegetation}} & \multicolumn{1}{c|}{94.98} & \multicolumn{1}{c|}{94.63} & \multicolumn{1}{c|}{90.12} & \multicolumn{1}{c|}{95.74} & \multicolumn{1}{c|}{95.80} & \multicolumn{1}{c|}{\textbf{91.88}} \\ \hline
\multicolumn{1}{|c|}{\textbf{Sky}} & \multicolumn{1}{c|}{97.86} & \multicolumn{1}{c|}{93.09} & \multicolumn{1}{c|}{91.23} & \multicolumn{1}{c|}{97.49} & \multicolumn{1}{c|}{93.39} & \multicolumn{1}{c|}{91.20} \\ \hline
\multicolumn{1}{|c|}{\textbf{Other}} & \multicolumn{1}{c|}{84.97} & \multicolumn{1}{c|}{62.71} & \multicolumn{1}{c|}{56.45} & \multicolumn{1}{c|}{84.83} & \multicolumn{1}{c|}{64.59} & \multicolumn{1}{c|}{\textbf{57.90}} \\ \hline
\multicolumn{1}{|c|}{\textbf{Overall}} & \multicolumn{1}{c|}{91.79} & \multicolumn{1}{c|}{93.29} & \multicolumn{1}{c|}{86.47} & \multicolumn{1}{c|}{92.75} & \multicolumn{1}{c|}{93.80} & \multicolumn{1}{c|}{\textbf{87.66}} \\ \hline
\multicolumn{1}{|c|}{\textbf{Mean}} & \multicolumn{1}{c|}{90.18} & \multicolumn{1}{c|}{84.98} & \multicolumn{1}{c|}{78.64} & \multicolumn{1}{c|}{89.93} & \multicolumn{1}{c|}{86.38} & \multicolumn{1}{c|}{\textbf{79.66}} \\ \hline
\multicolumn{1}{|c|}{\textbf{Weighted}} & \multicolumn{1}{c|}{88.64} & \multicolumn{1}{c|}{82.14} & \multicolumn{1}{c|}{75.27} & \multicolumn{1}{c|}{87.40} & \multicolumn{1}{c|}{84.15} & \multicolumn{1}{c|}{75.93} \\ \hline
\end{tabular}}
\caption{Performance of the modified U-Net (patches and rebuilt images from overlapping patches) on the 3-classes (up) and 5-classes (down) test datasets in terms of recall, accuracy and IoU (Intersection over Union).}
\label{tab:metricsFloatingPointUnet}
\end{table}

\begin{figure}[h!]
\begin{subfigure}{0.32\linewidth}
\centering
\includegraphics[width=5.4cm]{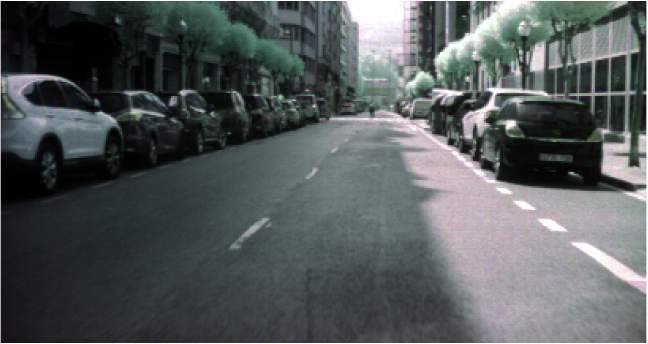}
\caption{Urban, visible.}
\label{fig:visUrban3clases}
\end{subfigure}
\begin{subfigure}{0.32\linewidth}
\centering
\includegraphics[width=5.4cm]{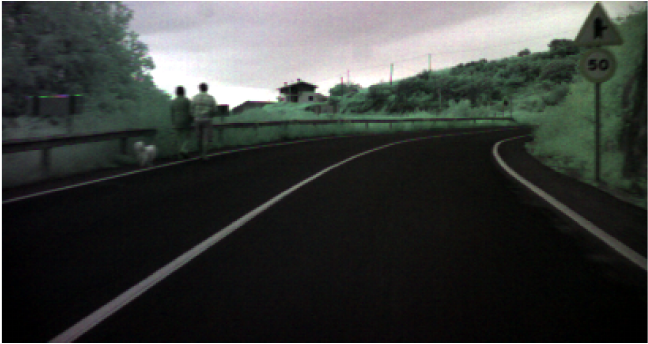}
\caption{Road, visible.}
\label{fig:visRoad3clases}
\end{subfigure}
\begin{subfigure}{0.32\linewidth}
\centering
\includegraphics[width=5.4cm]{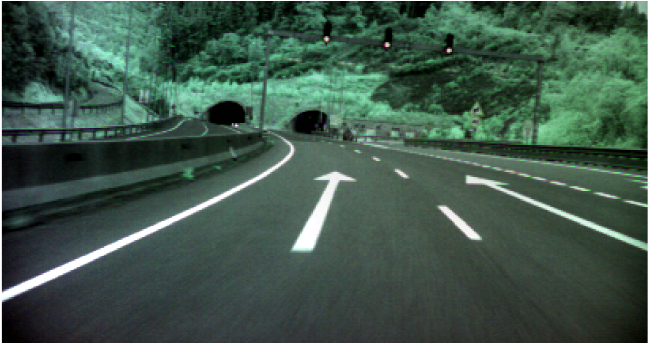}
\caption{Highway, visible.}
\label{fig:visHighway3clases}
\end{subfigure}\\[1ex]
\begin{subfigure}{0.32\linewidth}
\centering
\includegraphics[width=5.35cm]{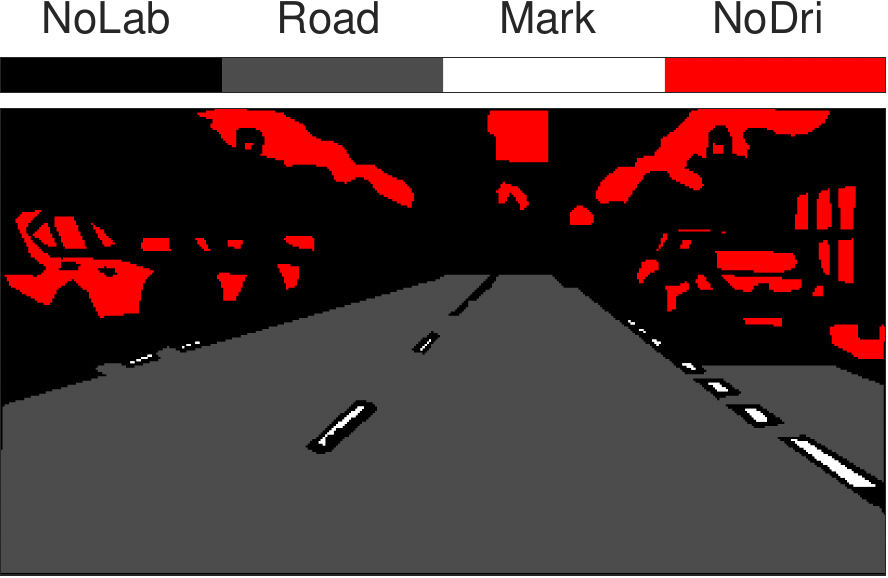}
\caption{Urban, GT (3 class).}
\label{fig:gtUrban3clases}
\end{subfigure}%
\begin{subfigure}{0.33\linewidth}
\centering
\includegraphics[width=5.4cm]{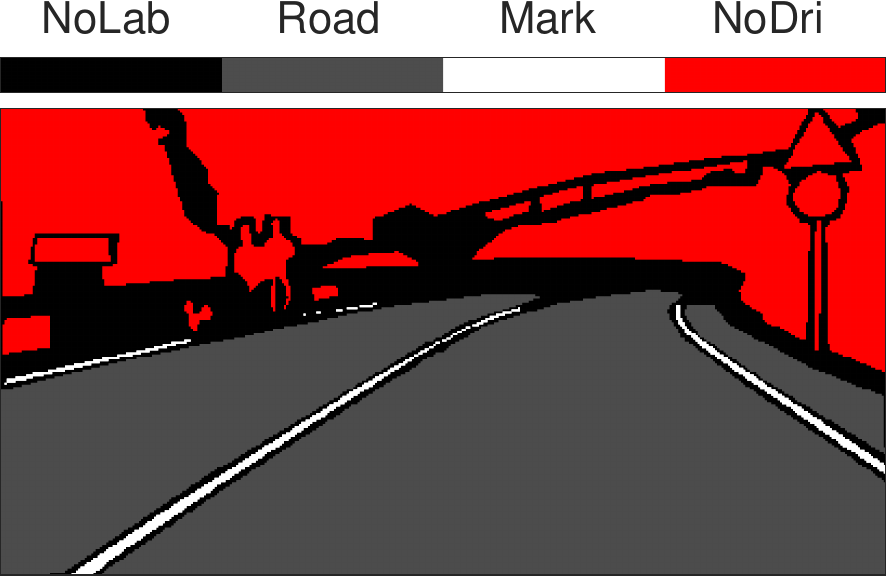}
\caption{Road, GT (3 class).}
\label{fig:gtRoad3clases}
\end{subfigure}
\begin{subfigure}{0.31\linewidth}
\centering
\includegraphics[width=5.35cm]{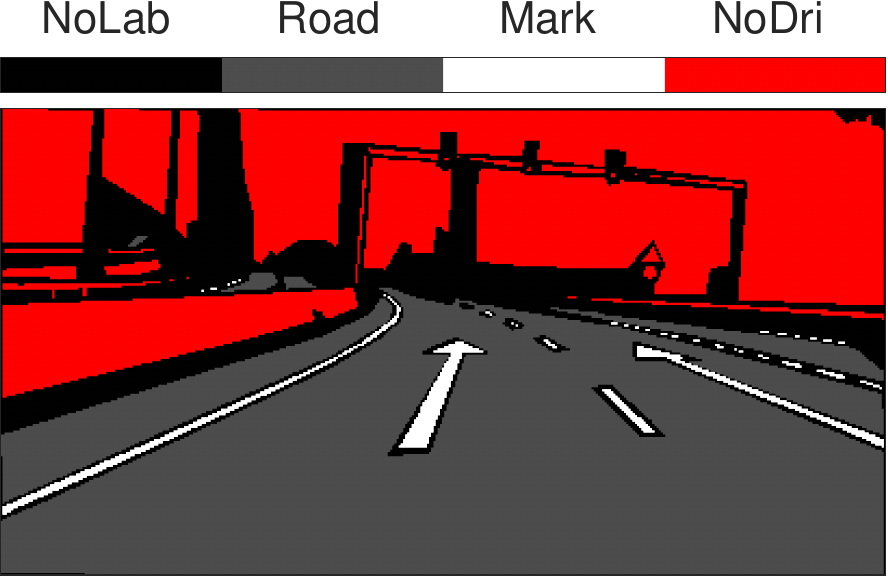}
\caption{Highway, GT (3 class).}
\label{fig:gtHighway3clases}
\end{subfigure}\\[1ex]
\begin{subfigure}{0.32\linewidth}
\centering
\includegraphics[width=5.35cm]{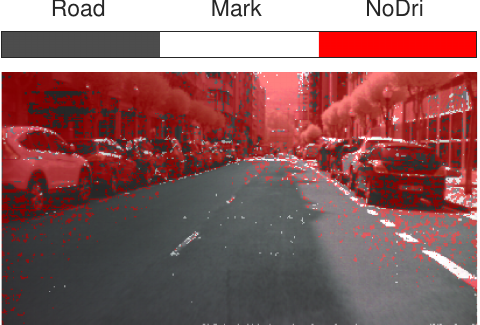}
\caption{Urban, seg. (3 class).}
\label{fig:segUrban3clasesANN}
\end{subfigure}%
\begin{subfigure}{0.33\linewidth}
\centering
\includegraphics[width=5.4cm]{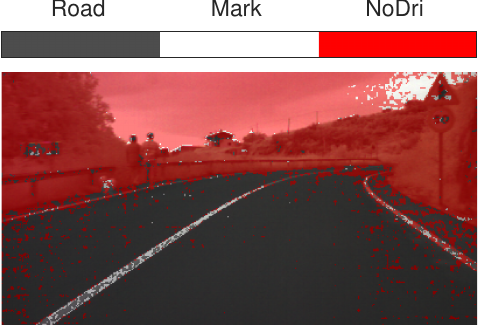}
\caption{Road, seg. (3 class).}
\label{fig:segRoad3clasesANN}
\end{subfigure}
\begin{subfigure}{0.31\linewidth}
\centering
\includegraphics[width=5.35cm]{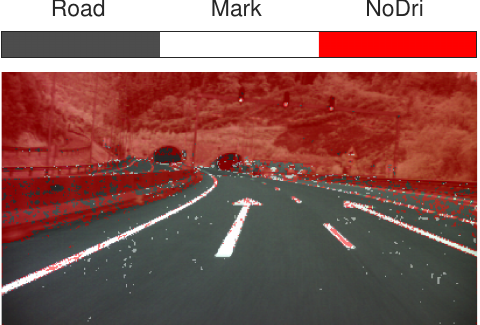}
\caption{Highway, seg. (3 class)}
\label{fig:segHighway3clasesANN}
\end{subfigure}\\[1ex]
\begin{subfigure}{0.32\linewidth}
\centering
\includegraphics[width=5.35cm]{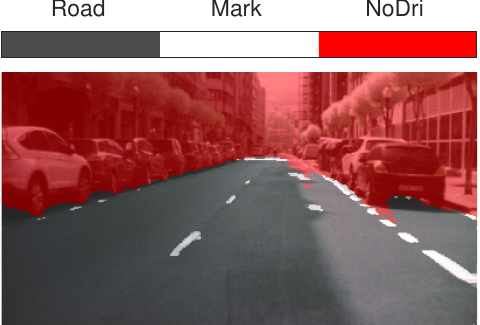}
\caption{Urban, seg. (3 class).}
\label{fig:segUrban3clases}
\end{subfigure}%
\begin{subfigure}{0.33\linewidth}
\centering
\includegraphics[width=5.4cm]{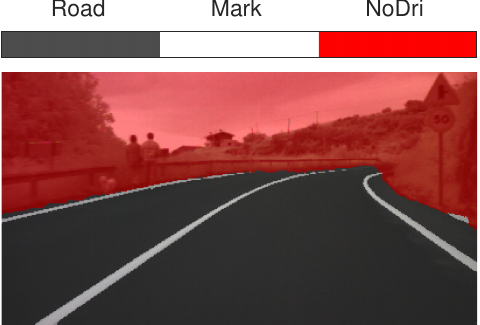}
\caption{Road, seg. (3 class).}
\label{fig:segRoad3clases}
\end{subfigure}
\begin{subfigure}{0.31\linewidth}
\centering
\includegraphics[width=5.35cm]{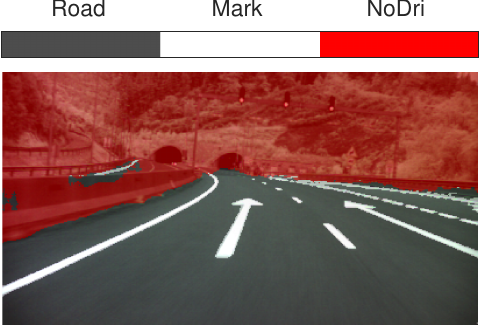}
\caption{Highway, seg. (3 class)}
\label{fig:segHighway3clases}
\end{subfigure}
\caption{Comparison among the visible (first row), ground truth (second row), ANN segmentation (third row), U-Net segmentation (forth row) images of three different scenarios: urban (first column), road (second column) and highway (third column) for the 3-class experiment.}
\label{fig:comparison3class}
\end{figure}

\begin{figure}[h!]
\begin{subfigure}{0.32\linewidth}
\centering
\includegraphics[width=5.4cm]{images/4FCNs/Vis_GT/2_nf3121_228_vis.pdf}
\caption{Urban, visible.}
\label{fig:visUrban5clases}
\end{subfigure}
\begin{subfigure}{0.32\linewidth}
\centering
\includegraphics[width=5.4cm]{images/4FCNs/Vis_GT/15_nf3232_101_vis.pdf}
\caption{Road, visible.}
\label{fig:visRoad5clases}
\end{subfigure}
\begin{subfigure}{0.32\linewidth}
\centering
\includegraphics[width=5.4cm]{images/4FCNs/Vis_GT/49_nf4313_234_vis.pdf}
\caption{Highway, visible.}
\label{fig:visHighway5clases}
\end{subfigure}\\[1ex]
\begin{subfigure}{0.32\linewidth}
\centering
\includegraphics[width=5.35cm]{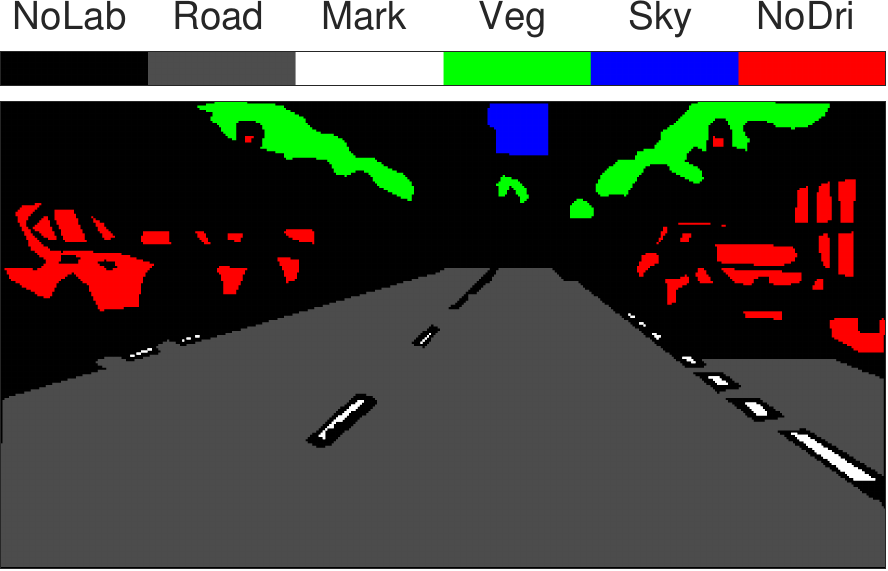}
\caption{Urban, GT.}
\label{fig:gtUrban5clases}
\end{subfigure}
\begin{subfigure}{0.32\linewidth}
\centering
\includegraphics[width=5.35cm]{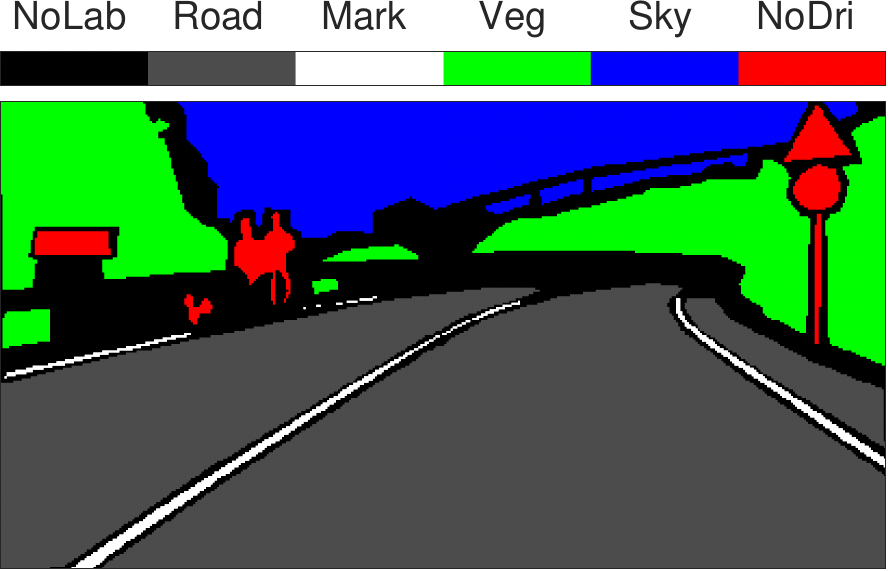}
\caption{Road, GT.}
\label{fig:gtRoad5clases}
\end{subfigure}
\begin{subfigure}{0.31\linewidth}
\centering
\includegraphics[width=5.35cm]{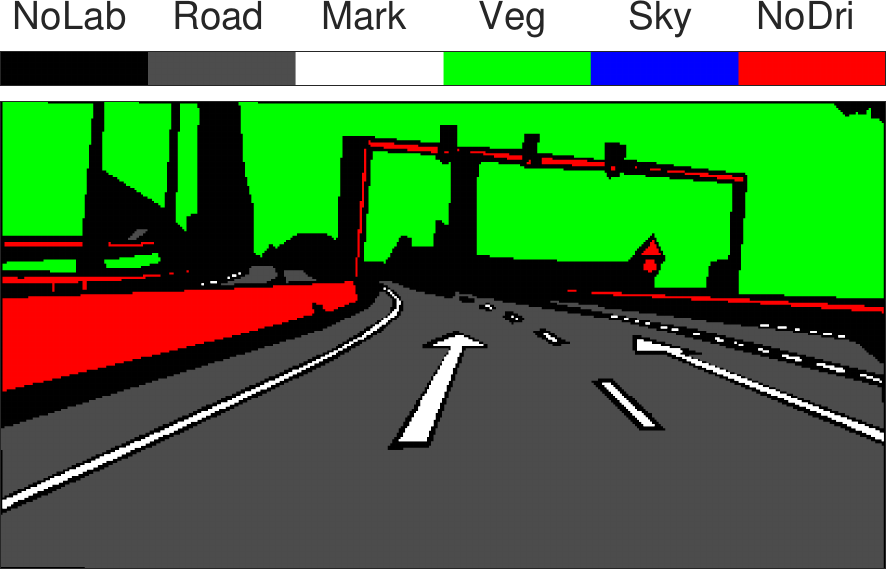}
\caption{Highway, GT.}
\label{fig:gtHighway5clases}
\end{subfigure}\\[1ex]
\begin{subfigure}{0.32\linewidth}
\centering
\includegraphics[width=5.35cm]{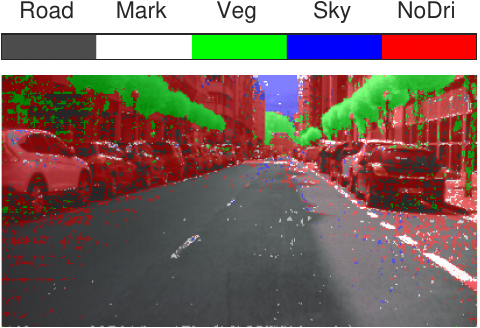}
\caption{Urban, seg. ANN.}
\label{fig:segUrban5clasesANN}
\end{subfigure}%
\begin{subfigure}{0.33\linewidth}
\centering
\includegraphics[width=5.4cm]{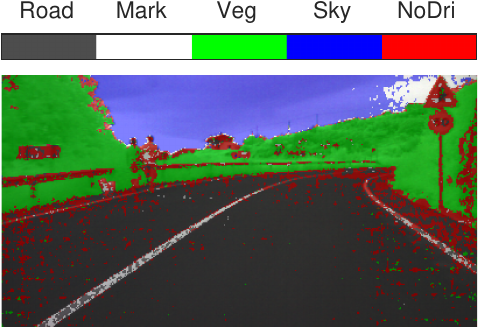}
\caption{Road, seg. ANN.}
\label{fig:segRoad5clasesANN}
\end{subfigure}
\begin{subfigure}{0.31\linewidth}
\centering
\includegraphics[width=5.35cm]{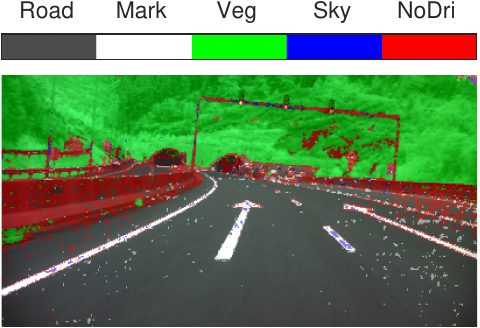}
\caption{Highway, seg. ANN.}
\label{fig:segHighway5clasesANN}
\end{subfigure}\\[1ex]
\begin{subfigure}{0.32\linewidth}
\centering
\includegraphics[width=5.35cm]{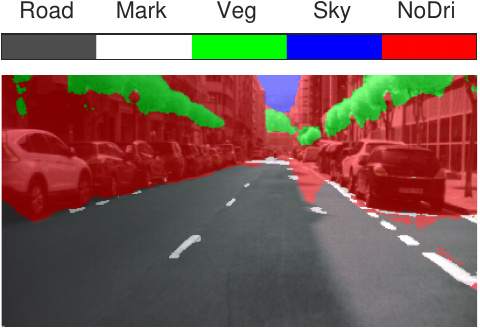}
\caption{Urban, seg. U-Net.}
\label{fig:segUrban5clases}
\end{subfigure}%
\begin{subfigure}{0.33\linewidth}
\centering
\includegraphics[width=5.4cm]{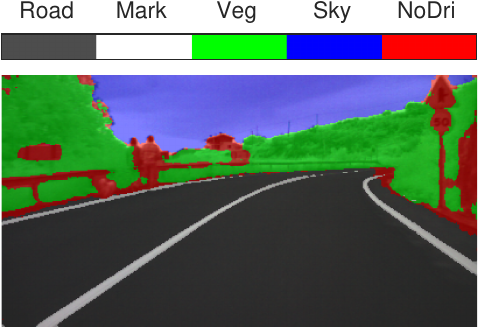}
\caption{Road, seg. U-Net.}
\label{fig:segRoad5clases}
\end{subfigure}
\begin{subfigure}{0.31\linewidth}
\centering
\includegraphics[width=5.35cm]{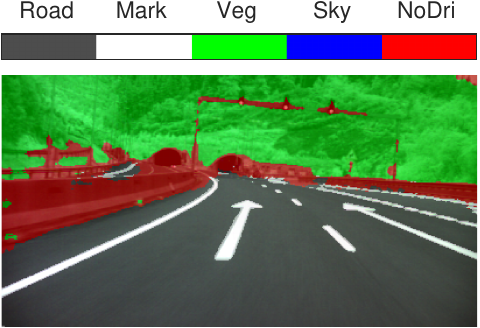}
\caption{Highway, seg. U-Net.}
\label{fig:segHighway5clases}
\end{subfigure}
\caption{Comparison among the visible (first row), ground truth (second row), ANN segmentation (third row), U-Net segmentation (forth row) images of three different scenarios: urban (first column), road (second column) and highway (third column) for the 5-class experiment.}
\label{fig:comparison5class}
\end{figure}

Analyzing the numerical results of the reconstructed images, it can be seen that all the classes show a good IoU value with the exception of the Road Marks class, which suffers from a low precision value. In the first example, in particular, for every 100 TPs of this class there are 37 FPs. However, as Road Marks is the minority class, this inaccuracy does not severally affect the overall metrics. This can be visually verified when depicting the segmentation outputs for the testing images. Figures \ref{fig:comparison3class} and \ref{fig:comparison5class} show the segmentation maps of three example images chosen from the three different driving scenarios in the dataset: an urban environment, an interurban road and a highway. The proposed FCN quite satisfactorily segments a typical driving scene for the 3-class experiment (Figure \ref{fig:segRoad3clases}) while it misinterprets some pixels in more challenging images such as those where there are objects casting their shadows on the road (Figure \ref{fig:segUrban3clases}) or where there are overlapping objects, as on the left side in (Figure \ref{fig:segHighway3clases}).

\newpage

Similarly, in the five-class example (Figures \ref{fig:segUrban5clases} to \ref{fig:segHighway5clases}), similar issues can be observed, but there is much more information about the scenes. For instance, it can be observed that the system is now able to identify the presence of some objects in the non-drivable sections of the images such as traffic signals, pedestrians and guardrails. 

\newpage

\begin{figure}[h!]
\centering
\includegraphics[width=9cm]{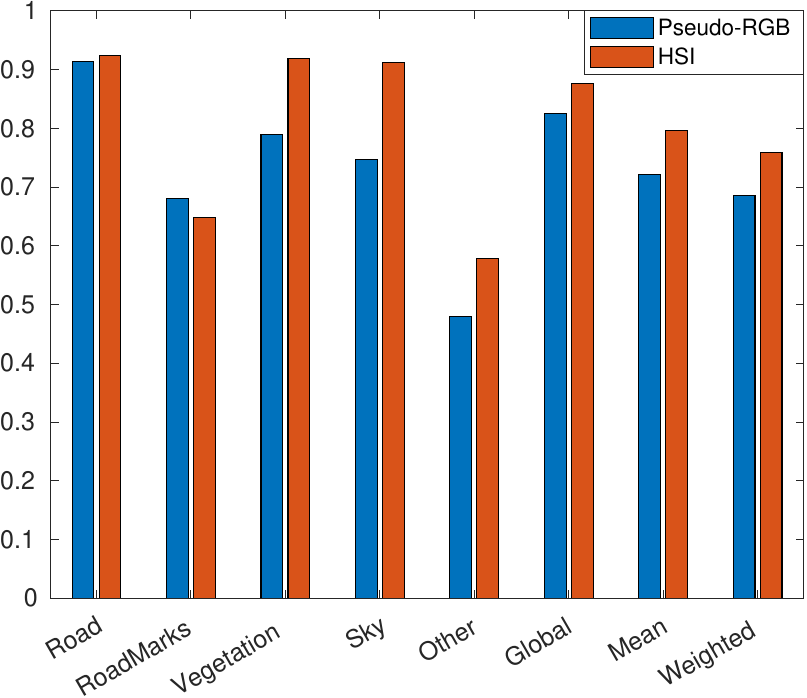}
\caption{IoU values of the rebuilt images for the 5-class-experiment using pseudoRGB images (blue) or HSI (red).}
\label{fig:HSIvsPseudoRGB5class}
\end{figure}

Following what has been done with the baseline spectral classifier, the same U-Net has been trained with pseudoRGB images which have been extracted as previously described in Subsection \ref{subsec:dataset}. Figure \ref{fig:HSIvsPseudoRGB5class} contains the IoU values of the rebuilt images for the 5-class-experiment. As it can be seen, although some classes (Road and Road Marks) are not affected by the reduction of the spectral number of channels, the rest of the classes do suffer a considerable degradation of their metrics.

\newpage

\begin{figure}[h!]
\begin{subfigure}{0.5\linewidth}
\centering
\includegraphics[width=5.5cm, center]{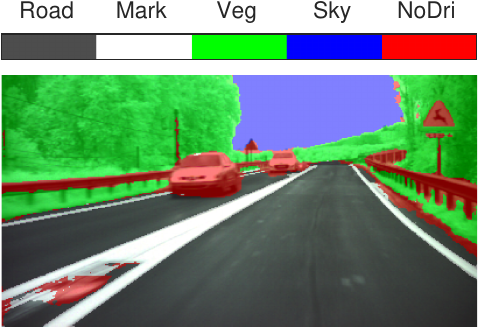}
\caption{}
\end{subfigure}
\begin{subfigure}{0.5\linewidth}
\centering
\includegraphics[width=5.5cm, center]{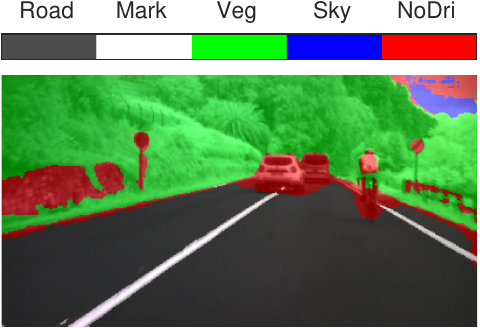}
\caption{}
\end{subfigure}\\[1ex]
\begin{subfigure}{0.5\linewidth}
\centering
\includegraphics[width=5.5cm]{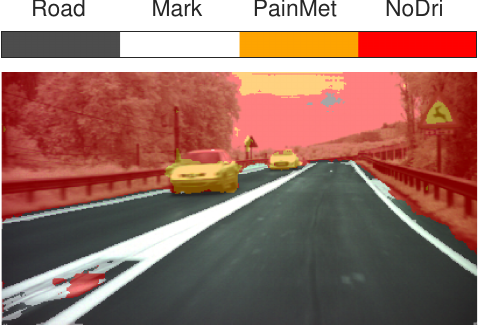}
\caption{}
\end{subfigure}
\begin{subfigure}{0.5\linewidth}
\centering
\includegraphics[width=5.5cm]{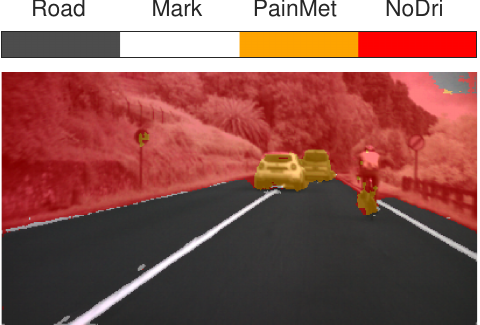}
\caption{}
\end{subfigure}
\caption{U-Net segmentation comparison of two selected images: 5-class experiment (a-b) and 3-class experiment + Painted Metal (c-d).}
\label{fig:PaintedMetal}
\end{figure}

In ADAS environments, it is of utmost importance to correctly identify traffic signals, cars and pedestrians. As Figures \ref{fig:PaintedMetal}a and \ref{fig:PaintedMetal}b show, our FCN is able to detect those obstacles for the 5-class experiment. However, when we want to determine the class to which those objects belong to, we have to train the FCN using some additional classes (as for Painted Metal in Figures \ref{fig:PaintedMetal}c and \ref{fig:PaintedMetal}d) which, for the mentioned examples, are undersampled. The results obtained are inconclusive and not robust, so we will investigate how to effectively combine spectral classifiers (which can easily cope with unbalanced datasets) with U-Net to improve the segmentation results. 

\newpage

\section{System Prototyping and Characterization}\label{sec:workflow}
As stated above, there are two main processes involved in the implementation of this HSI segmentation system: the preprocessing of the raw images provided by the snapshot camera and the FCN-based AI. In this section we describe in detail how both processing stages have been developed, implemented and tested on three different embedded processing systems: a low-cost single board computer, an embedded GPU SoC and a PSoC with an embedded FPGA. We also give details of the benchmarking setup used for the characterization of all three implementations and we analyze the results obtained.

\subsection{Benchmark Setup}
Proper hardware selection is essential for the successful deployment of AI in an edge device. In the particular case of the ADAS/ADS, this decision is conditioned by the need to satisfy strict specifications in terms of processing latency, cost and energy consumption without compromising the overall performance of the algorithm (safety criticality considerations aside). Most current edge-AI deployments rely on SoCs including AI-oriented coprocessors (domain specific or even application specific). However, depending on the complexity of the AI model (number of parameters and FLOPS) and the constraints associated with the specific application, the use of cheaper, lower-end devices could provide satisfactory results. Consequently, we have selected three different prototyping platforms to explore the performance of our application: a Raspberry Pi 4B, a NVIDIA Jetson Nano Development Kit and an AMD-Xilinx ZCU104 development board.

The Raspberry Pi 4 Model B (Pi4B) is the leading single-board computer of the Raspberry Pi family with significant enhancement in CPU, GPU and I/O performance. This single-board computer consists of a quad-core Cortex-A72 (ARM v8) 64-bit SoC 1.5GHz with 8GB of LPDDR4 SDRAM, which includes a recent upgrade that lets the cores work in turbo-boost mode and reach a peak frequency of 1.8GHz safely \cite{rpi4b}. The NVIDIA Jetson Nano is a small, powerful computer for embedded AI applications with 4 GB 64-bit LPDDR4 SDRAM. The SoC combines a Quad-core ARM Cortex-A57 MPCore processor (1.43GHz of maximum theoretical frequency) with a Maxwell architecture NVIDIA GPU containing 128 CUDA cores \cite{JetsonNanoUserGuide}. Finally, the ZCU104 development board includes a Zynq UltraScale+ MPSoC with a quad-core ARM Cortex-A53 processor (1.5GHz of maximum theoretical frequency) and a dual-core Cortex-R5 real-time processor in the Processing System (PS) connected to a 16nm FinFET Programmable Logic (PL) with access to a 2 GB 64-bit wide DDR4 external memory \cite{ZCU104userGuide}.

Although all three devices include ARM v8-A architecture Quad-Core Cortex processors (A53, A57 and A72), they do not only differ in CPU frequency but also in other aspects, such as L1 I/D Cache Size, L2 Cache Size, execution order, number of pipeline stages and so on. Cortex A53 is said to be a power-efficient processor when compared to the more powerful A57, while A72, which is the direct successor of the A57, was designed to improve its predecessor in the PPA metric: performance, power and area \cite{cortexcomparison}.

\newpage

\begin{table}[h!]
\centering
\resizebox{14cm}{!}{%
\begin{tabular}{|c|c|c|c|}
\hline
\textbf{\diagbox[]{\textbf{Properties}}{\textbf{Core}}} & \textbf{Cortex-A53} & \textbf{Cortex-A57} & \textbf{Cortex-A72} \\ \hline
\textbf{Revision}              & v8.0-A                                                                     & v8.0-A              & v8.0-A              \\ \hline
\textbf{Platform}              & Zynq UltraScale+ MPSoC   & Jetson Nano SoM     & Raspberry Pi 4B     \\ \hline
\textbf{OS}                    & Petalinux 2022.1\tablefootnote{This OS has been used to work more comfortably with AMD-Xilinx's Tools. However, for TensorFlow Lite model development, we have opted for Ubuntu 20.04.3 LTS.} & Ubuntu 18.04.6 LTS & Ubuntu 22.04.1 LTS     \\ \hline
\textbf{Number of cores}       & 4                        & 4                   & 4                   \\ \hline
\textbf{Execution}             & In-order                 & Out-of-order        & Out-of-order        \\ \hline
\textbf{Pipeline depth}        & 8                        & 15+                 & 15+                 \\ \hline
\textbf{Superscalar}           & Yes                      & Yes                 & Yes                 \\ \hline
\textbf{Neon and FPU}          & Yes                      & Yes                 & Yes                 \\ \hline
\textbf{L1 I Cache (per core)} & 32 KB                    & 48 KiB              & 48 KiB              \\ \hline
\textbf{L1 D Cache (per core)} & 32 KB                    & 32 KiB              & 32 KiB              \\ \hline
\textbf{L2 Cache}              & 1 MB                     & 2 MiB               & 1 MiB               \\ \hline
\textbf{Clock rate}            & 1.199 GHz                & 1.479 GHz           & 1.8 GHz             \\ \hline
\textbf{External DDR4 memory}  & 2 GB                     & 4 GB                & 8 GB                \\ \hline
\end{tabular}}
\label{tab:comparisonPlatformsProcessors}
\caption{Comparison among the selected platforms focusing on the processors' characteristics.}
\end{table}

\begin{figure}[h!]
\centering
\includegraphics[height=6cm]{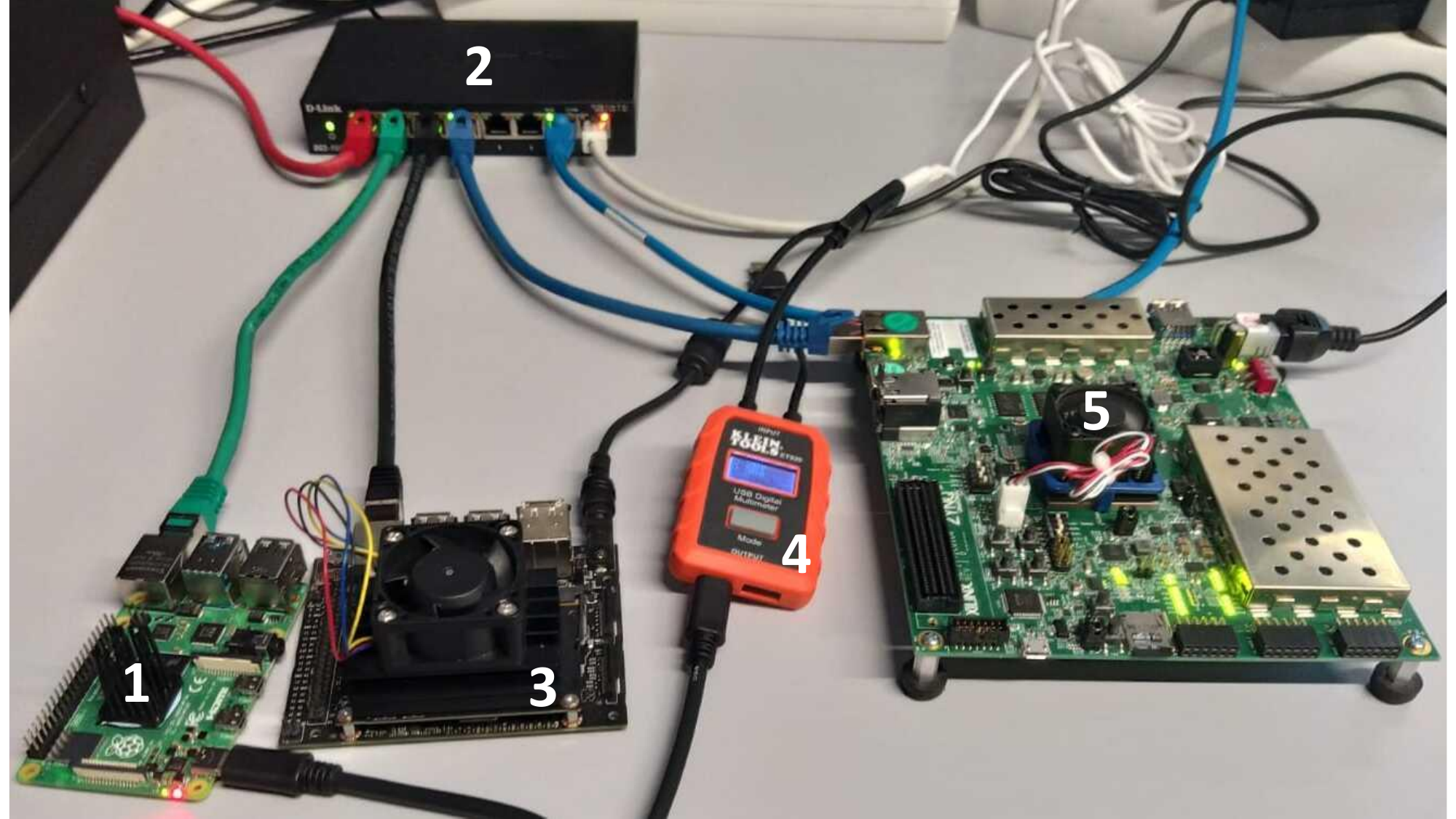}
\caption{Power measurement infrastructure and communication with computing boards: 1, Raspberry Pi 4B; 2, Router; 3, Jetson Nano; 4, Power meter and 5, ZCU104.}
\label{fig:infraestructura}
\end{figure}

In order to make a fair power and energy consumption comparison, we disabled as many peripherals as possible; this includes, in the case of the Pi4B, onboard LEDs, microHDMI, bluetooth or Wifi, for example. Power measurements have been performed either by software commands or by plugging an external power meter between the power supply and the device (Figure \ref{fig:infraestructura}). We have selected a USB-C digital meter from Klein Tools \cite{kleinTools} that reliably measures (accuracy of $\pm$ 1\%) the voltage and current (with a resolution of 0.01V and 0.01A respectively) that flows through both the Raspberry Pi 4B and the Jetson Nano boards. Furthermore, the \textit{tegrastats} command allows us to extract valuable information about the power consumption of the processing elements (CPU, GPU and overall) of the Jetson Nano, so a more comprehensive analysis can be derived. However, there is a mismatch between the values obtained from the \textit{tegrastats} command and the values read from the digital power meter, an issue that has also been reported previously, such as in \cite{holly2020profiling}. As for the Zynq UltraScale+ MPSoC, power monitoring is carried out with the Infineon IR PowerCenter GUI that obtains power rail values through Infineon IRPS5401 power controllers \cite{ZCU104userGuide}. The onboard Infineon IRPS5401 power controllers are accessed through an I2C connector included in the Infineon USB Cable \cite{infineonPowerController, infineonUSB, ZCU104schematic}.

With regard to the communication infrastructure developed to ease the benchmarking process on multiple devices, in Figure \ref{fig:infraestructura} a router can be seen that is used to create a local area network which allows us to communicate with various prototyping boards via the ssh protocol while their Internet access is granted. 

\subsection{Raw image preprocessing}\label{subsec:imagePre}
Recording in real driving conditions implies that the setup of the hyperspectral camera cannot be kept invariable. Moreover, lighting conditions cannot be controlled due to changes in the scenarios, weather conditions and daylight. Additionally, the use of a moving platform (the vehicle) and the presence of moving objects severely constrain the exposition time and, in consequence, the amount of light that can reach the sensor in each situation. The HSI-drive database contains images acquired in very diverse conditions that require different setups of the camera and optics. This diversity makes it more difficult to effectively preprocess the raw images in order to generate hyperspectral cubes with equivalent characteristics that can be used to successfully train AI models. In addition, this preprocessing has to be replicated on-the-fly during the inference phase, so the computational aspects of the preprocessing pipeline have also to be addressed.

\begin{figure}[h!]
\centering
\includegraphics[height=5.75cm]{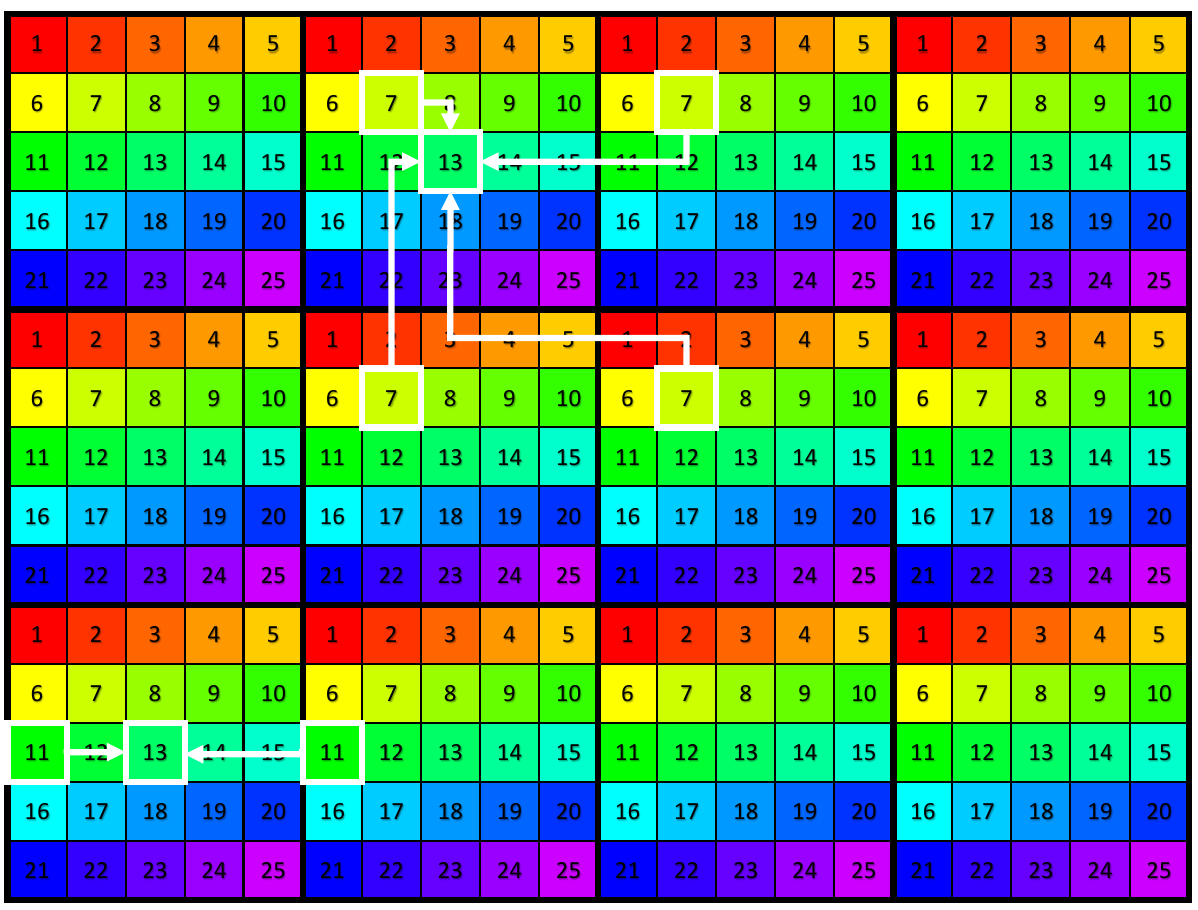}
\caption{Translation to center algorithm.}
\label{fig:translationToCenter}
\end{figure}

The processing stages involved in the pipeline of the raw image preprocessing to obtain the corresponding reflectance hyperspectral cubes depend on the type of application. In our context, the programmed preprocessing algorithm consists of the following four steps: crop and clip, reflectance correction, band extraction, and translation to center (partial demosaicing).

As reported in Section \ref{sec:expSetup}, the CMOS sensor has a standard size of 1088x2048. However, as the multispectral filter array (MSFA) does not cover the entire surface of the sensor, the resulting active area has a lower resolution (1080x2045) and so the acquired raw images need to be cropped and clipped. After that, the acquired radiance is normalized by taking a white reference frame as the highest response from the sensor, resulting in a reflectance signal which is assumed to be robust under different acquisition conditions. In order to eliminate the static noise of the sensor, bias correction is previously performed by subtracting a dark reference frame from both the image and the white reference frames. The next step, known as band extraction, transforms the 2D representation obtained from the mosaic-shaped MSFA (Figure \ref{fig:translationToCenter}) into a hyperspectral cube. As a consequence of the snapshot camera technology, each of the 25 pixels of the 5x5 mosaic contains spectral information about a different wavelength (Figure \ref{fig:translationToCenter}) and thus each pixel lacks the spectral content related to the remaining 24 wavelengths. Therefore, to reconstruct from the raw data either a full resolution image (1080x2045x25) or a partial resolution image (216x409x25) a typical approach is to use an interpolation algorithm in a process known as (partial) demosaicing. In our application, we have opted for a partial demosaicing process which consists of applying linear or bilinear interpolation techniques to interpolate the reflectance value of the missing spectral bands in certain spatial positions which are the center of the mosaic (hence its name, translation to center). Two examples are shown in Figure \ref{fig:translationToCenter}. On the upper part of the image, we want to interpolate in the spatial position which corresponds to band 13, the reflectance value that we would obtain if we placed there a band-7 filter. For that purpose, we take into account the values of the surrounding four band-7 pixels, and the distance from the band-13 pixel, and apply a bilinear interpolation to extract that value. We perform this operation to extract all the missing information from the remaining bands but only in the central position of each mosaic. However, as the other example suggests, there are some positions in the contour of the image for which there are not 4 neighbouring pixels and so a different treatment (such as linear interpolation) is required. 

Differently from what is made in \cite{gutierrez22}, the cube generation contains neither the median filtering nor the band normalization steps. This is because although a prior analysis demonstrated the benefits of using a median filter before training a multi-layer feedforward network, those effects are largely overshadowed in an FCN by the convolutional layers which act as a kind of spatial regularizer and, consequently, the overall latency is greatly reduced. Regarding the normalization, we have performed hardware acceleration on it, including this step as the first layer of U-Net. 

\subsubsection{Embedded Software Programming}
The raw image preprocessing algorithm has been programmed in C and compiled to be executed as an embedded Linux application in the microprocessors of the benchmark devices as part of the HW/SW codesign for the implementation of the system. To reduce its latency, we have combined thread-level parallelism (OpenMP pragmas) with data-level parallelism (Single instruction multiple data, SIMD, via Neon). OpenMP is an API to develop parallel applications on shared-memory processors, such as embedded systems and accelerators, in a flexible manner \cite{OpenMP}. OpenMP grants the programmer control of the thread creation, workload distribution among the threads (not only the scheduling type but also the restriction of a certain task to one thread), thread synchronization and variable attributes (i.e. which variables are shared among threads and which are kept private to avoid data incoherence and race conditions). The way of working with OpenMP is by adding compiler directives or pragmas and modifying environment variables that condition both the compile and runtime behaviour of the program in a fork-join model. OpenMP can be combined with Neon (ARM Advanced SIMD architecture) and Floating Point technologies which are fully integrated into the processor and share the processor resources for integer operation, loop control and caching, significantly reducing the area and power cost \cite{leaTheArcIntNeo}. SIMD instructions are carried out in the 32 128-bit SIMD/floating-point registers that the AArch64 architecture includes.

To implement those techniques in our software design, on the one hand, some compiler directives have been added to the code after analyzing how to parallelize the code execution among the different threads. The analysis consisted of identifying parallel regions (coarse-grained parallelism) and then focusing on the parallelizable loops (fine-grained parallelism). In order to parallelize loops, we have studied the dependency between loop indexes or data inside the loops. Besides this, as all threads can modify and access all variables, we have also examined when to change the scope of one of the variables and make them local to each thread to avoid undesirable behaviour. Finally, we have checked if the execution time of each iteration varies inside a loop to establish how to distribute the workload among threads. On the other hand, we have also included compiler autovectorization to boost the use of the Neon technology.

To evaluate the extent to which the use of SIMD instructions and multiprocessing accelerates our application, an ablation study has been performed. The results of the four combinations are shown in Table \ref{tab:ablationStudy}.

\begin{table}[h!]
\centering
\resizebox{10cm}{!}{%
\begin{tabular}{c|cccc|}
\cline{1-5}
\multicolumn{1}{|c|}{\diagbox[]{\textbf{Pipeline step}}{\textbf{Use of SIMD-MP}}} & \multicolumn{1}{c|}{\textbf{00}} & \multicolumn{1}{c|}{\textbf{01}} & \multicolumn{1}{c|}{\textbf{10}} & \multicolumn{1}{c|}{\textbf{11}} \\ \hline
\multicolumn{1}{|c|}{\textbf{Image cropping (ms)}} & \multicolumn{1}{c|}{6.654} & \multicolumn{1}{c|}{2.883} & \multicolumn{1}{c|}{6.654} & 2.886  \\ \hline
\multicolumn{1}{|c|}{\textbf{Reflectance correction (ms)}} & \multicolumn{1}{c|}{251.803} & \multicolumn{1}{c|}{90.902} & \multicolumn{1}{c|}{50.369} & 21.519 \\ \hline
\multicolumn{1}{|c|}{\textbf{Band extraction (ms)}} & \multicolumn{1}{c|}{53.063} & \multicolumn{1}{c|}{16.217} & \multicolumn{1}{c|}{311.803} & 13.042 \\ \hline
\multicolumn{1}{|c|}{\textbf{Translation to center (ms)}} & \multicolumn{1}{c|}{340.196} & \multicolumn{1}{c|}{162.219} & \multicolumn{1}{c|}{22.586} & 13.583  \\ \hline
\multicolumn{1}{|c|}{\textbf{Total (ms)}} & \multicolumn{1}{c|}{\textbf{651.716}} & \multicolumn{1}{c|}{\textbf{272.223}} & \multicolumn{1}{c|}{\textbf{111.453}} & \textbf{51.030}\\ \hline
\end{tabular}}
\caption{Mean execution time over 1000 iterations of the C image-preprocessing Linux application on the ZCU104 development board combining the use or not of SIMD and MP.}
\label{tab:ablationStudy}
\end{table}

The combination of SIMD instructions with the thread-level parallelism enabled by the QuadCore ARM-Cortex A we are working with significantly reduces latency with respect to the options where no optimization (12x), or only one of them (5x for MP-only and 2x for SIMD-only approaches), is used. In addition, by analyzing each of the preprocessing stages individually, some conclusions that are consistent with the characteristics of the stages can be drawn.

For the crop and clip stage, the only variable that conditions the execution time is the use or not of multiprocessing. This can be easily explained since this stage only consists of cropping and copying an array where the calculation of the start/end indexes is straightforward.

On the other hand, as previously mentioned, the reflectance correction stage consists of a normalization based on several references, so that a significant acceleration can be achieved with only the use of SIMD instructions, although this is more evident in the median filtering stage, as will be discussed below. 

As with the first stage, during partial demosaicing, the image is transformed from a two-dimensional representation to a three-dimensional representation, so the differential factor is whether or not to use multiprocessing. However, since the way in which the indexes for band extraction are calculated requires several mathematical operations that can be parallelized, the use of SIMD instructions additionally reduces the latency.

Of the four stages of preprocessing, band alignment is undoubtedly the one that contains the most mathematical operations, both when choosing the indexes of the pixels selected for interpolation and when performing the interpolation operation itself. This justifies the considerable acceleration observed between using either only SIMD or only opting just for multiprocessing.

\subsubsection{Test Results}
Table \ref{tab:cubeLat} shows the mean latency over 1000 iterations of the raw image preprocessing pipeline running on the Cortex A-72 (Raspberry Pi 4B), Cortex A-57 (Jetson Nano), and Cortex A-53 (ZCU104) quad-core processors. According to these figures and if no pipeline schemes are implemented, this processing stage would limit the reachable throughput to 19.10 FPS, 40.56 FPS and 19.84 FPS respectively.

\newpage

\begin{table}[h!]
\centering
\resizebox{8cm}{!}{%
\begin{tabular}{c|ccc|}
\cline{1-4}
\multicolumn{1}{|c|}{\diagbox[]{\textbf{Pipeline step}}{\textbf{Device type}}} & \multicolumn{1}{c|}{{\color[HTML]{CB0000} \textbf{RPI}}} & \multicolumn{1}{c|}{{\color[HTML]{32CB00} \textbf{JN}}} &\multicolumn{1}{c|}{{\color[HTML]{3166FF} \textbf{ZCU}}} \\ \hline
\multicolumn{1}{|c|}{\textbf{Image cropping (ms)}} & \multicolumn{1}{c|}{6.128} & \multicolumn{1}{c|}{2.309} & 2.886 \\ \hline
\multicolumn{1}{|c|}{\textbf{Reflectance correction (ms)}} & \multicolumn{1}{c|}{13.630} & \multicolumn{1}{c|}{7.729} & 21.519 \\ \hline
\multicolumn{1}{|c|}{\textbf{Band extraction (ms)}} & \multicolumn{1}{c|}{21.124} & \multicolumn{1}{c|}{8.324} & 13.042 \\ \hline
\multicolumn{1}{|c|}{\textbf{Translation to center (ms)}} & \multicolumn{1}{c|}{11.453} & \multicolumn{1}{c|}{6.290} & 13.583 \\ \hline
\multicolumn{1}{|c|}{\textbf{Total (ms)}} & \multicolumn{1}{c|}{\textbf{52.335}} & \multicolumn{1}{c|}{\textbf{24.652}} & \textbf{51.030} \\ \hline
\end{tabular}}
\caption{Mean execution time over 1000 iterations of the C image-preprocessing Linux application on the benchmarked devices: Raspberry Pi 4B (RPI), Jetson Nano (JN) and ZCU104 development board (ZCU).}
\label{tab:cubeLat}
\end{table}

\subsection{FCN Inference}
The implementation of the FCN has been carried out in several phases, starting with the study of the effect of different quantization levels on the ARM v8 quad-core Cortex processors and ending with the implementation of the optimized models on the AI hardware accelerators, when available, such as the NVIDIA GPU and the AMD-Xilinx DPUs on the FPGA.

\subsubsection{Implementation Workflow and Tools}\label{subsubsub:algorithmDeploymentInference}
The design, training and validation of the FCN has been performed with MATLAB's Deep Learning Toolbox. While the most recent version of MATLAB includes tools to directly export deep models, including segmentation FCNs, to other frameworks such as TensorFlow, and even provide tools for the automatic deployment of deep models  in FPGAs \cite{dlt}, this was not possible with the Matlab version used in this project (2021b).

\begin{figure}[h!]
\centering
\includegraphics[height=5.25cm]{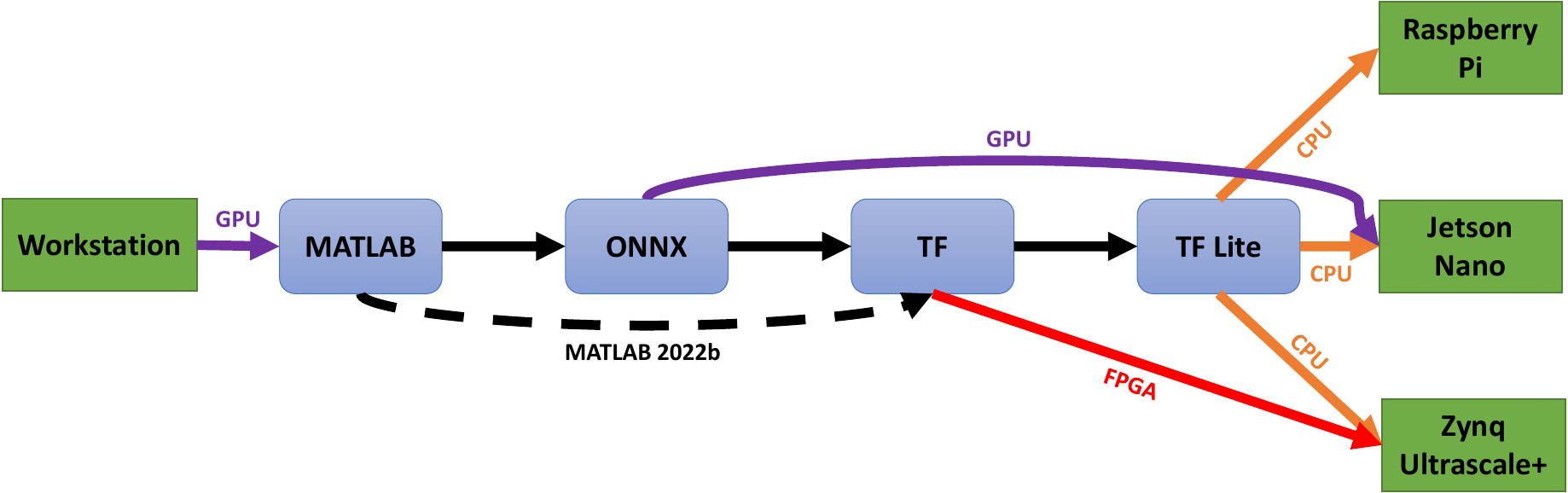}
\caption{Path adopted to transfer the FCN from one representation to another.}
\label{fig:frameworkDiagram}
\end{figure}

Consequently, to link our ML development environment to any implementation workflow, we opted for exporting the MATLAB-generated deep models to the Open Neural Network Exchange (ONNX) representation. ONNX is an open standard format for representing machine learning models that can be then used with a variety of frameworks, tools and compilers. In our workflow, the ONNX representation was exported to Keras/TensorFlow \cite{tensorflow2015-whitepaper} via \textit{onnx2keras} \cite{onnx2keras}. Figure \ref{fig:frameworkDiagram} shows the flowchart that represents the workflow adopted to transfer the FCN from one representation to another and adapt it to the requirements of both the DL frameworks and the computing platforms under study.

As has been remarked in Subsection \ref{subsec:FCNs}, we favoured the use of a smaller-size FCN architecture over the most accurate one in order to reduce memory footprint and computational load. In an initial approach, we tested the software execution of the FCN directly on the three Cortex microprocessors. With that aim in mind, we used TensorFlow Lite, a library for deploying models on smartphones, microcontrollers and other edge devices. By default, TensorFlow Lite utilizes CPU kernels that are optimized for the Neon instruction subset of ARM, but which are not necessarily optimized for the intensive computational workload of deep models. Thus, to reduce the final latency as much as possible, we partially used XNNPACK, a highly optimized library of floating-point, neural network inference operators for ARM \cite{XNNPACK}, as the inference engine, which we combined with the multicore execution. Nevertheless, the obtained latency values are not reasonable in ADAS; for this reason, in order to speed up the FCN inference and reduce the memory footprint in all cases, we opted for another strategy and performed post-training quantization using TensorFlow Lite.

As we have prioritized the highest possible speed-up, we have opted for the full integer quantization approach where the whole network (weights, biases, activation functions, input and output) is dynamically quantized (a calibration dataset is therefore needed) using 8-bit integer precision. It is interesting to remark that, although having applied full integer quantization, the size of the data buffer is not reduced fourfold. This is because in the quantization procedure, bias vectors are still quantized as 32-bit data. As is justified in \cite{jacob2018quantization}, there is a need for the high precision of bias vectors because, as they are added to many output activation functions, otherwise, this could result in an overall bias which could endanger the accuracy of the network. Besides, 8-bit integer products are accumulated using a 32-bit accumulator so 32-bit registers cannot be avoided.

Finally, in order to get the best possible performance out of the benchmark computing platforms, we have implemented the FCN on the dedicated AI coprocessors, i.e. the embedded GPU in the Jetson Nano and the DPUs in the MPSoC's FPGA. In order to make as fair a comparison as possible, the three devices were programmed using their preferred DL framework, i.e. Tensorflow Lite for the Raspberry Pi 4B, Tensor RT for the Jetson Nano, and Vitis AI for the AMD-Xilinx MPSoC. TensorRT, built on the NVIDIA CUDA parallel programming model, is an SDK provided by NVIDIA to perform low latency and high throughput DL inference applications by optimizing trained neural networks and calibrating them for low precision without accuracy degradation. The resultant network can be deployed in diverse platforms, ranging from hyperscale data centers to embedded processors \cite{tensorRT}. Vitis AI is a development platform for AI inference on AMD-Xilinx hardware platforms. Vitis AI Quantizer exploits quantization and the VAI Optimizer applies channel pruning techniques to reach the high-throughput and low-latency requirements of ADAS applications. According to \cite{quantModel}, by converting the 32-bit floating point weights and activations to an 8-bit integer format, Vitis AI Quantizer can reduce computing complexity without losing prediction accuracy and, as the fixed-point network model requires less memory bandwidth, a faster speed is also provided. After compiling the quantized FCN and the DPU with Vitis AI Compiler, the output product is loaded at runtime in the system composed of the ARM CPU and the DPU accelerator in the MPSoC by a C++ application.

The Deep Learning Processor Unit (DPU) is a configurable computation engine, optimized for convolutional neural networks, which can be customized to some extent by modifying its parallelism degree and logic resource utilization. It is implemented in the programmable logic of the MPSoC with direct AXI connections to the processing system. The DPU executes compiled microcode generated from a neural network graph by fetching the instructions from off-chip memory to control how the computing engine behaves. To achieve high throughput, efficiency and reduce external memory bandwidth requirements, on-chip memory is reused as much as possible to buffer input activations, intermediate feature-maps and output activations. The user-configurable parameters are DSP slices, LUT, block RAM and UltraRAM usage. There are even other options for additional functions such as channel augmentation, average pooling, depthwise convolution and softmax. For more detailed information, the reader is referred to \cite{dpu}.

\begin{figure}[h!]
\centering
\includegraphics[height=10cm]{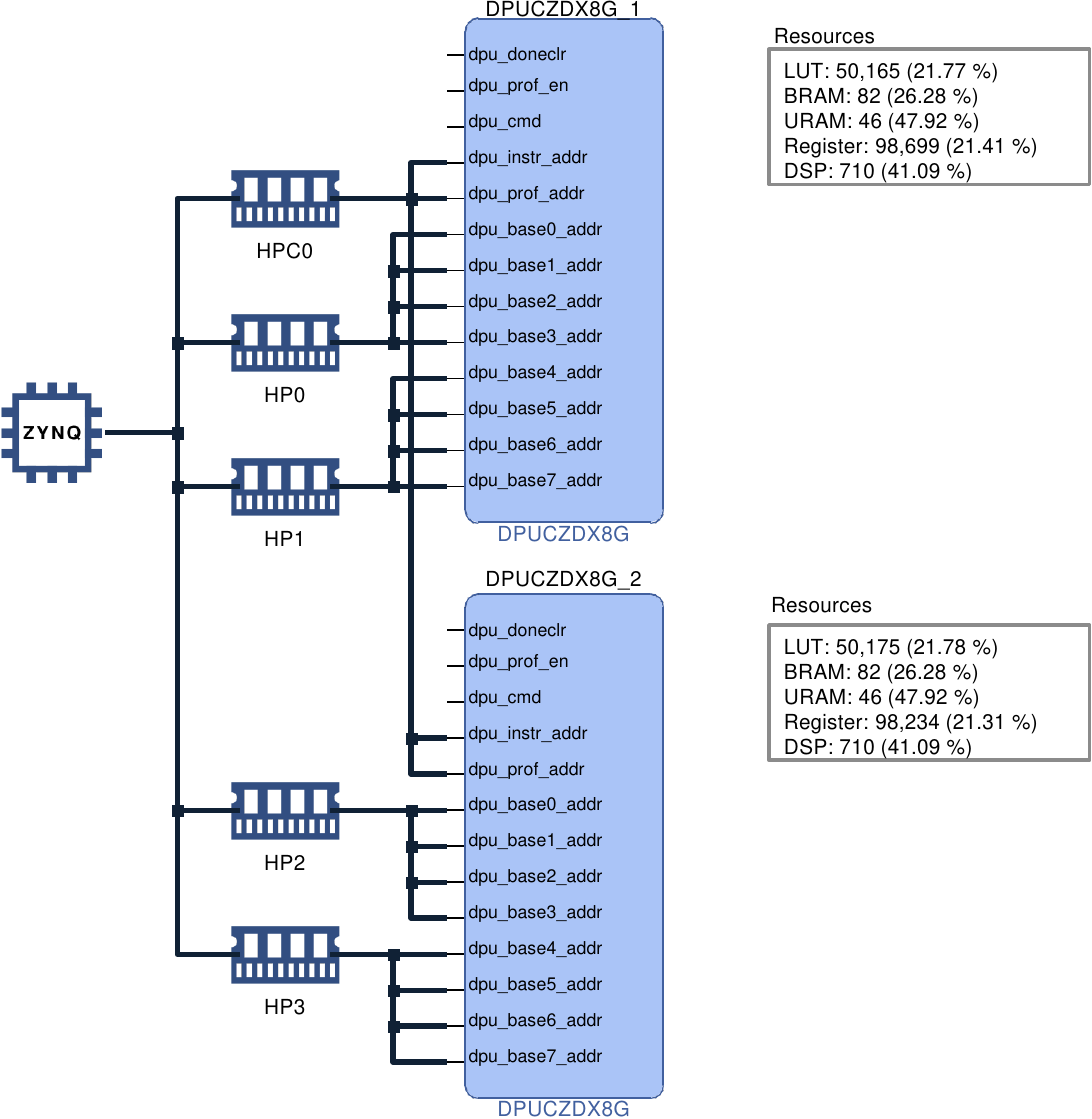}
\caption{System diagram of the DPU.}
\label{fig:systemDiagram}
\end{figure}

As can be seen in Figure \ref{fig:systemDiagram}, our implementation contains 2 DPUCZDX8G cores, each of which features a B4096 architecture which means that 4096 operations are performed per DPU clock cycle, which is set to 300MHz. More specifically, in each clock cycle, 8 pixels of the input feature map (which can have, at most, 16 channels) are multiplied by their respective portions of the 16 convolutional filters, so 8x16x16 multiplication operations and sums are performed; this results in a total of 4096 operations. Those three values are architecture-specific parameters known as pixel parallelism, input channel parallelism and output channel parallelism, the latter two sharing the same value. Figure \ref{fig:systemDiagram} also shows the logic resources occupied by this implementation.

Although the three DL frameworks allow quantizing models to 8 bits, not all devices are capable of performing computations in this type of representation. The FCN has thus been quantized according to what the respective microarchitectures allow: an 8-bit integer format for the Raspberry Pi, a 16-bit FP format for the Jetson Nano (although as indicated by NVIDIA, TensorRT would choose a higher-precision FP32 kernel if this results in overall lower runtime, or if no low-precision implementation exists), and an 8-bit format representation for the AMD-Xilinx MPSoC.

\subsubsection{Test Results}\label{subsubsub:benchmarkInference}
The implementation of the neural network layers varies slightly from TensorFlow to TensorFlow Lite so, it has first been verified that there are no significant deviations when exporting the U-Net floating point model from TensorFlow to the Lite version.

The first test has been performed on the software execution of the FCN on the CPU of the three devices both before and after quantization. As can be seen in Table \ref{tab:cpuInferenceLatency}, the throughput associated to the FP32 inference latency hardly reaches 1 FPS, while the 8-bit integer implementation produces a notable speed-up, although not enough to meet the application requirements (1.2 FPS to 2.4 FPS).

\begin{table}[h!]
\centering
\resizebox{10cm}{!}{%
\begin{tabular}{|c|cccccccc|}
\hline
\textbf{Framework (Accelerator)} &
  \multicolumn{8}{c|}{\textbf{TF Lite (CPU)}} \\ \hline
\textbf{Data quantization} &
  \multicolumn{4}{c|}{\textbf{FP32}} &
  \multicolumn{4}{c|}{\textbf{INT8}} \\ \hline
  \diagbox[]{\textbf{Measurement}}{\textbf{Platform}} &
  \multicolumn{1}{c|}{{\color[HTML]{CB0000} \textbf{RPI}}} &
  \multicolumn{2}{c|}{{\color[HTML]{32CB00} \textbf{JN}}} &
  \multicolumn{1}{c|}{{\color[HTML]{3166FF} \textbf{ZCU}}} &
  \multicolumn{1}{c|}{{\color[HTML]{CB0000} \textbf{RPI}}} &
  \multicolumn{2}{c|}{{\color[HTML]{32CB00} \textbf{JN}}} &
  {\color[HTML]{3166FF} \textbf{ZCU}} \\ \hline
\textbf{Latency (s)} &
  \multicolumn{1}{c|}{1.910} &
  \multicolumn{2}{c|}{0.835} &
  \multicolumn{1}{c|}{1.345} &
  \multicolumn{1}{c|}{0.613} &
  \multicolumn{2}{c|}{0.416} &
  {0.820} \\ \hline
\end{tabular}}
\caption{CPU-inference latency characterization of the unquantized and quantized FCN on the three devices: Raspberry Pi 4B (RPI), Jetson Nano (JN) and ZCU104 development board (ZCU).}
\label{tab:cpuInferenceLatency}
\end{table}

The second test has been performed on the implementations that use the full power of the SoCs, either with the embedded GPU or the embedded DPU/FPGA. As can be observed in Table \ref{tab:inferenceBenchmarking}, the latency is dramatically reduced in both cases, with a x4.47 speed-up in the Jetson Nano and a x22.16 speed-up in the AMD-Xilinx MPSoC. To obtain those values, as each image consists of 18 patches, we have opted for using a 18-patch batch size, although further experiments could be carried out to find the best size to optimize latency–throughput trade-off \cite{hanhirova2018latency}. 

Regarding power consumption, the 4.258W measured in the Raspberry Pi corresponds to the utilization of the whole board although, as previously explained, we have disabled as many peripherals as possible to make a fairer comparison. As for the Jetson Nano, 3.995W is obtained after applying a correction term (empirically inferred comparing the output of \textit{tegrastats} command and power meter measurement) to the addition of the CPU and GPU consumptions. Properly isolating the chip's consumption from the peripherals allows us to make a finer comparison with the Zynq Ultrascale+ whose 8.8W also represent the chip's consumption. With respect to energy utilization, the figures show that the higher power consumption of the AMD-Xilinx MPSoC is fully compensated by the latency acceleration achieved, making this the device with the most efficient processing, slightly better than the Jetson Nano.

\begin{table}[h!]
\centering
\resizebox{11.5cm}{!}{%
\begin{tabular}{|c|c|c|c|}
\hline
\textbf{Framework (Accelerator)} &
  \textbf{TF Lite (CPU)} &
  \multicolumn{1}{c|}{\textbf{TRT (GPU)}} &
  \textbf{Vitis AI (FPGA)} \\ \hline
\textbf{Data quantization} &
  \textbf{INT8} &
  \multicolumn{1}{c|}{\textbf{FP16/FP32}} &
  \textbf{INT8} \\ \hline
  \diagbox[]{\textbf{Measurement}}{\textbf{Platform}} &
  {\color[HTML]{CB0000} \textbf{RPI}} &
  \multicolumn{1}{c|}{{\color[HTML]{32CB00} \textbf{JN}}} &
  {\color[HTML]{3166FF} \textbf{ZCU}} \\ \hline
\textbf{Power (W)} &
  4.258 &
  \multicolumn{1}{c|}{3.995} & 8.8 \\ \hline
\textbf{Latency (s)} &
  {0.613} &
  \multicolumn{1}{c|}{0.093} &
  {0.037} \\ \hline
\textbf{Energy (J)} &
  {\textbf{2.608}} &
  \multicolumn{1}{c|}{\textbf{0.372}} &
  {\textbf{0.326}} \\ \hline
\end{tabular}}
\caption{Power, latency and energy consumption measurements regarding the quantization technique (8-bit integer, INT8 or 16/32-bit floating point FP16/FP32), framework (TensorFlow Lite, TF Lite; Tensor RT, TRT and Vitis AI), accelerator and platform (Raspberry Pi 4B (RPI), Jetson Nano (JN) and ZCU104 development board (ZCU)) used.}
\label{tab:inferenceBenchmarking}
\end{table}

In a two-stage pipelined processing implementation (image preprocessing + FCN inference), currently achievable throughput limit is imposed by the image preprocessing stage, imposing a limit of around 19,6 FPS. Although this stage could be hardware accelerated in the future to improve throughput, achieved image segmentation latency (50 ms) is adequate for most ADS/ADAS systems such as forward collision warning systems, emergency breaking systems, trajectory planning systems, etc \cite{estimateLatency}.

\newpage

Finally, we present results from the analysis of the memory footprint and the real performance of the segmentation. On Table \ref{tab:modelSizeChange}, it is interesting to note how the non-data buffers (responsible for storing, among other things, operators and subgraphs) have the same size for both experiments when quantization has still not been performed on the network. This is due to the fact that the difference between the two models lies solely in the number of filters of the final convolution layer and not in any other operator.

\begin{table}[h!]
\centering
\resizebox{14cm}{!}{%
\begin{tabular}{c|cccc|cccc|}
\cline{2-9}
 & \multicolumn{4}{c|}{\textbf{3-class experiment}} & \multicolumn{4}{c|}{\textbf{5-class experiment}} \\ \hline
\multicolumn{1}{|c|}{\textbf{Quantization}} & \multicolumn{2}{c|}{\textbf{No}} & \multicolumn{2}{c|}{\textbf{Yes}} & \multicolumn{2}{c|}{\textbf{No}} & \multicolumn{2}{c|}{\textbf{Yes}} \\ \hline
\multicolumn{1}{|c|}{\textbf{Buffer type}} & \multicolumn{1}{c|}{\textbf{Memory (KB)}} & \multicolumn{1}{c|}{\textbf{\%}} & \multicolumn{1}{c|}{\textbf{Memory (KB)}} & \textbf{\%} & \multicolumn{1}{c|}{\textbf{Memory (KB)}} & \multicolumn{1}{c|}{\textbf{\%}} & \multicolumn{1}{c|}{\textbf{Memory (KB)}} & \textbf{\%} \\ \hline
\multicolumn{1}{|c|}{\textbf{Non-data}} & \multicolumn{1}{c|}{\textbf{11.432}} & \multicolumn{1}{c|}{8.42} & \multicolumn{1}{c|}{\textbf{17.108}} & 35.06 & \multicolumn{1}{c|}{11.432} & \multicolumn{1}{c|}{8.41} & \multicolumn{1}{c|}{17.156} & 35.11 \\ \hline
\multicolumn{1}{|c|}{\textbf{Data}} & \multicolumn{1}{c|}{\textbf{124.392}} & \multicolumn{1}{c|}{91.58} & \multicolumn{1}{c|}{\textbf{31.684}} & 64.94 & \multicolumn{1}{c|}{124.464} & \multicolumn{1}{c|}{91.59} & \multicolumn{1}{c|}{31.708} & 64.89 \\ \hline
\multicolumn{1}{|c|}{\textbf{Total}} & \multicolumn{1}{c|}{135.824} & \multicolumn{1}{c|}{100} & \multicolumn{1}{c|}{48.792} & 100 & \multicolumn{1}{c|}{135.896} & \multicolumn{1}{c|}{100} & \multicolumn{1}{c|}{48.864} & 100 \\ \hline
\end{tabular}}
\caption{U-Net model size variation after applying full integer quantization.}
\label{tab:modelSizeChange}
\end{table}

Table \ref{tab:modelSizeChange} also shows that data buffers take up more than 90\% of the whole unquantized model. This is what allows the size of the complete model to be reduced considerably when quantization is applied, allowing it to be stored in a device with fewer memory resources. However, it has to be mentioned that there is a 50\% increase in the size of the non-data buffers after quantization. This fact is a consequence of the inclusion of new operators (scaling or casting, for example) which are needed for the aforementioned process. As for the segmentation variations, we show in Table \ref{tab:metricsOptimumConfiguration} the recall, precision and IoU after the different quantization approaches.

\begin{table}[h!]
\centering
\resizebox{12cm}{!}{
\begin{tabular}{cccccccccc}
\cline{2-10}
\multicolumn{1}{c|}{} & \multicolumn{9}{c|}{\textbf{Rebuilt}} \\ \hline
\multicolumn{1}{|c|}{\textbf{Metric}} & \multicolumn{3}{c|}{\textbf{Recall}} & \multicolumn{3}{c|}{\textbf{Precision}} & \multicolumn{3}{c|}{\textbf{IoU}} \\ \hline
\multicolumn{1}{|c|}{\diagbox[]{\textbf{Class name}}{\textbf{Device}}} & \multicolumn{1}{c|}{\textbf{RPI}} & \multicolumn{1}{c|}{\textbf{Nano}} & \multicolumn{1}{c|}{\textbf{ZCU}} & \multicolumn{1}{c|}{\textbf{RPI}} & \multicolumn{1}{c|}{\textbf{Nano}} & \multicolumn{1}{c|}{\textbf{ZCU}} & \multicolumn{1}{c|}{\textbf{RPI}} & \multicolumn{1}{c|}{\textbf{Nano}} & \multicolumn{1}{c|}{\textbf{ZCU}} \\ \hline
\multicolumn{1}{|c|}{\textbf{Road}} & \multicolumn{1}{c|}{98.28} & \multicolumn{1}{c|}{98.19} & \multicolumn{1}{c|}{97.97} & \multicolumn{1}{c|}{95.44} & \multicolumn{1}{c|}{95.72} & \multicolumn{1}{c|}{97.11} & \multicolumn{1}{c|}{93.87} & \multicolumn{1}{c|}{94.06} & \multicolumn{1}{c|}{95.19} \\ \hline
\multicolumn{1}{|c|}{\textbf{Road Marks}} & \multicolumn{1}{c|}{86.13} & \multicolumn{1}{c|}{84.42} & \multicolumn{1}{c|}{86.71} & \multicolumn{1}{c|}{80.98} & \multicolumn{1}{c|}{79.91} & \multicolumn{1}{c|}{81.12} & \multicolumn{1}{c|}{71.63} & \multicolumn{1}{c|}{69.64} & \multicolumn{1}{c|}{72.15} \\ \hline
\multicolumn{1}{|c|}{\textbf{Non-Drivable}} & \multicolumn{1}{c|}{92.87} & \multicolumn{1}{c|}{93.45} & \multicolumn{1}{c|}{95.59} & \multicolumn{1}{c|}{97.79} & \multicolumn{1}{c|}{97.71} & \multicolumn{1}{c|}{97.40} & \multicolumn{1}{c|}{90.96} & \multicolumn{1}{c|}{91.44} & \multicolumn{1}{c|}{93.21} \\ \hline
\multicolumn{1}{|c|}{\textbf{Overall}} & \multicolumn{1}{c|}{95.86} & \multicolumn{1}{c|}{95.98} & \multicolumn{1}{c|}{96.74} & \multicolumn{1}{c|}{95.84} & \multicolumn{1}{c|}{95.96} & \multicolumn{1}{c|}{96.77} & \multicolumn{1}{c|}{92.27} & \multicolumn{1}{c|}{92.50} & \multicolumn{1}{c|}{93.78} \\ \hline
\multicolumn{1}{|c|}{\textbf{Mean}} & \multicolumn{1}{c|}{92.42} & \multicolumn{1}{c|}{92.02} & \multicolumn{1}{c|}{93.42} & \multicolumn{1}{c|}{91.40} & \multicolumn{1}{c|}{91.11} & \multicolumn{1}{c|}{91.88} & \multicolumn{1}{c|}{85.49} & \multicolumn{1}{c|}{85.05} & \multicolumn{1}{c|}{86.85} \\ \hline
\multicolumn{1}{|c|}{\textbf{Weighted}} & \multicolumn{1}{c|}{87.08} & \multicolumn{1}{c|}{85.61} & \multicolumn{1}{c|}{87.78} & \multicolumn{1}{c|}{82.75} & \multicolumn{1}{c|}{81.80} & \multicolumn{1}{c|}{82.88} & \multicolumn{1}{c|}{73.89} & \multicolumn{1}{c|}{72.16} & \multicolumn{1}{c|}{74.52} \\ \hline
 &  &  &  &  &  &  &  &  &  \\ \hline
\multicolumn{1}{|c|}{\textbf{Road}} & \multicolumn{1}{c|}{90.22} & \multicolumn{1}{c|}{93.26} & \multicolumn{1}{c|}{90.37} & \multicolumn{1}{c|}{99.24} & \multicolumn{1}{c|}{98.75} & \multicolumn{1}{c|}{99.05} & \multicolumn{1}{c|}{89.60} & \multicolumn{1}{c|}{92.17} & \multicolumn{1}{c|}{89.59} \\ \hline
\multicolumn{1}{|c|}{\textbf{Road Marks}} & \multicolumn{1}{c|}{78.66} & \multicolumn{1}{c|}{75.91} & \multicolumn{1}{c|}{78.30} & \multicolumn{1}{c|}{78.53} & \multicolumn{1}{c|}{78.61} & \multicolumn{1}{c|}{78.69} & \multicolumn{1}{c|}{64.74} & \multicolumn{1}{c|}{62.92} & \multicolumn{1}{c|}{64.60} \\ \hline
\multicolumn{1}{|c|}{\textbf{Vegetation}} & \multicolumn{1}{c|}{96.90} & \multicolumn{1}{c|}{95.86} & \multicolumn{1}{c|}{96.02} & \multicolumn{1}{c|}{91.85} & \multicolumn{1}{c|}{95.73} & \multicolumn{1}{c|}{95.63} & \multicolumn{1}{c|}{89.23} & \multicolumn{1}{c|}{91.93} & \multicolumn{1}{c|}{91.99} \\ \hline
\multicolumn{1}{|c|}{\textbf{Sky}} & \multicolumn{1}{c|}{97.64} & \multicolumn{1}{c|}{97.51} & \multicolumn{1}{c|}{95.40} & \multicolumn{1}{c|}{92.85} & \multicolumn{1}{c|}{93.56} & \multicolumn{1}{c|}{95.40} & \multicolumn{1}{c|}{90.81} & \multicolumn{1}{c|}{91.38} & \multicolumn{1}{c|}{91.21} \\ \hline
\multicolumn{1}{|c|}{\textbf{Other}} & \multicolumn{1}{c|}{78.47} & \multicolumn{1}{c|}{84.01} & \multicolumn{1}{c|}{85.50} & \multicolumn{1}{c|}{57.29} & \multicolumn{1}{c|}{64.44} & \multicolumn{1}{c|}{57.34} & \multicolumn{1}{c|}{49.51} & \multicolumn{1}{c|}{57.40} & \multicolumn{1}{c|}{52.25} \\ \hline
\multicolumn{1}{|c|}{\textbf{Overall}} & \multicolumn{1}{c|}{90.58} & \multicolumn{1}{c|}{92.62} & \multicolumn{1}{c|}{91.08} & \multicolumn{1}{c|}{92.34} & \multicolumn{1}{c|}{93.61} & \multicolumn{1}{c|}{93.20} & \multicolumn{1}{c|}{84.41} & \multicolumn{1}{c|}{87.42} & \multicolumn{1}{c|}{85.66} \\ \hline
\multicolumn{1}{|c|}{\textbf{Mean}} & \multicolumn{1}{c|}{88.38} & \multicolumn{1}{c|}{89.31} & \multicolumn{1}{c|}{89.12} & \multicolumn{1}{c|}{83.95} & \multicolumn{1}{c|}{86.22} & \multicolumn{1}{c|}{85.22} & \multicolumn{1}{c|}{76.78} & \multicolumn{1}{c|}{79.16} & \multicolumn{1}{c|}{77.93} \\ \hline
\multicolumn{1}{|c|}{\textbf{Weighted}} & \multicolumn{1}{c|}{86.92} & \multicolumn{1}{c|}{86.25} & \multicolumn{1}{c|}{85.67} & \multicolumn{1}{c|}{82.65} & \multicolumn{1}{c|}{83.96} & \multicolumn{1}{c|}{82.28} & \multicolumn{1}{c|}{74.58} & \multicolumn{1}{c|}{75.06} & \multicolumn{1}{c|}{73.06} \\ \hline
\end{tabular}}
\caption{Recall, accuracy and IoU (Intersection over Union) of the quantized neural network inference running on the three tested computing platforms: Raspberry Pi 4B (RPI), Jetson Nano (JN) and ZCU104 development board (ZCU).}
\label{tab:metricsOptimumConfiguration}
\end{table}

Table \ref{tab:metricsOptimumConfiguration} includes some interesting results. With regard to the 3-class experiment, the 8-bit integer Raspberry Pi 4B and the AMD-Xilinx MPSoC quantization approaches (each device with its own quantization technique) offer similar and even better performance in several cases than the 16/32-bit floating point Jetson Nano. Furthermore, these results also verify that the use of overlapping patches improves the segmentation recall/precision. Nevertheless, the results from the 5-class experiment show that having added two classes to the initial segmentation application increases the challenge difficulty to such an extent that the best results are now obtained with the Jetson Nano. These results highlight the need to either explore alternative quantization or finetuning techniques or directly train the neural network in such a way that it is aware that it will be later quantized, a process known as Quantization Aware Training (QAT) \cite{quantModel}. 

Although metrics such as those listed in Table \ref{tab:metricsOptimumConfiguration} are a good indicator of the quality of inference, it is essential to inspect the inference results in the images, especially when dealing with weakly labelled datasets such as the one we are working with. Otherwise, the prediction of the neural network in the unlabelled pixels, which are equally as important as the other ones, is unknown.

\begin{figure}[h!]
\begin{subfigure}{0.32\linewidth}
\centering
\includegraphics[width=5.35cm]{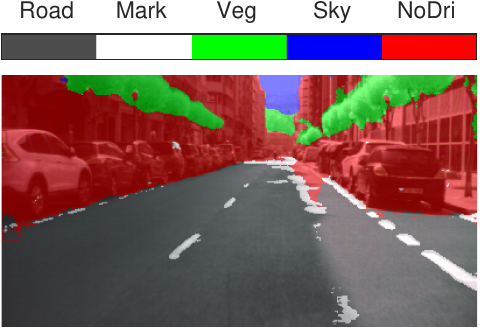}
\caption{Urban, Rasp. Pi 4B.}
\label{fig:segUrbanRPI5clases}
\end{subfigure}
\begin{subfigure}{0.32\linewidth}
\centering
\includegraphics[width=5.35cm]{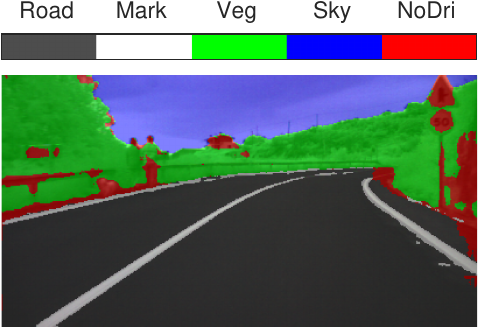}
\caption{Road, Rasp. Pi 4B.}
\label{fig:segRoadRPI5clases}
\end{subfigure}
\begin{subfigure}{0.31\linewidth}
\centering
\includegraphics[width=5.35cm]{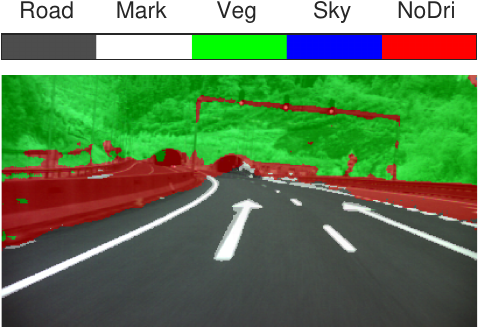}
\caption{Highway, Rasp. Pi 4B.}
\label{fig:segHighwayRPI5clases}
\end{subfigure}\\[1ex]
\begin{subfigure}{0.32\linewidth}
\centering
\includegraphics[width=5.35cm]{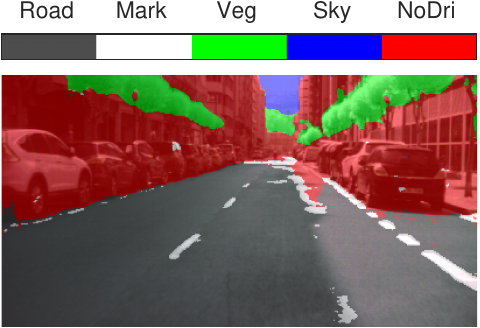}
\caption{Urban, Jetson Nano.}
\label{fig:segUrbanNano5clases}
\end{subfigure}%
\begin{subfigure}{0.33\linewidth}
\centering
\includegraphics[width=5.35cm]{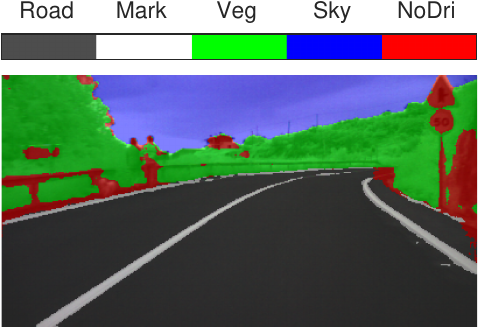}
\caption{Road, Jetson Nano.}
\label{fig:segRoadNano5clases}
\end{subfigure}
\begin{subfigure}{0.31\linewidth}
\centering
\includegraphics[width=5.35cm]{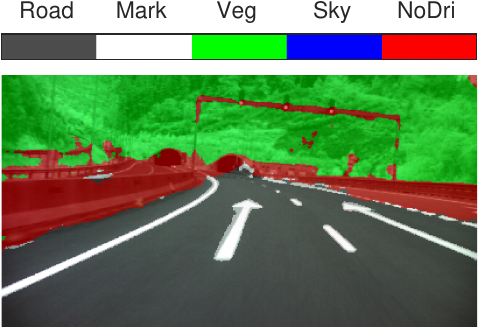}
\caption{Highway, Jetson Nano.}
\label{fig:segHighwayNano5clases}
\end{subfigure}\\[1ex]
\begin{subfigure}{0.32\linewidth}
\centering
\includegraphics[width=5.35cm]{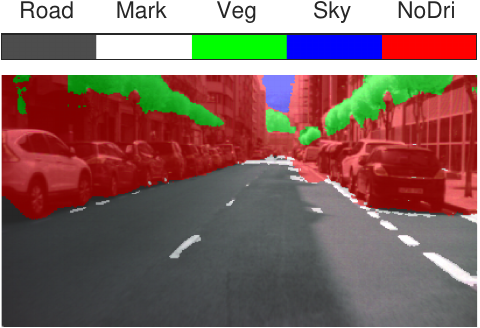} %(OIoU, 93.78)
\caption{Urban, ZCU 104.}
\label{fig:segUrbanZCU5clases}
\end{subfigure}%
\begin{subfigure}{0.33\linewidth}
\centering
\includegraphics[width=5.35cm]{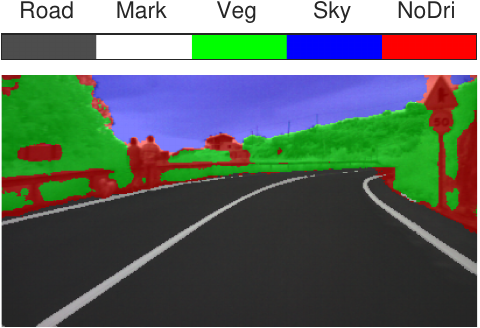} %(OIoU, 93.61)
\caption{Road, ZCU 104.}
\label{fig:segRoadZCU5clases}
\end{subfigure}
\begin{subfigure}{0.31\linewidth}
\centering
\includegraphics[width=5.35cm]{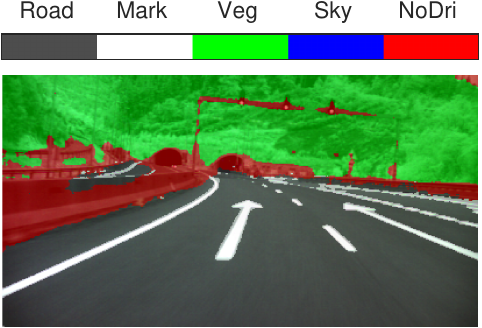} %OIoU, 94.96
\caption{Highway, ZCU 104.}
\label{fig:segHighwayZCU5clases}
\end{subfigure}
\caption{Results from the 5-class quantized U-Net inference on the benchmarked devices of three diverse images.}
\label{fig:comparisonQuantized}
\end{figure}

For the visual inspection of the results, we have chosen the same images as before. Although, at first sight, the three images all look similar, we will thoroughly analyze them one by one. As for the urban image, the output from the AMD-Xilinx MPSoC (Figure \ref{fig:segHighwayZCU5clases}) is that which has the smallest white spot in the foreground; however, it is also the one that fails to perceive the left side thin road lines. The output from Raspberry Pi 4B and Jetson Nano are almost identical (Figures \ref{fig:segHighwayRPI5clases} and \ref{fig:segHighwayNano5clases}). Regarding the road landscape, the segmentation of the Raspberry Pi 4B (Figure \ref{fig:segRoadRPI5clases}) is the one that makes the smallest error, followed by the Jetson Nano's (Figure \ref{fig:segRoadNano5clases}) in the vegetation to the right of the road sign and in the sky/vegetation intersections. However, they have less ability than the AMD-Xilinx MPSoC (Figure \ref{fig:segRoadZCU5clases}) to detect objects located at medium-long distances; this is evident in the sharpness of both the signal on the left and the guardrail underneath it, in the couple who is going for a walk and in the house at the back. Finally, in relation to the highway scene, which is challenging, once again the Raspberry Pi 4B and the Jetson Nano (Figures \ref{fig:segHighwayRPI5clases}, \ref{fig:segHighwayNano5clases}) show a different output compared to the AMD-Xilinx MPSoC (Figure \ref{fig:segHighwayZCU5clases}): while the former better identify
the lane lights before the tunnel entry, the latter is the only one to detect the road on the right side and to partially predict the road on the left. Nevertheless, the overall segmentation is extremely good. 

\section{Conclusions}\label{sec:conc}
HSI contains relevant spectral information that can help to develop more capable intelligent vision systems. In the field of ADAS and ADS, HSI can provide sufficient information to produce the correct segmentation of relevant items in driving scenarios and thus provide the vehicles with meaningful data for scene understanding. However, ADAS and ADS require capturing and processing HSI at video rates, which posses an additional challenge to the task of image segmentation. Nowadays, hyperspectral imaging at video rates in small formats is possible mainly by using snapshot cameras that incorporate standard CMOS imaging sensors with filter-on-chip technology. Filter-array technology constrains, to some extent, the quality of the captured spectral and spatial information, especially in outdoor uncontrolled environments, as is the case in real driving situations. Consequently, to achieve good classification performance and robustness, it is necessary to apply additional techniques beyond the mere use of raw spectral information.

It is known that the use of spatial information via convolution operations allows for obtaining a more robust segmentation of images by applying deep-learning techniques, but it is at the cost of a huge increase in the complexity of the models. However, the requirements of ADAS applications impose tight constraints on the complexity of the AI models, since they must be deployed in compliance with the low cost, low latency and low power requirements of embedded computing platforms. In this regard, we demonstrate that satisfactory segmentation results can be obtained for low/medium-complexity use cases when combining spectral information from a snapshot camera with the spatial information obtained with light encoder-decoder FCN models. In particular, we provide evidence that with a careful design and a hyperparameter optimization process, it is possible to achieve good performance with FCNs of sizes in the range of tens of thousands of parameters.  

We provide a benchmark of designed segmentation systems by the deployment on three significative embedded computing platforms: a multicore microprocessor, a SoC featuring an embedded GPU, and a PSoC containing an FPGA. We show that the FPGA-based PSoC approach is superior in terms of both latency and energy efficiency. The PSoC implementation allows for achieving almost 20 fps of processing throughput with a simple two-stage processing pipeline (preprocessing stage and inference stage), which meets the latency requirements of the ADAS industry standards. In fact, it is important to highlight that, unlike other works published in this field of application, our tested prototypes incorporate not only the acceleration of the neural network inference via the AI dedicated coprocessor, but also the necessary preprocessing of the acquired raw HSI which, in fact, reveals itself to be a bottleneck that needs to be taken into account in the image segmentation pipeline. In this sense, we also show that accelerating the preprocessing stage by taking advantage of both the thread-level and the data-level parallelism of modern microprocessor architectures in the writing of the code, makes the difference in terms of throughput.

Although the presented results are promising, HSI-based intelligent vision research is still at its early stages and there is much room for improvement. Future work will focus on extending the capabilities of the system at both the algorithmic and the implementation levels. It is expected that the ML-based algorithmic development and model training will be improved as a result of the availability of a greater amount of data. In this sense, the recently published new v2.0 release of the HSI-Drive dataset, which provides 496 additional labeled images extracted from new recordings performed in Winter and Autumn, will facilitate the training of models thanks to the increase of data available, particularly for the minority classes and for challenging scenarios such as rainy weather, low-lighting environments etc. In this regard, it will be necessary to investigate how to extract more meaningful spectral features that would be able to enhance class separability and how to effectively combine this information with the spatial features at different scales.

Regarding the deployment on embedded computing platforms that fit ADAS/ADS application requirements, on the one hand, it is necessary to further explore optimization techniques regarding model quantization, pruning, the use of separable convolutions etc. On the other hand, we are assessing the possibility of designing a more customized AI coprocessor to further accelerate the data flow and reduce resource occupation in order to enhance the system throughput and reduce power consumption.

\bibliographystyle{splncs04}
\bibliography{biblio.bib}

\end{document}